\documentclass{article}

\usepackage{arxiv}
\usepackage[utf8]{inputenc} % allow utf-8 input
\usepackage[T1]{fontenc}    % use 8-bit T1 fonts
\usepackage{hyperref}       % hyperlinks
\usepackage{url}            % simple URL typesetting
\usepackage{booktabs}       % professional-quality tables
\usepackage{amsfonts}       % blackboard math symbols
\usepackage{nicefrac}       % compact symbols for 1/2, etc.
\usepackage{microtype}      % microtypography
\usepackage{lipsum}
\usepackage{graphicx}
\graphicspath{ {./images/} }
\usepackage{authblk}         % 上标编号式作者/单位
\usepackage[numbers,sort&compress]{natbib}
\usepackage{placeins}          % 提供 \FloatBarrier
\usepackage{enumitem}
\usepackage[table]{xcolor} 
\usepackage{amsmath}
\usepackage{tabularx,makecell,array,ragged2e}
\newcolumntype{Y}{>{\raggedright\arraybackslash}X}         % 左对齐自适应
\newcolumntype{C}{>{\centering\arraybackslash}p{1.8cm}}    % 居中定宽（可微调）
\usepackage{chngcntr}
\counterwithin{table}{section}  % 表格按 section 计数：Table 4.1, 4.2, ...

\usepackage{listings}
\usepackage{xcolor}

\lstdefinelanguage{yaml}{
  keywords={true,false,null,yes,no,True,False,NULL},
  sensitive=false,
  comment=[l]{\#},
  morestring=[b]",           % strings in double quotes
  morestring=[s]{'}{'},      % strings in single quotes
}

\renewcommand{\today}{}      % 不显示日期
\usepackage{xltabular}  % 在导言区添加此包（自动加载 longtable + tabularx）
\usepackage{booktabs}
\usepackage[table]{xcolor}
\usepackage{caption}
\usepackage{adjustbox}  % 等比缩放
\usepackage{afterpage}

\title{Memory in Large Language Models: Mechanisms, Evaluation and Evolution}

\author{Dianxing Zhang\textsuperscript{1}}
\author{Wendong Li\textsuperscript{1,*}}
\author{Kani Song\textsuperscript{1,*}}
\author{Jiaye Lu\textsuperscript{1,*}}
\author{Gang Li\textsuperscript{1}}
\author{Liuchun Yang\textsuperscript{1}}
\author{Sheng Li\textsuperscript{1,\textdagger}}

\affil{Digital China AI Research Institute\\
\texttt{zhangdxh@digitalchina.com}\quad \texttt{lishengh@digitalchina.com}}

\usepackage{tikz}
\usetikzlibrary{mindmap}   % 必须加载 mindmap 库
\usepackage{subcaption}    % 如果你用 subfigure

\begin{document}
\maketitle

% 在标题页底部集中解释符号
\begingroup
\renewcommand{\thefootnote}{\fnsymbol{footnote}}
\footnotetext[1]{Equal contribution.}
\footnotetext[2]{Corresponding author.}
\endgroup

\begin{abstract}Under a unified operationalized definition, we define LLM "memory" as a persistent state that is written during pretraining, finetuning, or inference, can be subsequently addressed, and stably influences outputs. On this basis, we propose a four-way taxonomy (parametric, contextual, external, and procedural/episodic) and a "memory quadruple" (storage location—persistence—write/access path—controllability), and connect mechanism, evaluation, and governance through a causal chain of "write—read—inhibit/update." To avoid distorted comparisons arising from heterogeneous settings, we provide a three-setting parallel protocol (parameter-only, offline retrieval, online retrieval) that decouples model capability from information availability on the same data slice and timeline; we then construct a layered evaluation: parametric memory (closed-book recall, edit differential, memorization/privacy risks), contextual memory (position–performance curves and the "mid-sequence drop"), external memory (decoupling correctness from snippet-level attribution/faithfulness), and procedural/episodic memory (cross-session consistency and timeline replay, E-MARS+). The framework uniformly incorporates temporal governance and leakage auditing (Freshness hits, outdated answers, refusal slices), as well as uncertainty reporting via inter-rater agreement and paired tests/multiple-comparison correction. For updating and forgetting, we propose DMM-Gov dynamic governance: coordinating DAPT/TAPT, PEFT, model editing (ROME/MEND/MEMIT/SERAC), and RAG to form an auditable closed loop of "admission thresholds—progressive rollout—online monitoring—reversible rollback—change audit certificates", together with operational and acceptance specifications for timeliness/conflict handling and long-horizon consistency. Finally, we put forward four classes of testable propositions (minimum identifiability, a minimally sufficient evaluation card, causally constrained editing and verifiable forgetting, and the conditional boundary under which retrieval plus small-window replay is preferable to ultra-long-context direct reading), thereby providing a shared coordinate system and methodological baselines—reproducible, comparable, and governable—for research and deployment.
\end{abstract}


\section{Introduction}"When a physician consults an AI assistant for the latest treatment options but receives recommendations for a drug that was phased out three years ago; when an attorney asks an AI to retrieve case law and it confidently cites a statute that does not exist—these are not scenes from science fiction, but real risks arising from deficiencies in the ‘memory’ mechanisms of today’s large language models. As LLMs move from the lab into high-stakes domains such as healthcare, finance, and law, their ‘memory’—namely, the ability to store, update, and forget knowledge—has evolved from an academic topic into the lifeblood of safety, compliance, and trust." \citep{vu2023freshllmsrefreshinglargelanguage,chen2021datasetansweringtimesensitivequestions,Dhingra_2022,925786,owasp2023llmtop10}

The memory of large language models (LLMs) refers to a persistent state that is written at any of the stages of pretraining, finetuning, or inference and can subsequently be stably addressed, thereby systematically influencing model outputs. This capability spans the model’s entire life cycle and is the cornerstone for the transition from "language understanding" to "knowledge application." Early studies—such as the LAMA probes—first systematically validated, under the parametric-only (PO) setting, the model’s ability to recall facts from parameters, revealing its potential as an "unsupervised knowledge base." \citep{petroni2019languagemodelsknowledgebases,petroni2021kiltbenchmarkknowledgeintensive} Subsequent mechanistic research further indicated that the Transformer’s feed-forward layers (FFN) can be interpreted as key–value memory, providing a mechanistic handle for localizing and editing knowledge at the parameter level. \citep{geva2021transformerfeedforwardlayerskeyvalue,elhage2021mathematical,elhage2022toymodelssuperposition}

However, a flourishing academic literature cannot mask practical difficulties. The current research ecosystem on LLM memory faces three core challenges that seriously hinder reliable deployment:

Blurred conceptual boundaries. Definitions, life cycles, and intervention modes for parametric, contextual, and external memory are often conflated. Basic questions such as "In a RAG system, are documents ‘external memory’ or ‘contextual memory’?" lack consensus, leading to irreproducible experimental designs and non-comparable results. \citep{lewis2021retrievalaugmentedgenerationknowledgeintensivenlp,guu2020realmretrievalaugmentedlanguagemodel,borgeaud2022retro,gao2024retrievalaugmentedgenerationlargelanguage,joren2025sufficientcontextnewlens,Pan_2024}

Fragmented evaluation lenses. Assessments often lump disparate aspects together indiscriminately. RAG systems must measure retrieval quality and also evaluate the faithfulness and source attribution of generated outputs to evidence, yet existing benchmarks (e.g., RAGAS, RAGChecker) often conflate the two, or exhibit blind spots when handling cross-source conflicts and identifying key evidence in long documents . \citep{saadfalcon2024aresautomatedevaluationframework,ru2024ragcheckerfinegrainedframeworkdiagnosing,es2025ragasautomatedevaluationretrieval,friel2025ragbenchexplainablebenchmarkretrievalaugmented,yang2024cragcomprehensiverag,sorodoc2025garagebenchmarkgroundingannotations,suri2025visdommultidocumentqavisually,wang2025retrievalaugmentedgenerationconflictingevidence,fabbri2022qafactevalimprovedqabasedfactual,min2023factscorefinegrainedatomicevaluation,li2024towards}

Bias in automated judging. Evaluations relying on LLM-as-a-Judge have been shown to suffer from serious position, order, and self-preference biases, causing "spurious significance" to be mistaken for real progress. \citep{zheng2023judgingllmasajudgemtbenchchatbot,wang2023largelanguagemodelsfair,castillobolado2024promptsdynamicconversationalbenchmarking}

\begin{figure}[htbp]
  \centering
  \includegraphics[width=\textwidth]{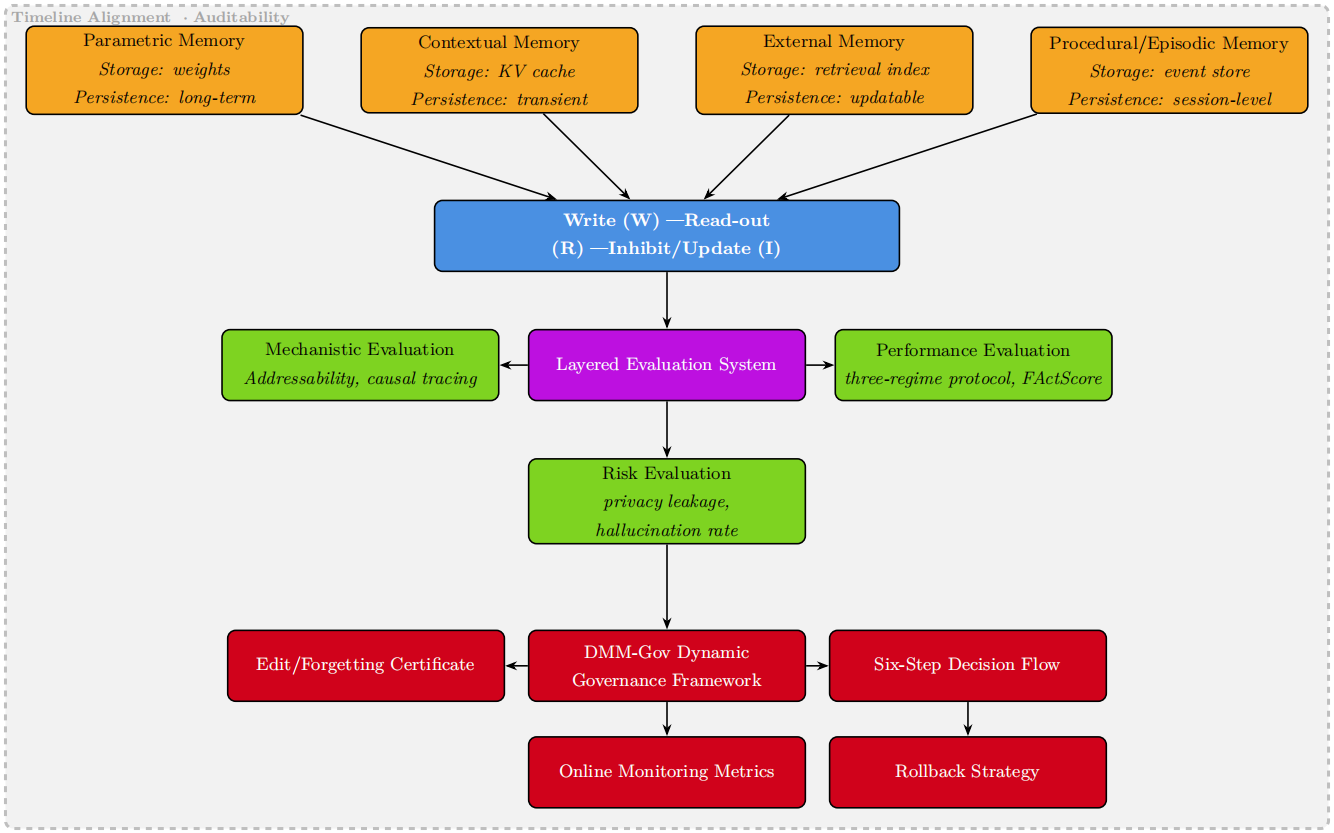}
  \caption{A unified framework for LLM memory research: mechanisms, evaluation, and governance.}
  \label{fig:llm_memory_framework}
\end{figure}

In response to these challenges, this paper aims to build an end-to-end "operating system" for LLM memory governance. Rather than merely proposing a set of new methods, we seek to construct a unified, operational analytical framework(Figure~\ref{fig:llm_memory_framework}) that bridges the gap between academic research and industrial practice. Our key contributions are:
\begin{enumerate}[label=(\roman*)]
\item \textbf{ Unified definitions and taxonomy.} We propose an operationalized definition—memory = persistent and addressable state—and, via a quadruple of storage location—persistence—update path—access method, characterize parametric, contextual (working), external (non-parametric), and procedural (episodic) memory; we clarify their boundaries vis-à-vis knowledge/ability/context state, linking them to reproducible experimental designs; \citep{petroni2019languagemodelsknowledgebases,geva2021transformerfeedforwardlayerskeyvalue,Pan_2024,packer2024memgptllmsoperatingsystems}
\item \textbf{ Survey of memory mechanisms.}  Using Retrieval (R)—Write (W)—Inhibit (I) as the main thread, we cover the training phase (W: compressive writing of the corpus into weights; R: differentiable retrieval/data selection), the inference phase (R: context retrieval/external injection; W: context loading and procedural writes), and post-training shaping (I: control interfaces for instruction/preference alignment and for editing/forgetting), assembling mechanistic evidence and delineating attainable limits and risks (verbatim memorization, privacy exposure, and side-effect propagation) ; \citep{raffel2023exploringlimitstransferlearning,gururangan2020dontstoppretrainingadapt,houlsby2019parameterefficienttransferlearningnlp,lester2021powerscaleparameterefficientprompt,li2021prefixtuningoptimizingcontinuousprompts,lewis2021retrievalaugmentedgenerationknowledgeintensivenlp,guu2020realmretrievalaugmentedlanguagemodel,izacard2021leveragingpassageretrievalgenerative,izacard2022atlasfewshotlearningretrieval,karpukhin2020densepassageretrievalopendomain,khattab2020colbertefficienteffectivepassage,thakur2021beirheterogenousbenchmarkzeroshot,meng2023locatingeditingfactualassociations,mitchell2022fastmodeleditingscale,mitchell2022memorybasedmodeleditingscale,meng2023masseditingmemorytransformer,yin2023historymatterstemporalknowledge,peng2024eventlevelknowledgeediting,shi2024musemachineunlearningsixway,maini2024tofutaskfictitiousunlearning,zhang2024negativepreferenceoptimizationcatastrophic,jin2024rwkubenchmarkingrealworldknowledge,li2024wmdpbenchmarkmeasuringreducing,belrose2023leace,ravfogel2024linearadversarialconcepterasure,ravfogel2022adversarial,ravfogel2020nulloutguardingprotected,carlini2021extracting,duan2024membershipinferenceattackswork,carlini2023quantifyingmemorizationneurallanguage,kandpal2022deduplicatingtrainingdatamitigates,lee2022deduplicatingtrainingdatamakes,gao2020pile800gbdatasetdiverse}
\item \textbf{ Evaluation framework.} We propose layered evaluation and unified metric design across the four memory types: parametric memory uses the PO setting plus pre/post-edit differentials; long-context focuses on positional robustness and sensitivity to lost-in-the-middle; external memory adopts a dual-channel assessment of retrieval quality × faithfulness and source attribution; procedural memory emphasizes cross-session consistency and trajectory replay. We additionally recommend timestamp-aligned protocols for time-sensitive scenarios; \citep{liu2023lostmiddlelanguagemodels,hsieh2024middlecalibratingpositionalattention,hsieh2024rulerwhatsrealcontext,bai2024longbenchbilingualmultitaskbenchmark,yen2025helmetevaluatelongcontextlanguage,zhang2024inftybenchextendinglongcontext,shaham2022scrollsstandardizedcomparisonlong,dong2024bamboocomprehensivebenchmarkevaluating,yuan2024lvevalbalancedlongcontextbenchmark,wang2025multimodalneedlehaystackbenchmarking,saadfalcon2024aresautomatedevaluationframework,ru2024ragcheckerfinegrainedframeworkdiagnosing,es2025ragasautomatedevaluationretrieval,friel2025ragbenchexplainablebenchmarkretrievalaugmented,yang2024cragcomprehensiverag,sorodoc2025garagebenchmarkgroundingannotations,suri2025visdommultidocumentqavisually,fabbri2022qafactevalimprovedqabasedfactual,min2023factscorefinegrainedatomicevaluation,li2024towards,wu2025longmemevalbenchmarkingchatassistants,huet2025episodicmemoriesgenerationevaluation,Dhingra_2022,vu2023freshllmsrefreshinglargelanguage,chen2021datasetansweringtimesensitivequestions}
\item \textbf{ Forgetting and updating.} We discuss evaluation protocols and side-effect boundaries for model editing/machine unlearning, with a Pareto analysis along three axes—target suppression, neighborhood preservation, and downstream steady state—supplemented by security regression tests such as membership inference and data extraction. \citep{bourtoule2021machine,shi2024musemachineunlearningsixway,maini2024tofutaskfictitiousunlearning,ji2024reversingforgetretainobjectivesefficient,zhang2024negativepreferenceoptimizationcatastrophic,jin2024rwkubenchmarkingrealworldknowledge,li2024wmdpbenchmarkmeasuringreducing,duan2024membershipinferenceattackswork,carlini2021extracting,carlini2023quantifyingmemorizationneurallanguage}
\item \textbf{ Deployment-oriented engineering principles.} We distill five principles—external-memory-first, small-step editing, long-task read/write strategies, timestamp alignment, and debiasing for privacy and evaluation—forming an actionable governance checklist (see §5.4), together with evaluation and online monitoring metrics such as RAGBench/RAGAS/RAGChecker. \citep{joren2025sufficientcontextnewlens,gao2024retrievalaugmentedgenerationlargelanguage,dai2019transformerxlattentivelanguagemodels,rae2019compressivetransformerslongrangesequence,xiao2024efficientstreaminglanguagemodels,munkhdalai2024leavecontextbehindefficient,chen2023extendingcontextwindowlarge,su2023roformerenhancedtransformerrotary,press2022trainshorttestlong,tang2024razorattentionefficientkvcache,hu2025epicefficientpositionindependentcaching,kim2025kvzipqueryagnostickvcache,hu2025efficientlongcontextllminference,lee2025infinitehipextendinglanguagemodel,kim2024infinipotinfinitecontextprocessing,925786,owasp2023llmtop10}
\end{enumerate}

Through this framework, we not only provide the research community with reproducible and comparable baselines, but also furnish the industry with a methodology for building LLM applications that are safe, reliable, and compliant. We hope this "operating system" will bridge theory and practice, advancing LLM memory research from a patchwork of techniques toward a unified engineering blueprint.

The remainder of this paper is organized as follows: Section 2 introduces the technical background and core pain points of LLM memory; Section 3 presents unified definitions, taxonomy, and formation mechanisms; Section 4—the core of this paper—details a layered evaluation framework covering the four memory types; Section 5 discusses strategies and governance frameworks for memory updating and forgetting; Section 6 analyzes current challenges and outlines future directions; Section 7 concludes.

\section{Technical Background of LLM Memory}
\subsection{From Language Models to the Problem Space of "Memory Systems"}
Contemporary large language models are not merely conditional distribution approximators; rather, along the full chain of pretraining–finetuning–inference they continually deposit and repeatedly invoke information states that are addressable. Early evaluations under the parametric-only (PO) (closed-book) setting show that, without accessing external evidence, models can directly recall a substantial amount of relational knowledge from parameters—an observation summarized as the feasibility and limits of "LM-as-KB" . \citep{petroni2019languagemodelsknowledgebases,petroni2021kiltbenchmarkknowledgeintensive,luo2023systematicassessmentfactualknowledge} Subsequent mechanistic studies further indicate that the Transformer’s feed-forward layers (FFN/MLP) form mid-layer storage structures approximating key–value pairs, and that specific pathways can read out subject→attribute associations from the weights, making these layers the primary carriers of factual recall . \citep{geva2021transformerfeedforwardlayerskeyvalue,elhage2021mathematical,elhage2022toymodelssuperposition,meng2023locatingeditingfactualassociations} Meanwhile, observations at the level of attention circuits have revealed reusable substructures such as induction heads, which copy patterns and align sequence positions under few-shot prompting, furnishing visible evidence for style continuation and short-term maintenance at inference time. \citep{olsson2022incontextlearninginductionheads,akyurek2023what,vonoswald2023transformerslearnincontextgradient,oren2024transformersmultistaternns} From this perspective, language models naturally possess multiform memory: part of it sinks into parameters and architecture, and part of it surfaces during inference as activations or external components, jointly shaping the upper bounds of factual recall, long-document utilization, and dialog coherence.
Formal definitions and type boundaries are given in §3.1–§3.2; causal evidence for "write—read—inhibit/update (R–W–I)" is presented in §3.3.
\subsection{Where Memory Lives and How It Is Invoked: From Network Internals to System Extensions}
A commonly held mechanistic hypothesis is that multi-head attention acts as a content-addressable read operator, aggregating the context under a given query; the feed-forward network provides a parametric storage/write channel, imprinting high–mutual-information structures into the weights, thereby explaining the coexistence of robustness for commonsense patterns and fragility for long-tail knowledge . \citep{elhage2021mathematical,geva2021transformerfeedforwardlayerskeyvalue,elhage2022toymodelssuperposition} This division of labor is not exclusive; inter-layer and intra-layer coupling and exceptions persist. To break the temporal and capacity limits of single-segment inputs, Transformer-XL extends the visible time horizon to multi-segment sequences via segment-level recurrence and relative positional encoding, enabling cross-segment replay of historical activations and key–value caches; the Compressive Transformer retains more distant cues through hierarchical compression without significantly increasing cost. \citep{dai2019transformerxlattentivelanguagemodels,rae2019compressivetransformerslongrangesequence} However, "visible" does not automatically mean "usable": as context length increases, attention becomes diluted and positional biases systematically depress the utilization of mid-span information—an failure mode repeatedly validated in subsequent long-context evaluations . \citep{liu2023lostmiddlelanguagemodels,hsieh2024rulerwhatsrealcontext,hsieh2024middlecalibratingpositionalattention} The corresponding position–performance curves and mid-span drop will serve as core robustness metrics in §4.3.

Relying on weights alone cannot cover timeliness and the long tail; accordingly, systems introduce non-parametric evidence paths at inference time. Retrieval-augmented generation (RAG) incorporates external documents into the generation loop via index–retrieve–fuse, improving factual consistency and traceability without altering weights; going further, REALM jointly optimizes the retriever and the language model end-to-end so that "finding sources" is acquired already during pretraining; RETRO connects by cross-attention to trillion-scale external stores, structurally loosening the equation "parameter scale = knowledge capacity"; kNN-LM performs local posterior correction via nearest neighbors, providing plug-and-play support for long-tail tokens and new-domain concepts . \citep{lewis2021retrievalaugmentedgenerationknowledgeintensivenlp,guu2020realmretrievalaugmentedlanguagemodel,borgeaud2022retro,khandelwal2020generalizationmemorizationnearestneighbor,gao2024retrievalaugmentedgenerationlargelanguage} Existing evidence shows that end-to-end performance is highly sensitive to retrieval recall, and reranking often yields notable joint gains in recall and faithfulness. \citep{karpukhin2020densepassageretrievalopendomain,thakur2021beirheterogenousbenchmarkzeroshot,khattab2020colbertefficienteffectivepassage,salemi2024evaluatingretrievalqualityretrievalaugmented,saadfalcon2024aresautomatedevaluationframework}

As models move from dialog assistants to agents with sustained goals and planning, the system must also structure and replay key events and intermediate states from interaction, in order to maintain cross-session consistency and track long-term objectives—Generative Agents and MemGPT provide operational blueprints via layered memory stores and "virtual memory"-style scheduling, respectively . \citep{park2023generativeagentsinteractivesimulacra,packer2024memgptllmsoperatingsystems} The resulting picture is one of internal–external collaboration: the network interior consolidates general patterns, while system extensions supply fresh evidence and a process timeline; together they determine both achievable performance and governability. Component-level metrics and alignment criteria for external memory are given in §4.4, and evaluation for procedural/episodic memory appears in §4.5.
\subsection{Principal Pain Points and Design Trade-offs of Memory Systems}

\begin{enumerate}[label=(\arabic*), leftmargin=2.2em, itemsep=0.8ex]
\item \textbf{The tension between controllable writes and revocability.}
Self-supervised training tends to imprint high–mutual-information and highly repeated fragments into the weights, which not only yields robust commonsense recall but also elevates risks of verbatim reproduction and privacy leakage. Under black-box conditions, training-data extraction has been demonstrated on early models, showing that specially crafted queries can reconstruct long spans of training samples. \citep{carlini2021extracting}
Engineering de-duplication can reduce regurgitation rates, shorten convergence, and weaken privacy threats that use highly repetitive samples as attack vectors, all without materially harming perplexity; however, poorly chosen thresholds introduce new trade-offs between coverage and quality. \citep{lee2022deduplicatingtrainingdatamakes,kandpal2022deduplicatingtrainingdatamitigates}
Regarding the decision of ``seen vs. unseen,'' conventional membership inference often approaches random-guess accuracy under standard pretraining, and is highly sensitive to distributional and temporal drift; hence, more robust risk assessment should jointly consider de-duplication thresholds, model scale, the corpus timeline, and evaluation protocols within a unified framework, rather than drawing conclusions from a single attack metric. \citep{duan2024membershipinferenceattackswork,carlini2023quantifyingmemorizationneurallanguage,gao2020pile800gbdatasetdiverse}

\item \textbf{For long context, ``visible'' is not the same as ``usable.''}
Enlarging the context window increases visibility, but attention allocation and positional bias systematically depress the utilization of mid-span evidence, so long context does not automatically translate into effective read/write. This failure mode has been repeatedly observed in Lost in the Middle and subsequent evaluations. \citep{liu2023lostmiddlelanguagemodels,hsieh2024rulerwhatsrealcontext}
Accordingly, structured reordering, anchor-guided prompting, and positional reweighting become necessary strategic complements. \citep{hsieh2024middlecalibratingpositionalattention,press2022trainshorttestlong,chen2023extendingcontextwindowlarge,su2023roformerenhancedtransformerrotary,xiao2024efficientstreaminglanguagemodels,munkhdalai2024leavecontextbehindefficient}

\item \textbf{Reliability and governance cost of external evidence paths.}
External memory (e.g., RAG) provides timeliness and traceability, while also importing the errors of the index–embedding–reranking–fusion pipeline into the generation loop: mismatches between embeddings and chunking can introduce noisy documents; cross-source evidential conflicts require explicit consistency adjudication and refusal mechanisms; the freshness and rollback capability of the index directly affect release cadence and the compliance audit trail. \citep{ru2024ragcheckerfinegrainedframeworkdiagnosing,saadfalcon2024aresautomatedevaluationframework,niu2024ragtruthhallucinationcorpusdeveloping,yang2024cragcomprehensiverag,friel2025ragbenchexplainablebenchmarkretrievalaugmented,sorodoc2025garagebenchmarkgroundingannotations,suri2025visdommultidocumentqavisually,wang2025retrievalaugmentedgenerationconflictingevidence,vu2023freshllmsrefreshinglargelanguage,Dhingra_2022,925786,owasp2023llmtop10}
Layered evaluation of retrieval quality and faithfulness/source attribution metrics is presented in Chapter 4.

\item \textbf{Management of event and temporal structure.}
As tasks expand from single-turn Q\&A to long-horizon interaction, factual memory alone is insufficient; the system must treat process as a first-class citizen, maintaining behavioral consistency and progress toward long-term goals via queryable timelines and summary replay. \citep{park2023generativeagentsinteractivesimulacra,packer2024memgptllmsoperatingsystems,ge2025tremuneurosymbolictemporalreasoning,huet2025episodicmemoriesgenerationevaluation,wu2025longmemevalbenchmarkingchatassistants,xiong2025memorymanagementimpactsllm,zhou2025mementofinetuningllmagents,gutierrez2025hipporagneurobiologicallyinspiredlongterm,qian2025memoragboostinglongcontext}
Omitting this layer leads to failure to retrieve key evidential snippets; conversely, writing procedural details directly into the weights induces non-revertibility and accumulation of side effects.

\item \textbf{Choosing the locus for in-parameter edits and compliant forgetting.}
Pointwise editing shows that write locations can be localized and manipulated: ROME rewrites factual associations via rank-1 updates at causal mediation sites in mid-layer MLPs; MEND performs rapid local adjustments via low-rank gradient transformations; MEMIT scales direct editing to tens of thousands of facts. \citep{meng2023locatingeditingfactualassociations,mitchell2022fastmodeleditingscale,mitchell2022memorybasedmodeleditingscale,meng2023masseditingmemorytransformer}
Nonetheless, there is inherent tension among effectiveness, locality, and generalization; sequential or batched edits readily trigger off-target effect diffusion. \citep{gu2024modeleditingharmsgeneral,rosati2024longformevaluationmodelediting,hsueh2024editingmindgiantsindepth,zhang2025uncoveringoverfittinglargelanguage,li2024reallyeditlanguagemodels}
A reasonable division of responsibilities and boundaries vis-à-vis external evidence paths determines the engineering feasibility of updates and forgetting. \citep{yin2023historymatterstemporalknowledge,peng2024eventlevelknowledgeediting,jiang2024learningeditaligningllms,wang2024wiserethinkingknowledgememory,belrose2023leace,ravfogel2024linearadversarialconcepterasure,ravfogel2022adversarial,ravfogel2020nulloutguardingprotected,maini2024tofutaskfictitiousunlearning,shi2024musemachineunlearningsixway,ji2024reversingforgetretainobjectivesefficient,li2024wmdpbenchmarkmeasuringreducing,jin2024rwkubenchmarkingrealworldknowledge}
\end{enumerate}

\noindent\textbf{Synthesis.}
Parametric memory offers recall under the PO setting but is pressured on precise controllability and revocability; runtime contextual memory (CM) enlarges visibility, whose usability depends on positional and read/write strategies; external memory (EM) and procedural memory bring timeliness and traceability, while requiring rigorous component-level governance and consistency adjudication.
Type taxonomy and decision criteria appear in Chapter 3; evaluation dimensions and protocols in Chapter 4; and engineering pathways and governance frameworks for updating/forgetting in Chapter 5. \citep{petroni2019languagemodelsknowledgebases,liu2023lostmiddlelanguagemodels,lewis2021retrievalaugmentedgenerationknowledgeintensivenlp,park2023generativeagentsinteractivesimulacra,packer2024memgptllmsoperatingsystems,Pan_2024,bai2024longbenchbilingualmultitaskbenchmark,hsieh2024rulerwhatsrealcontext,saadfalcon2024aresautomatedevaluationframework,ru2024ragcheckerfinegrainedframeworkdiagnosing,es2025ragasautomatedevaluationretrieval,yang2024cragcomprehensiverag}

\section{Taxonomy and Mechanisms of LLM Memory}
\subsection{Conceptual Boundaries and an Operationalized Definition}
The prevailing literature indicates that large language models do more than conditionally generate on a given context: across different stages of training and inference they form collections of states that can be stably addressed at subsequent time steps and influence outputs. Accordingly, we define LLM memory as a persistent state that is written during pretraining, finetuning, or inference, and that can later be read to produce systematic effects on the model’s outputs. To avoid equating "memory" with any single substrate, our analysis records four dimensions—storage location, persistence, write path, and access method—without presupposing a one-to-one mapping to specific implementations. Thus, long-term representations fixed in parameters and activations/caches during inference fall under the same semantic category, differing primarily in persistence and access path; external document stores, vector stores, and tool/API returns are likewise treated as addressable external states, provided they enter the generation loop and exert reproducible influence on outputs. This definition aligns with empirical observations that language models can recall facts under the parametric-only (PO) setting, and it is corroborated by mechanistic evidence that interprets Transformer feed-forward layers as key–value memory, thereby furnishing a unified semantics and testable interface for subsequent measurement and ablations \citep{petroni2019languagemodelsknowledgebases,geva2021transformerfeedforwardlayerskeyvalue,elhage2021mathematical}.

Within this framework, memory, knowledge, ability, and context state exhibit a layered relationship. Knowledge denotes facts or relations that can be stably reproduced and verified, an important subset of which exists in parametric form and can be recalled under the PO setting by probes such as LAMA \citep{petroni2019languagemodelsknowledgebases,petroni2021kiltbenchmarkknowledgeintensive}. Ability emphasizes task-facing algorithmic processes, often loaded on-the-fly during the forward pass as in-context learning; at the circuit level, studies document induction heads that copy and align across positions, while at the algorithmic level this loading is explained as implicit Bayesian updating or approximate gradient descent under fixed weights \citep{olsson2022incontextlearninginductionheads,akyurek2023what,vonoswald2023transformerslearnincontextgradient,oren2024transformersmultistaternns}. Context state refers to the instantaneous token sequence and KV cache exposed to the model at inference; its effective time horizon can be extended via segment-level recurrence or compressed replay, but it does not automatically become long-term memory \citep{dai2019transformerxlattentivelanguagemodels,rae2019compressivetransformerslongrangesequence}. Compared with these notions, memory stresses a sustainable, addressable, and read–writable set of states that encompasses both long-term parametric representations and evidence/intermediate variables retrieved or injected via tools \citep{lewis2021retrievalaugmentedgenerationknowledgeintensivenlp,guu2020realmretrievalaugmentedlanguagemodel,borgeaud2022retro,khandelwal2020generalizationmemorizationnearestneighbor}.

From a life-cycle perspective, pretraining (with cross-entropy objectives) imprints high–mutual-information structures into parameters; finetuning and model editing alter activation thresholds and call probabilities of these structures; inference loads and reads relevant states via in-context learning, retrieval, and tool use. Risks along this path can be described uniformly: when the corpus contains low-entropy, highly repeated fragments, the probability of verbatim memorization and privacy exposure increases; conversely, systematic de-duplication can reduce regurgitation and extraction rates without materially harming perplexity, suggesting that memory evaluation should be co-designed with data governance \citep{carlini2021extracting,lee2022deduplicatingtrainingdatamakes,kandpal2022deduplicatingtrainingdatamitigates,carlini2023quantifyingmemorizationneurallanguage,duan2024membershipinferenceattackswork}. The present section provides operational terminology and observables for the unified evaluation in §4 and for update/forgetting governance in §5.

\subsection{Memory Types and Their Boundaries}
\noindent
Without altering the above definition, we minimally distinguish memory types by their observable carriers and invocation modes.

\begin{enumerate}[label=(\roman*), leftmargin=2.2em, itemsep=0.8ex]
  \item \textbf{Parametric memory.} Representations consolidated in the weights (especially FFN layers) through pretraining/finetuning; they support factual recall and pattern continuation under the PO setting but are sensitive to time-critical knowledge and long-tail facts, with controllability depending on dedicated editing and rollback mechanisms \citep{petroni2019languagemodelsknowledgebases,geva2021transformerfeedforwardlayerskeyvalue,meng2023locatingeditingfactualassociations,mitchell2022fastmodeleditingscale,meng2023masseditingmemorytransformer}.
  \item \textbf{Contextual memory.} Activations and caches formed on the fly during inference by inputs and history; they enable style alignment and rule fitting under few-shot prompting, but are markedly affected by position and length. Multiple evaluations report that mid-span evidence is harder to utilize effectively in ultra-long sequences; this effect persists even as the visible window is enlarged \citep{liu2023lostmiddlelanguagemodels,hsieh2024rulerwhatsrealcontext,hsieh2024middlecalibratingpositionalattention}.
  \item \textbf{External or ``non-parametric'' memory.} Mechanisms that incorporate documentary evidence or nearest neighbors into the generative distribution via retrieval; advantages include updatability and auditability, with canonical pathways including retrieval-augmented generation, differentiable retrieval during pretraining, and nearest-neighbor interpolation. At the same time, errors in indexing, chunking, and fusion can inject noise and conflict into answers, necessitating component-level constraints on faithfulness and source attribution \citep{lewis2021retrievalaugmentedgenerationknowledgeintensivenlp,guu2020realmretrievalaugmentedlanguagemodel,borgeaud2022retro,khandelwal2020generalizationmemorizationnearestneighbor,salemi2024evaluatingretrievalqualityretrievalaugmented,saadfalcon2024aresautomatedevaluationframework,ru2024ragcheckerfinegrainedframeworkdiagnosing,min2023factscorefinegrainedatomicevaluation,li2024towards,fabbri2022qafactevalimprovedqabasedfactual,honovich2022truereevaluatingfactualconsistency}.
\end{enumerate}

\noindent
When cross-session persistence and long-horizon behavioral consistency are required, intermediate steps, event snippets, and timelines from interaction can be stored in structured form and replayed as needed; this procedural/episodic memory emphasizes temporal structure and re-playability and often shares infrastructure with external memory in practice \citep{park2023generativeagentsinteractivesimulacra,packer2024memgptllmsoperatingsystems,maharana2024evaluatinglongtermconversationalmemory,hu2025evaluatingmemoryllmagents}.

\medskip

\noindent\textbf{Boundary ambiguity and coupling in practice: the ``dual identity'' problem in RAG.}
The boundary between contextual and external memory is not cleanly separable at the system level; instead, there is substantial coupling and overlap. The canonical case is RAG: documents returned by the retriever must be serialized and concatenated into the prompt before reaching the generator, thereby becoming part of the model’s current context window.
% (概念阐释段落可不必引文)

\medskip

\noindent\textbf{Evaluation priorities and metric decoupling.}
To address this, we advocate, in coupled scenarios (e.g., RAG), a source-first, effect-separation principle:

\begin{description}[leftmargin=2.2em, style=nextline]
  \item[Priority:] prioritize source attribution. For diagnosing bottlenecks or reporting core capabilities, first assess the contribution of external memory—i.e., retrieval quality (Recall@k, nDCG) and faithfulness/attribution to evidence (e.g., FActScore, Citation Recall). External memory is the defining value of RAG relative to purely parametric models; its updatability and auditability are the design raison d’être. If the retrieved document is wrong or irrelevant, higher downstream context-utilization cannot compensate.
  \item[Metric choice:] explicit decoupling. To avoid misattributing poor context utilization as ineffective external knowledge, employ layered metrics:
\end{description}

\begin{enumerate}[label=Layer~\arabic*:, leftmargin=2.6em, itemsep=0.6ex]
  \item \textbf{Retrieval-quality metrics.} Does the system find the correct evidence? (e.g., Recall@5, nDCG@10) \citep{thakur2021beirheterogenousbenchmarkzeroshot,karpukhin2020densepassageretrievalopendomain,salemi2024evaluatingretrievalqualityretrievalaugmented}.
  \item \textbf{Attribution and faithfulness metrics.} Is the generated answer faithful to the retrieved evidence, with correct citations? (e.g., FActScore, Citation Precision/Recall) \citep{min2023factscorefinegrainedatomicevaluation,li2024towards,fabbri2022qafactevalimprovedqabasedfactual,honovich2022truereevaluatingfactualconsistency,wang2020askingansweringquestionsevaluate}.
  \item \textbf{Context-utilization efficiency.} Conditional on correct and cited evidence, evaluate how efficiently the model uses this now-contextualized external evidence—e.g., place the same evidence at beginning/middle/end to measure positional losses \citep{liu2023lostmiddlelanguagemodels,hsieh2024middlecalibratingpositionalattention,hsieh2024rulerwhatsrealcontext}.
\end{enumerate}

\medskip

\noindent\textbf{Boundary judgments should follow observable behavior.}
Stable answering under the PO setting without external materials indicates parametric memory; phenomena that require placing information in the prompt/history indicate contextual memory; answers that depend on citable external evidence reflect external memory; and behaviors that require cross-turn replay of structured events/timelines to maintain consistency involve procedural/episodic memory. For borderline cases, combine clues such as whether information entered the weights, whether it persisted across sessions, and whether it entered the generation loop via retrieval/tools. Note that retrievers and caches do not automatically constitute memory; they count only when their returns enter the generation loop and reproducibly affect outputs \citep{lewis2021retrievalaugmentedgenerationknowledgeintensivenlp,guu2020realmretrievalaugmentedlanguagemodel,borgeaud2022retro}. Regarding parametric vs. contextual memory, it is apt to treat weights as ``what is knowable on call,'' and activations/caches as ``what is currently called.'' They differ systematically in substrate, time scale, write path, and risk profile: the former is persistent and hard to revoke, prone to verbatim memorization/data extraction; the latter is transient and failure-prone, with issues arising from positional bias and misreads. In measurement and governance, these differences map to distinct metrics and processes \citep{carlini2021extracting,lee2022deduplicatingtrainingdatamakes,kandpal2022deduplicatingtrainingdatamitigates,liu2023lostmiddlelanguagemodels,hsieh2024rulerwhatsrealcontext}. This typology provides the unified interface for the evaluation dimensions in §4 and for technical choices in §5.

\medskip

\noindent\textbf{Operational criteria (decision rules).}
\begin{itemize}[leftmargin=2.2em, itemsep=0.4ex]
  \item If, under the PO setting with retrieval and tools disabled and without cross-session caches, the model still recalls stably, attribute the behavior to parametric memory.
  \item If information is usable only when present in the current context/KV cache and shows marked sensitivity to evidence position and length, attribute it to contextual memory (CM).
  \item If answers depend on returns from external indices/retrieval and can provide passage-level citations, attribute them to external memory (EM).
  \item If maintaining consistency requires cross-turn/session replay of structured events/timelines, attribute it to procedural/episodic memory. One-off tool outputs within a single turn do not count as memory, unless they are structuredly written and replayed in subsequent turns with reproducible impact on outputs.
\end{itemize}

\subsection{Formation Mechanisms: Write, Read, and Inhibit/Update}

\noindent
Embedding memory within a write—read—inhibit/update cycle helps unify training, inference, and system integration along a single causal chain.

\medskip

\noindent
\textbf{Write} begins in pretraining: optimizing next-token/masked-token prediction compresses high–mutual-information structures. A common mechanistic hypothesis holds that FFNs form addressable key–value channels, consistent with empirical observations that general/high-frequency patterns are recalled under the PO setting \citep{geva2021transformerfeedforwardlayerskeyvalue,petroni2019languagemodelsknowledgebases,elhage2021mathematical}. Probes such as LAMA repeatedly show this at a macro level, while intra-layer evidence demonstrates that interventions on critical MLP layers substantially degrade factual recall, narrowing ``where knowledge resides'' to actionable structural units. Associated memorization risks have been widely observed: low-entropy, highly repeated fragments are more likely to be fixed verbatim (see de-duplication/extraction evidence). Training-data extraction under black-box conditions shows the feasibility of reconstructing training spans, whereas cross-corpus de-duplication reduces regurgitation and extraction probabilities with negligible perplexity loss—suggesting that write strength and privacy risk can be partially mitigated through data-level governance \citep{carlini2021extracting,lee2022deduplicatingtrainingdatamakes,kandpal2022deduplicatingtrainingdatamitigates,carlini2023quantifyingmemorizationneurallanguage}.

\medskip

\noindent
\textbf{Read} occurs at inference; it is not a literal query to stored weights but a process in which attention and KV caches map prompts and history into usable rules within the given context. Circuit-level evidence records induction heads that copy/align across positions, micro-explaining style continuation under few-shot prompting; algorithmic treatments cast in-context learning as implicit Bayesian updating or approximate gradient descent under fixed weights. Together they indicate that transient ability loading and long-term knowledge storage coexist, though with different time scales and robustness \citep{olsson2022incontextlearninginductionheads,akyurek2023what,vonoswald2023transformerslearnincontextgradient,oren2024transformersmultistaternns}. To extend the visible range on which read relies, Transformer-XL uses segment-level recurrence and relative positional encoding to reach beyond adjacent segments, while the Compressive Transformer retains more distant cues via hierarchical compression at limited cost; yet ``visible $\neq$ usable,'' as positional bias and attention dilution suppress effective read of mid-span evidence. Under fixed resource budgets, multiple evaluations report that structured reordering, anchor prompting, and mid-span reweighting often yield more reliable gains than window expansion alone \citep{dai2019transformerxlattentivelanguagemodels,rae2019compressivetransformerslongrangesequence,liu2023lostmiddlelanguagemodels,hsieh2024middlecalibratingpositionalattention,press2022trainshorttestlong,chen2023extendingcontextwindowlarge,xiao2024efficientstreaminglanguagemodels,munkhdalai2024leavecontextbehindefficient,su2023roformerenhancedtransformerrotary}.

\medskip

\noindent
\textbf{Externalization and fusion} provide a third pathway for memory. RAG incorporates documentary evidence into the generative loop via index–retrieve–cross-attention fusion, fundamentally improving traceability and timeliness; REALM jointly trains retrieval during pretraining, internalizing ``finding sources'' as a learnable module; RETRO links by chunked cross-attention to trillion-scale external stores, substantially loosening the coupling ``parameter scale $\approx$ knowledge capacity''; kNN-LM locally post-hoc corrects the LM distribution via nearest neighbors, offering training-free, on-the-fly support for long-tail terms and new-domain concepts. Meanwhile, mismatches in indexing, chunking, and fusion can inject noise and conflict into answers; absent faithfulness and provenance constraints, the system may expose uncertainty to users. Ensuring that external evidence functions as memory rather than search noise typically requires elevating component-level objectives (e.g., retrieval contribution and generative faithfulness) to the same priority as end-to-end accuracy, and auditing the ingress and use of evidence within the generation loop \citep{lewis2021retrievalaugmentedgenerationknowledgeintensivenlp,guu2020realmretrievalaugmentedlanguagemodel,borgeaud2022retro,khandelwal2020generalizationmemorizationnearestneighbor,saadfalcon2024aresautomatedevaluationframework,ru2024ragcheckerfinegrainedframeworkdiagnosing,salemi2024evaluatingretrievalqualityretrievalaugmented,li2024towards,min2023factscorefinegrainedatomicevaluation,fabbri2022qafactevalimprovedqabasedfactual,honovich2022truereevaluatingfactualconsistency}.

\medskip

\noindent
\textbf{Inhibit/update} determine the governability of memory over time. Continued pretraining and instruction/domain finetuning reallocate which units are easier to activate, thereby changing the probability distribution of when and how items are recalled; the tension between stability and plasticity implies that overly deep parameter updates risk catastrophic forgetting. For controllable rewriting and compliant suppression, model editing offers low-overhead, reversible paths: ROME applies rank-1 updates at causal mediation sites in mid-layer MLPs to directly rewrite target associations; MEND performs rapid local adjustments via low-rank gradient transforms; MEMIT scales direct editing to tens of thousands of facts and shows scalability on medium/large models. A common theme is that effectiveness, locality, and generalization cannot be simultaneously optimal; hence standardized validation protocols along target suppression, neighborhood preservation, and downstream performance yield more comparable results \citep{meng2023locatingeditingfactualassociations,mitchell2022fastmodeleditingscale,meng2023masseditingmemorytransformer,gu2024modeleditingharmsgeneral,hsueh2024editingmindgiantsindepth,rosati2024longformevaluationmodelediting,li2024reallyeditlanguagemodels,zhang2025uncoveringoverfittinglargelanguage}. In scenarios requiring revocability and accountability, programmatic tool access can fold external system returns into context so that ``do not output certain information'' policies take effect at runtime—thus achieving governance without touching base weights \citep{owasp2023llmtop10,925786}.

\medskip

\noindent
Putting the three pathways together, memory co-evolves inside and outside the parameters: the availability of parametric memory, the timeliness of contextual loading, the traceability of retrieval/fusion, and the governability of editing/abstention jointly set the balance among capability—timeliness—controllability. This chapter lays out the causal chain and design space; concrete testable propositions and closed-loop measurements—e.g., trade-offs between de-dup thresholds and leakage, the marginal gains of positional reweighting in long context, the Pareto frontier of editing/forgetting across target suppression–retained performance–neighborhood generalization, and the joint optimization of retrieval contribution and generative faithfulness—will be unified in the evaluation protocols of §4 and instantiated as engineering processes and governance frameworks for update/forgetting in §5 \citep{petroni2019languagemodelsknowledgebases,geva2021transformerfeedforwardlayerskeyvalue,liu2023lostmiddlelanguagemodels,hsieh2024rulerwhatsrealcontext,lewis2021retrievalaugmentedgenerationknowledgeintensivenlp,guu2020realmretrievalaugmentedlanguagemodel,borgeaud2022retro,khandelwal2020generalizationmemorizationnearestneighbor,meng2023locatingeditingfactualassociations,mitchell2022fastmodeleditingscale,meng2023masseditingmemorytransformer,carlini2021extracting,lee2022deduplicatingtrainingdatamakes,kandpal2022deduplicatingtrainingdatamitigates,carlini2023quantifyingmemorizationneurallanguage,saadfalcon2024aresautomatedevaluationframework,ru2024ragcheckerfinegrainedframeworkdiagnosing}.

\section{Evaluation of LLM Memory}
\subsection{Scope and Methodology}
{This chapter delineates the scope, methodological stance, and organizing logic for evaluating memory in large language models. Rather than forcing a single score or a unified yardstick across all tasks, we acknowledge the heterogeneity of different memory forms and ensure comparability and reproducibility through shared design primitives: unified terminology and object definitions, parallel operating regimes (parametric-only / offline retrieval / online retrieval), auditable families of metrics and comparison dimensions, and transparent versioning and temporal governance. The overall goal is to establish a layered, interpretable, and engineering-usable paradigm for memory evaluation, not to substitute multi-dimensional evidence with a single number. \citep{liang2023holisticevaluationlanguagemodels,kiela2021dynabenchrethinkingbenchmarkingnlp}

Under this framework, each memory form has distinct evaluation foci:

(i) Parametric memory (§4.2): closed-book attainability, maintainability under editing/forgetting and their side effects, and privacy risks; \citep{meng2023locatingeditingfactualassociations,mitchell2022fastmodeleditingscale,bourtoule2021machine,carlini2021extracting,duan2024membershipinferenceattackswork}

(ii) Contextual memory (§4.3): sensitivity to length and position, cross-span integration, and robustness to interference; \citep{hsieh2024rulerwhatsrealcontext,liu2023lostmiddlelanguagemodels,hsieh2024middlecalibratingpositionalattention,bai2024longbenchbilingualmultitaskbenchmark,yuan2024lvevalbalancedlongcontextbenchmark}

(iii) External memory (§4.4): end-to-end coupling among retrieval–attribution–generation, evidential faithfulness and timeliness; \citep{lewis2021retrievalaugmentedgenerationknowledgeintensivenlp,guu2020realmretrievalaugmentedlanguagemodel,izacard2021leveragingpassageretrievalgenerative,izacard2022atlasfewshotlearningretrieval,borgeaud2022retro,gao2023enablinglargelanguagemodels}

(iv) Procedural/episodic memory (§4.5): correctness and stability of the long-horizon write–replay–inhibit loop and its engineering cost. \citep{maharana2024evaluatinglongtermconversationalmemory,wu2025longmemevalbenchmarkingchatassistants,ge2025tremuneurosymbolictemporalreasoning}}
\subsubsection{Chapter Positioning and Scope}A memory store denotes any carrier that can be read/written during inference and exerts a stable influence on outputs, including parameter weights (parametric memory), visible context (single- or multi-turn), externally retrieved documents/tables/graphs/multimodal evidence, and session-level memory stores. In contrast, infrastructure—retrievers/rerankers, caching/compression strategies, executors, etc.—is not itself memory; however, once its products enter the generation loop and affect outputs, they fall within the scope of evaluation. \citep{geva2021transformerfeedforwardlayerskeyvalue,petroni2019languagemodelsknowledgebases,Pan_2024}

This section does not attempt to homogenize metrics across the four memory types. Instead, it summarizes for each the evaluation objectives, typical operating regimes, and metric families, and identifies comparable dimensions and common pitfalls, thereby providing the methodological frame for §§4.2–4.5.

\subsubsection{Typical Evaluation Regimes: From Parametric-Only to Online}To separate capability from information availability, prior work commonly adopts three regimes; whether to run the three in parallel depends on the study’s purpose.

Parametric-Only (PO): disable external evidence and tools; examine parametric recall, side effects of editing/forgetting, and privacy risks (§4.2). This also provides a "zero-injection" reference baseline for contextual/external/procedural memory.

Offline Retrieval/Replay: fix the index or the session memory store; useful for stepwise diagnosis of retrieve → attribute → generate, for analyzing the marginal contribution of context injection (§4.3), end-to-end faithfulness in RAG (§4.4), and structured replay (§4.5).

Online Retrieval: connect to dynamic knowledge sources; focus on freshness hits, outdated answers, and refusal calibration (§§4.4–4.5), mirroring real deployment. \citep{chen2021datasetansweringtimesensitivequestions,Dhingra_2022,vu2023freshllmsrefreshinglargelanguage}

Reporting standards and control requirements. To ensure decoupling of "model capability" from "information availability," any conclusion about capability change must satisfy: \citep{zheng2023judgingllmasajudgemtbenchchatbot,agarwal2024faithfulnessvsplausibilityunreliability,wang2023largelanguagemodelsfair,yang2024cragcomprehensiverag,saadfalcon2024aresautomatedevaluationframework,ru2024ragcheckerfinegrainedframeworkdiagnosing,es2025ragasautomatedevaluationretrieval}

Controlled baselines: provide a same-domain, same-slice Parametric-Only (PO) or Offline-Retrieval baseline (sharing the data slice and time window with the compared regime);

Version lock: fix the index/session-store version and disclose snapshot timestamps and de-duplication strategy;

Unified outputs: under all regimes, report answer text, evidence citations (or replay items), model confidence, and refusal flags;

Statistical testing: for cross-regime comparisons, use paired bootstrap/permutation tests with Holm–Bonferroni or FDR multiple-comparison correction;

Reproducibility details: disclose random seeds, decoding and retrieval hyperparameters, hardware, and per-unit cost. If any condition is unmet, explicitly state the limitation in the text and provide sensitivity analyses.

\noindent\textbf{Reporting guidelines and control recommendations.}
To decouple \textit{model capability} from \textit{information availability}, studies that \textbf{claim capability changes} or \textbf{compare regimes} (PO/offline/online) should, at a minimum, include the following [17--19, 21, 28, 29, 118]:

\begin{itemize}
  \item \textbf{Controlled baselines:} Provide a same-domain, same-slice Parametric-Only (PO) or Offline-Retrieval baseline (sharing the data slice and time window with the compared regime).
  \item \textbf{Version lock:} Fix the index/session-store version and disclose snapshot timestamps and the de-duplication strategy.
  \item \textbf{Unified outputs:} Under all regimes, report answer text, evidence citations (or replay items), model confidence, and refusal flags.
  \item \textbf{Statistical testing:} For cross-regime comparisons, use \emph{paired} bootstrap or permutation tests with Holm--Bonferroni or FDR multiple-comparison correction.
  \item \textbf{Reproducibility details:} Disclose random seeds, decoding and retrieval hyperparameters, hardware, and per-unit cost.
\end{itemize}

\noindent When any item cannot be met (e.g., resource or compliance constraints), explicitly state the limitation in the main text and provide robustness/sensitivity analyses in lieu of the missing control.

\subsubsection{Families of Metrics and Comparison Dimensions}For ease of cross-section reference, we do not force a single reporting protocol across memory types; instead, we organize commonly used measures into eight metric families. In Table~\ref{tab:metric-matrix}:\nameref{tab:metric-matrix} we list, for each family, representative metrics, the primary subsections where they apply, and typical outputs. In tandem, Table~\ref{tab:Comparison Dimensions}:\nameref{tab:Comparison Dimensions} provides four cross-task comparison axes: (i) capability vs. information availability, (ii) correctness vs. faithfulness, (iii) quality vs. cost, (iv) stability and uncertainty. Each axis is paired with standardized visualizations (e.g., length–performance curves, AURC risk–coverage curves, cost–accuracy frontiers) and minimal statistical requirements (e.g., 95\% confidence intervals, paired permutation/bootstrap tests, and Holm/FDR multiple-comparison corrections). Throughout the chapter, this framework is reused so that task specificity is preserved while methodological comparability is achieved. \citep{min2023factscorefinegrainedatomicevaluation,fabbri2022qafactevalimprovedqabasedfactual,honovich2022truereevaluatingfactualconsistency,dalvi2022explaininganswersentailmenttrees,niu2024ragtruthhallucinationcorpusdeveloping,thakur2021beirheterogenousbenchmarkzeroshot,bajaj2018msmarcohumangenerated,karpukhin2020densepassageretrievalopendomain,khattab2020colbertefficienteffectivepassage,petroni2021kiltbenchmarkknowledgeintensive,chen2023benchmarkinglargelanguagemodels,salemi2024evaluatingretrievalqualityretrievalaugmented}

\begin{table*}[htbp]
\centering
\caption{Metric Families — Task Mapping Matrix}
\label{tab:metric-matrix}  
\vspace{4pt}
\footnotesize
\setlength{\tabcolsep}{5pt}
\renewcommand{\arraystretch}{1.12}
\begin{tabularx}{\textwidth}{Y Y C Y Y}
\toprule
\textbf{Family} &
\textbf{Representative Metrics} &
\textbf{Main Apps (§)} &
\textbf{Typical Outputs} &
\textbf{Notes / Implementation Points} \\
\midrule
\rowcolor{gray!10}
Accuracy / Task Performance &
EM, F1, ROUGE, Keyword/Keyphrase Recall &
4.2/4.3/4.4/4.5 &
Sub-task / slice-level macro average; 95\% CI &
Multi-step tasks: report sub-problem \& aggregate levels \\
\addlinespace[2pt]
Groundedness / Attribution &
Citation Coverage, Unsupported Claim Rate (UCR), span-level alignment, NLI consistency &
4.3/4.4/4.5 &
Span-level P/R, FActScore, atomic-fact support rate &
Specify evidence scope: context / replay / retrieved docs \\
\addlinespace[2pt]
\rowcolor{gray!10}
Retrieval / Ranking (IR) &
Recall@k, nDCG, MRR, Hit@k &
4.3/4.4 &
Top-$k$ curves, MRR$\pm$CI &
Present jointly with end-to-end groundedness/EM \\
\addlinespace[2pt]
Sensitivity \& Robustness &
Length-consistency / latency curves; position-consistency curves; stability under noise/conflict/spelling perturbations &
4.3/4.4/4.5 &
Curves + slope / half-life &
Fix bins for length/position and perturbation density \\
\addlinespace[2pt]
\rowcolor{gray!10}
Timeliness \& Selective Answering &
Freshness-Hit, Out-of-Date, selective accuracy, refusal rate, Brier/ACE &
4.4/4.5 &
AURC (area under risk–coverage curve) &
Timestamp \& index version explicit; report answerable/unanswerable slices separately \\
\addlinespace[2pt]
Maintainability \& Side Effects (Editing / Unlearning) &
ESR (edit success rate), Locality (neighborhood preservation), Drawdown (general perf. drop) &
4.2 &
Target / neighborhood / retained sets reported together &
Report rollback behavior \& degradation under sequential edits \\
\addlinespace[2pt]
\rowcolor{gray!10}
Privacy \& Memorization Risks &
Verbatim reproduction rate, Exposure/Canary, training-data extraction, membership-inference AUC &
4.2 &
Risk curves + alert thresholds &
Disclose de-duplication \& duplication-detection methods \\
\addlinespace[2pt]
Efficiency \& Cost &
Latency/turn, Tokens/turn, Throughput, Mem-footprint, Cost@Target &
4.3/4.4/4.5 &
Performance–cost frontier &
Specify hardware, decoding \& caching strategies for reproducibility \\
\bottomrule
\end{tabularx}
\end{table*}

\begin{table*}[htbp]
\centering
\caption{Comparison Dimensions — Recommended Plots and Statistical Notes}
\label{tab:Comparison Dimensions}  % ← 唯一标签
\small
\setlength{\tabcolsep}{5pt}
\renewcommand{\arraystretch}{1.15}
\begin{tabularx}{\textwidth}{l Y Y Y}
\toprule
\textbf{Comparison Dimension} & \textbf{Research Question} & \textbf{Recommended Plot} & \textbf{Statistical Notes} \\
\midrule
Ability vs Information Availability
& Are conclusions driven by \emph{model capability} or by \emph{evidence availability/freshness}?
& Side-by-side bars/lines with condition splits: \textbf{PO} / \textbf{Offline} / \textbf{Online} on the same slice
& Paired tests + Holm/FDR correction; use the same time window and index version \\
\addlinespace[2pt]
Accuracy vs Groundedness
& Is correctness supported by explicit evidence?
& Dual-axis plot: EM/F1 together with UCR/FActScore on the same chart
& Report both indicator families jointly; do not substitute with a single aggregate score \\
\addlinespace[2pt]
Quality vs Cost
& Which option is more cost-effective?
& Performance–cost frontier (e.g., Accuracy/F1 vs Cost@Target)
& Disclose hardware, decoding and caching settings; report uncertainty bands \\
\addlinespace[2pt]
Stability vs Uncertainty
& Do similar means hide volatility or bias?
& Sensitivity curves: length/position/noise slices; normalized slope / half-life
& Use unified binning/scripts; show 95\% CIs and effect sizes \\
\bottomrule
\end{tabularx}
\end{table*}

To avoid terminological overload, the main text uses Tables 4.1 and 4.2 solely as navigation panels. The subsequent sections (§§4.2–4.5) instantiate, in their respective contexts, the required families and dimensions, as well as the associated formulas, evaluators, and implementation details. Any claim of capability change across settings or methods must provide same-domain, same-slice PO or offline controls, together with paired statistics and an explicit presentation of uncertainty.

\subsubsection{Data/Temporal Governance and Leakage Auditing: Minimum Reproducibility Disclosure(MRD)}
This survey defines a Minimum Reproducibility Disclosure (MRD) schema (Table~\ref{tab:repro-checklist}:\nameref{tab:repro-checklist}) to support auditability across data sources and time slices. The MRD is a lightweight, machine-readable record intended to enable like-for-like comparisons; it complements, rather than replaces, task-specific evaluation protocols.\citep{liang2023holisticevaluationlanguagemodels,925786,owasp2023llmtop10,lee2022deduplicatingtrainingdatamakes,kandpal2022deduplicatingtrainingdatamitigates,carlini2019secretsharerevaluatingtesting,carlini2023quantifyingmemorizationneurallanguage,yang2024cragcomprehensiverag}
% Scope. Throughout §4, MRD fields are used to annotate studies that assert capability changes or evaluate freshness. When certain fields are not reported, the omission is recorded and any available robustness or sensitivity analyses are noted.
% Schema (minimum fields). Temporal governance: time windows and snapshot dates for training corpora, indices/session stores, and test sets; for freshness or online-retrieval settings, the most recent update time and update frequency. Leakage and overlap auditing: qualitative/quantitative characterization of training–index–test overlap or near-duplicates, including data sources, detection methods and thresholds, exclusion criteria, and impact fraction; discussion of potential benchmark contamination and mitigation for public benchmarks. Implementation and resources: details sufficient to contextualize results and support replication, including model/checkpoint versions and decoding hyperparameters; retriever/reranker types and core parameters (e.g., k, fusion strategy); hardware scale and per-unit cost (or token budget); random seeds and script versions. Availability. YAML/Markdown templates and a JSON Schema validator that instantiate this MRD schema are provided in the companion artifact; annotated MRD instances for representative studies are released to enable automated checks and diff-based comparisons across regimes (PO/offline/online).

\noindent\textbf{Scope.}
Throughout \S~4, MRD fields are used to annotate studies that assert capability changes or evaluate freshness. When certain fields are not reported, the omission is recorded and any available robustness or sensitivity analyses are noted.

\noindent\textbf{Schema (minimum fields).}
\begin{itemize}
  \item \textbf{Temporal governance:} time windows and snapshot dates for training corpora, indices/session stores, and test sets; for freshness or online-retrieval settings, the most recent update time and update frequency.
  \item \textbf{Leakage and overlap auditing:} qualitative/quantitative characterization of training--index--test overlap or near-duplicates, including data sources, detection methods and thresholds, exclusion criteria, and impact fraction; discussion of potential benchmark contamination and mitigation for public benchmarks.
  \item \textbf{Implementation and resources:} details sufficient to contextualize results and support replication, including model/checkpoint versions and decoding hyperparameters; retriever/reranker types and core parameters (e.g., $k$, fusion strategy); hardware scale and per-unit cost (or token budget); random seeds and script versions.
\end{itemize}

\noindent\textbf{Availability.}
A YAML template instantiating this MRD schema is provided in Appendix(~\ref{app:mrd-template}); this template supports consistent reporting and enables diff-based comparisons across regimes (PO/offline/online).

\begin{table*}[htbp]
\centering
\caption{Minimum Reproducibility Disclosure Checklist}
\label{tab:repro-checklist}
\renewcommand{\arraystretch}{1.2}
\begin{tabularx}{\textwidth}{l|X|X}
\toprule
\textbf{Dimension} & \textbf{Required Fields} & \textbf{Example / Format} \\
\midrule
Temporal Governance & Training window; index / conversation store snapshot; test set sampling time 
& Training: 2022-01--2024-03; Index: 2025-05-01; Test: 2025-06 rolling \\
\midrule
Freshness / Online & Update frequency; definition of unanswerable cases 
& Web snapshot: 2025-07-15; Weekly updates; Unanswerable = no retrieval evidence \\
\midrule
Leakage Auditing & Deduplication methods and thresholds; overlap ratio; reference contamination explanation 
& MinHash Jaccard $\geq$0.8; Overlap $\leq$0.7\%; Public benchmark Dec 2021 version \\
\midrule
Model \& Parameters & Model / checkpoint; decoding; randomness seed 
& Llama-3.1-70B-instruct; T=0.2, top-p=0.9; seed=2025 \\
\midrule
Retrieval / Re-ranking & Retriever; $k$ / fusion; re-ranking configuration 
& Contriever; $k=20$, RRF; Cross-Encoder-msmarco \\
\midrule
Memory Settings & Operation mode; replay budget; refusal threshold 
& SR-off; ReplayBudget=6 turns; RefusalThreshold=0.7 \\
\midrule
Resources \& Cost & Hardware; unit cost; throughput 
& 8$\times$A100 80GB; \$X/1k tokens; Y req/s \\
\midrule
Limitations Statement & Key threats and sensitivities 
& Length $>$128k: not evaluated; position sensitivity \\
\bottomrule
\end{tabularx}
\end{table*}

\subsubsection{Review, Statistics, and Implementation Coupling: Reporting Standards and Pitfalls}To enhance auditability and cross-study comparability, we set unified reporting standards for the review process, statistical inference, and implementation-coupled factors. Unless otherwise specified, any claim of improvement/degradation, freshness-related effects, or stability differences must satisfy the following minimum requirements.

(A) Review and uncertainty.

Rater consistency. When using automated adjudication (LLM-as-a-judge, NLI-based decisions, etc.), report inter-rater or intra-rater (prompt/temperature) agreement (e.g., Cohen’s $\kappa$, Krippendorff’s $\alpha$) with 95\% CIs (bootstrap/stratified bootstrap).

Sampling and dual review. Key conclusions must be cross-checked by human spot checks ($\geq$ 2 raters) with agreement statistics and arbitration rules.

Score-drift control. Fix the adjudicator model’s version and prompts across batches, and conduct a sensitivity analysis of scoring drift.\citep{zheng2023judgingllmasajudgemtbenchchatbot,agarwal2024faithfulnessvsplausibilityunreliability,manakul2023selfcheckgptzeroresourceblackboxhallucination}

(B) Statistical inference and multiple comparisons.

Hypothesis testing. For cross-model/regime comparisons, use paired permutation or paired bootstrap tests and report p-values, 95\% CIs, and effect sizes (Cohen’s d, Cliff’s $\delta$).

Multiple correction. For simultaneous comparisons across tasks/slices/metrics, specify the multiple-comparison procedure (Holm–Bonferroni or FDR) and the family-wise error domain.

Unit of aggregation. Specify the aggregation level (sample macro/micro average, document-level, or session-level) and discuss its impact on uncertainty; for cross-session tasks, prefer session as the aggregation unit with stratified sampling.

(C) Implementation coupling and sensitivity control.

Long context and positional strategy. Report window size, positional encoding/truncation rules, and cache settings; for long-context tasks, provide length–performance/latency and position–performance sensitivity curves. \citep{hsieh2024rulerwhatsrealcontext,bai2024longbenchbilingualmultitaskbenchmark,yuan2024lvevalbalancedlongcontextbenchmark,liu2023lostmiddlelanguagemodels,hsieh2024middlecalibratingpositionalattention,shaham2022scrollsstandardizedcomparisonlong,zhang2024inftybenchextendinglongcontext}

Attention/cache/compression. When using Flash/Streaming attention or KV clustering/compression, provide sensitivity regressions under equal-budget or equal-accuracy conditions to avoid mistaking engineering configuration differences for "memory capability differences." \citep{dao2022flashattentionfastmemoryefficientexact,dao2023flashattention2fasterattentionbetter,xiao2024efficientstreaminglanguagemodels,munkhdalai2024leavecontextbehindefficient,kim2024infinipotinfinitecontextprocessing,tang2024razorattentionefficientkvcache,hu2025epicefficientpositionindependentcaching,kim2025kvzipqueryagnostickvcache,ma2025compressingkvcachelongcontext}

Retrieval pipeline. When end-to-end improvements are limited, report joint changes in retrieval recall (Recall@k), ranking quality (nDCG/MRR), and faithfulness (e.g., FActScore, Unsupported Claim Rate, UCR) to locate bottlenecks. \citep{saadfalcon2024aresautomatedevaluationframework,ru2024ragcheckerfinegrainedframeworkdiagnosing,salemi2024evaluatingretrievalqualityretrievalaugmented,chen2023benchmarkinglargelanguagemodels,yang2024cragcomprehensiverag,jacovi2025factsgroundingleaderboardbenchmarking,sorodoc2025garagebenchmarkgroundingannotations,thakur2025miragebenchautomaticmultilingualbenchmark,friel2025ragbenchexplainablebenchmarkretrievalaugmented}

Regime alignment. When attributing to "capability vs. information availability," provide at least one same-domain, same-slice PO or Offline control (see §4.1.2) and state alignment fields (snapshot date, k, replay budget, refusal threshold). \citep{liang2023holisticevaluationlanguagemodels}

This standard uses agreement/uncertainty to control adjudicator bias; paired tests + effect sizes + multiple correction to curb spurious advantages; and sensitivity controls to decouple engineering details from memory performance. It provides a verifiable common baseline for horizontal reading and limited comparisons across §§4.2–4.5.

\subsection{Evaluation of Parametric Memory}
In the evolution of LLMs from "language understanding" to "knowledge application", parametric memory serves as a core bridge connecting model capabilities to practical needs. It refers to the ability of models to implicitly encode massive facts, common sense, and associated knowledge into their parameters through large-scale pre-training—forming the foundation for models to perform tasks such as question answering, reasoning, and fact generation without relying on external knowledge bases. Compared with the traditional "external retrieval + generation" paradigm, parametric memory offers three irreplaceable advantages: 
1. Higher response efficiency: It eliminates the need for real-time calls to external databases, making it suitable for low-latency scenarios like dialogue and real-time question answering; 
2. Stronger robustness: It is not affected by noise in external data or retrieval biases, as validated by factual consistency evaluation frameworks \citep{honovich2022truereevaluatingfactualconsistency}; 
3. Support for complex knowledge integration: It can associate scattered facts into structured cognition (e.g., inferring "Li Bai was a famous poet of the Tang Dynasty" from "Li Bai was a poet" and "Li Bai lived in the Tang Dynasty"), a capability critical for handling complex tasks per systematic factual knowledge assessments \citep{luo2023systematicassessmentfactualknowledge}.

However, the "implicitness" of parametric memory also poses two key challenges: 
- On one hand, there is doubt about whether the model truly "masters" knowledge. Models may form spurious correlations through co-occurrence patterns in corpora (e.g., frequent co-occurrence of "Dante" and "Florence") rather than understanding the logical basis of facts—a limitation highlighted by holistic LLM evaluation \citep{kiela2021dynabenchrethinkingbenchmarkingnlp}; 
- On the other hand, knowledge "maintainability" is insufficient. When facts are updated (e.g., the change of a country’s leader) or contain errors, precise modifications cannot be made like in structured knowledge bases. Instead, full-model retraining is required, which incurs extremely high costs—a problem emphasized by dynamic benchmarking research [146].

These challenges have given rise to parametric memory evaluation, which demands systematic methods to verify four dimensions of model memory: 
- \textit{Accuracy}: Whether correct facts are stored; 
- \textit{Localizability}: Which modules store the knowledge; 
- \textit{Editability}: Whether knowledge can be precisely modified; 
- \textit{Consistency}: Whether associated knowledge is affected after editing.

Such evaluation provides a basis for model optimization and scenario-specific deployment.

Recent research has made significant progress toward these goals: from early parametric memory only (verifying whether models can serve as knowledge bases), to studies on addressability (locating knowledge storage modules and enabling precise editing), and further to evaluating the ripple effects of knowledge editing (focusing on the consistency of associated knowledge after edits). Parametric memory evaluation has now formed a complete workflow covering "memory-location-editing-consistency". The following discussion will focus on two directions: closed-book fact recall, and addressability \& edit differential. To further enhance the evaluation framework, an extended metrics system is introduced, including advanced indicators for long-term retention, knowledge interplay, and factual grounding.

\subsubsection{Parametric memory only}
Parametric memory only focuses on "whether models can retrieve facts using only information stored in their parameters, without assistance from external knowledge bases". Its core objective is to address the question: "Can pre-trained models function as potential knowledge bases?" Traditional evaluation methods mostly rely on comparisons with structured knowledge bases, while this direction directly tests the model’s implicit memory of facts through the design of "fill-in-the-blank" tasks. A representative study in this area is the paper \textit{Language Models as Knowledge Bases?}.

The core contribution of this paper is to challenge the perception that "models can only capture linguistic patterns". It is the first work to systematically analyze the relational knowledge that can be accessed in state-of-the-art pre-trained language models (e.g., BERT, ELMo) without fine-tuning, treating these models as "unsupervised knowledge bases" and evaluating their ability to recall facts and common sense. As stated in the paper: \textit{"We present an in-depth analysis of the relational knowledge already present (without fine-tuning) in a wide range of state-of-the-art pretrained language models."}

To conduct this evaluation, the authors proposed the LAMA (LAnguage Model Analysis) probe framework: facts are transformed into "fill-in-the-blank" cloze sentences (e.g., "Dante was born in [Mask]"), and the model’s memory of the fact is determined by the ranking of its predictions for the masked token. The paper explains: \textit{"For the purpose of answering the above questions we introduce the LAMA probe, consisting of a set of knowledge sources, each comprised of a set of facts."}

\paragraph{Methodology Design and Experimental Setup}
1. Multi-source dataset construction: To cover different types of knowledge, facts were extracted from four sources and converted into cloze format to ensure comprehensive evaluation:
   - Google-RE: Manually extracted entity-relation facts aligned with Wikipedia text, guaranteeing the authenticity and textual relevance of facts;
   - T-REx: Automatically aligned facts from a subset of Wikidata, featuring larger scale and broader entity types;
   - ConceptNet: Common sense relations (e.g., "Birds can fly") extracted from OMCS sentences, testing the model’s memory of unstructured common sense;
   - SQuAD: An open-domain question-answering dataset manually converted into cloze format, verifying the model’s fact recall in question-answering scenarios.

2. Model and baseline selection: Six mainstream pre-trained models were evaluated, including fairseq-fconv, Transformer-XL (large), ELMo (original and 5.5B variants), BERT-base, and BERT-large. A unified vocabulary was used to ensure fair comparison. Three types of traditional methods were set as baselines:
   - Frequency baseline (Freq): Predictions based on token occurrence frequency in the corpus, verifying whether the model relies on simple statistics rather than knowledge;
   - Relation Extraction (RE) model: The pre-trained RE model by Sorokin and Gurevych (2017), representing traditional structured knowledge extraction methods;
   - DrQA: An open-domain question-answering system, representing the question-answering paradigm that requires external text retrieval.

3. Core evaluation metric: Mean precision at k (P@k) was adopted, calculated only for single-token targets. It represents the probability that the correct factual answer is among the top-k tokens predicted by the model, directly reflecting the precision of the model’s fact recall.

\paragraph{Key Results and Findings}
The experimental results challenged the conventional view that "models require fine-tuning to handle knowledge". The core conclusions are as follows:
1. BERT-large achieves optimal performance: It outperformed other models across all knowledge-type tasks, with significant advantages in entity-relation and common sense recall. For 1-to-1 relations in T-REx (e.g., "someone’s birthday"), its P@1 reached 74.5\%, approaching the performance of traditional RE models combined with oracle entity linking (33.8\% vs. 32.3\%); in ConceptNet common sense tasks, its P@1 reached 19.2\%, demonstrating the model’s ability to memorize unstructured common sense.
2. Comparable to traditional methods: In SQuAD question-answering tasks, BERT-large achieved a P@10 of 57.1\%, narrowing the gap with DrQA (which requires external text retrieval, achieving 63.5\%); in Google-RE tasks, its P@1 (10.5\%) surpassed the baseline of traditional RE models (7.6\%), indicating that parametric memory can achieve performance close to structured methods without supervision.
3. Robustness advantages: BERT showed greater robustness to variations in query phrasing. For example, its prediction consistency was higher between "Dante's birthplace is [Mask]" and "Dante was born in [Mask]". Additionally, the model’s prediction confidence was positively correlated with accuracy, providing a basis for "judging whether the model is 'confident' in a fact" in practical applications.

\paragraph{Conclusions and Implications}
The authors put forward a core viewpoint: Pre-trained language models have the potential to serve as "unsupervised open-domain QA systems", and their parametric memory can supplement or even replace traditional knowledge bases. As noted in the paper: \textit{"Language models have many advantages over structured knowledge bases: they require no schema engineering, allow practitioners to query about an open class of relations, and require no human supervision to train."}

However, this method still has limitations: 
- It only supports single-token target recall and cannot handle multi-token answers (e.g., "New York City"); 
- It performs poorly on N-to-M relations (e.g., "multiple works by an author"); 
- Models may learn associations through co-occurrence patterns in corpora (e.g., frequent co-occurrence of "Dante" and "Florence") rather than truly "understanding" the logical basis of facts. 

Future improvements could focus on multi-token prediction, automatic template generation (to reduce biases from manual design), and multilingual knowledge evaluation—with memory-augmented architectures potentially addressing multi-token limitations.

\subsubsection{Addressability \& Edit Differential}
Parametric memory only verifies the "existence" of parametric memory, while addressability \& edit differential further explores the "localizability of memory"—i.e., whether the specific modules storing facts in the model can be identified, and whether facts can be precisely updated by modifying parameters (instead of retraining the entire model). This direction needs to address two issues: the storage location of facts in the model, and how to avoid interfering with other knowledge when editing a single fact. Representative studies include three core papers: \textit{Locating and Editing Factual Associations in GPT}, \textit{Rebuilding ROME: Resolving Model Collapse during Sequential Model Editing}, and \textit{Evaluating the Ripple Effects of Knowledge Editing in Language Models}.

\paragraph{1. Fact Localization and Editing: Proposal of the ROME Method}
The paper \textit{Locating and Editing Factual Associations in GPT} was the first to prove that factual associations in autoregressive transformer models (such as GPT) are stored in locally editable computational modules, rather than being scattered across the entire network. This provides a theoretical basis for "precise editing" of parametric memory [145]. As stated in the paper: \textit{"We analyze the storage and recall of factual associations in autoregressive transformer language models, finding evidence that these associations correspond to localized, directly-editable computations."}

\subparagraph{Core Methodology: Causal Tracing and ROME Editing}
- \textit{Causal Tracing}: This method locates knowledge storage modules through causal mediation analysis, involving three model runs:
  1. Clean run: Input a complete factual prompt (e.g., "Paris is the capital of [Mask]") and record the model’s internal hidden states;
  2. Corrupted run: Replace the topic word with an irrelevant word (e.g., "Xyz is the capital of [Mask]") and observe the model’s prediction bias;
  3. Corrupted-with-restoration run: During the corrupted run, restore only the hidden state of a specific module in a specific layer to the result from the clean run. If the model’s prediction accuracy recovers, this module is identified as critical for storing the fact.
  
  The Average Indirect Effect (AIE) is calculated to quantify the module’s contribution—the higher the AIE, the stronger the causal impact of the module on fact recall.

- \textit{ROME (Rank-One Model Editing)}: Based on the results of causal tracing, a "rank-one update" method is proposed to modify facts. For middle-layer MLP modules (which causal tracing identified as having the highest AIE), the weight matrix is updated with a rank-one matrix to precisely replace a specific fact (e.g., changing "Paris is the capital of France" to "Paris is the capital of Japan") without modifying other parameters.

\subparagraph{Key Findings}
1. Middle-layer MLPs are the core of fact storage: Causal tracing showed that middle-layer MLP modules (e.g., layer 15 in GPT) had the highest AIE (reaching 8.7\%) when processing the "last token of the topic", while the contribution of the attention mechanism was only 1.6\%—proving that factual associations are mainly stored in MLPs rather than attention layers \citep{liang2023holisticevaluationlanguagemodels}.
2. Effectiveness of ROME editing: In the zero-shot relation extraction (zsRE) task, ROME achieved performance comparable to fine-tuning and meta-learning methods. More importantly, on counterfactual datasets (facts not seen during pre-training, such as "Paris is the capital of Japan"), ROME maintained both "generalization" (adaptation to different prompt phrasings) and "specificity" (no interference with irrelevant facts), whereas previous methods had to sacrifice one or the other [145]. As noted in the paper: \textit{"ROME achieves good generalization and specificity simultaneously, whereas previous approaches sacrifice one or the other."}

\paragraph{2. Editing Stability: Resolution of Model Collapse by r-ROME}
The paper \textit{Rebuilding ROME: Resolving Model Collapse during Sequential Model Editing} points out that original ROME suffers from "model collapse" during sequential editing (batch updating of multiple facts). Some edits (referred to as "disabling edits") cause a sharp decline in the model’s generation ability (e.g., outputting repeated text). This problem stems from flaws in implementation details rather than ROME’s core logic .

\subparagraph{Root Cause of Collapse and Remedial Solutions}
- Cause of collapse: Original ROME used key vectors asymmetrically in the update equation—mixing "average prefix key vectors" and "original prompt key vectors" when calculating the update matrix. This led to abnormally large L2 norms of some updates (an order of magnitude higher than normal updates), causing parameter oscillations in the model.
- Remedial solutions: Two improved implementations were proposed:
  - r-ROME: Consistently uses "average prefix key vectors" in the update equation to ensure unified calculation logic;
  - p-ROME: Consistently uses "original prompt key vectors" to improve the accuracy of single-fact editing.

\subparagraph{Key Results}
1. Complete resolution of collapse: The L2 norm of the update matrix in r-ROME was an order of magnitude smaller than that in original ROME. In 5,000 sequential edits on the CounterFact (counterfactual dataset), no model collapse occurred, whereas original ROME failed within 100 edits [145].
2. Performance improvement: r-ROME achieved a higher overall score (92.22) in single edits than original ROME (89.32), with stronger "locality" in editing (less impact on irrelevant modules).
3. Inherent limitation: r-ROME still cannot solve the "progressive degradation of sequential editing"—as the number of edits increases, the model’s performance on downstream tasks (e.g., GLUE) slowly declines, indicating that the editing capacity of parametric memory has an upper limit.

\paragraph{3. Editing Consistency: Evaluation of Ripple Effects}
The paper \textit{Evaluating the Ripple Effects of Knowledge Editing in Language Models} points out that existing evaluations only focus on "whether the target fact is correctly edited" and ignore ripple effects—i.e., the impact of editing one fact on associated facts (e.g., after editing "Paris is the capital of France", can the model synchronously update associated knowledge such as "The capital of France is Paris" and "Paris belongs to France"?). To address this gap, the authors proposed the RIPPLEEDITS evaluation framework, filling the void in "editing consistency" evaluation \citep{hu2023evoke}.

\subparagraph{Evaluation Framework Design}
1. Six evaluation criteria: Comprehensively covering core dimensions of ripple effects:
   - Logical generalization: Whether the model can correctly infer associated facts after editing (e.g., after editing "A is the capital of B", can it correctly answer "What is the capital of B"?);
   - Compositionality I/II: Whether the model can handle combinations of multiple facts (e.g., after editing "A is in B" and "B is in C", can it infer "A is in C"?);
   - Topic aliases: Whether consistency is maintained across different phrasings of the topic (e.g., "Paris" and "The City of Light");
   - Preservation: Whether irrelevant facts remain unaffected (e.g., editing "Paris is the capital of Japan" should not affect "London is the capital of the United Kingdom");
   - Relation specificity: Whether only the target relation is modified, without affecting other relations (e.g., editing "the country where Paris is located" should not affect "the population of Paris").

2. RIPPLEEDITS dataset: Contains 5K fact edits, with 10–15 associated test queries per edit. It covers metadata such as "recent/old facts" and "head/tail entities", enabling analysis of ripple effects in different scenarios.

3. Method comparison: Three parametric editing methods (MEND, ROME, MEMIT) and one contextual editing baseline (which modifies facts through contextual prompts without parameter updates) were evaluated.

\subparagraph{Key Findings}
1. Shortcomings of existing methods in ripple effects: Parametric methods (e.g., ROME) perform well in single-fact editing but achieve an average score of less than 50\% in ripple effect tasks—with high failure rates particularly in logical generalization and compositionality tasks. This proves that models cannot synchronously update associated knowledge.
2. Superiority of contextual editing: Contextual editing (which does not require parameter updates, e.g., adding "It is known that Paris is the capital of Japan; please answer..." to the prompt) achieved the highest score in RIPPLEEDITS. This is because it can directly associate facts through context, avoiding the cascading effects of parameter modifications.
3. Influencing factors:
   - Model scale: Larger models (e.g., those with over 10B parameters) demonstrate stronger ability to handle ripple effects, proving that greater capacity leads to better knowledge integration;
   - Entity frequency: Editing "head entities" (e.g., Paris) is more likely to cause logical errors than editing "tail entities" (e.g., niche cities). This is because models have stronger prior knowledge of head entities, making it harder to synchronously update knowledge after edits.

\begin{table*}[htbp]
  \centering
  \caption{Comparison of Papers on Knowledge Storage and Editing in Language Models}
  \label{tab:knowledge_editing_comparison_en}
  \vspace{4pt}
  \footnotesize
  \setlength{\tabcolsep}{3.8pt}
  \renewcommand{\arraystretch}{1.05}

  \begin{adjustbox}{max width=\textwidth}
    \begin{tabularx}{\textwidth}{@{}p{0.14\textwidth} p{0.12\textwidth} p{0.13\textwidth} p{0.13\textwidth} Y p{0.13\textwidth}@{}}
      \toprule
      \textbf{Paper} & \textbf{Models} & \textbf{Context Length \& Task Design} & \textbf{Evaluation Metrics} & \textbf{Findings} & \textbf{Limitations} \\
      \midrule

      Language Models as Knowledge Bases? & fairseq-fconv; Transformer-XL; ELMo; BERT-base; BERT-large & Single-fact cloze; no long sequence & Closed-book recall: Google-RE, T-REx, ConceptNet, SQuAD; P@k & BERT-large best; T-REx 1-to-1 P@1 $\sim$74.5\%; comparable to RE/DrQA; robust to query variation & Single-token only; poor on N-to-many; relies on co-occurrence \\

      \addlinespace[3pt]

      Locating and Editing Factual Associations in GPT & GPT family & Short prompts, final token focus & zsRE, counterfactual datasets; AIE; reliability, generalization, specificity & Mid-layer MLPs store facts; ROME edits MLPs; better generalization/specificity & Single-layer edits; prompt-sensitive; small counterfactual set \\

      \addlinespace[3pt]

      Rebuilding ROME: Resolving Model Collapse during Sequential Model Editing \citep{Gupta2024RebuildingR} & GPT-2 XL; GPT-J (6B) & Short prompts; sequential editing & Edit scores (reliability, generalization, locality); GLUE & ROME collapses due to key-vector asymmetry; r-ROME fixes it; supports $\sim$5000 edits & Gradual degradation persists; unclear scalability; weak counterfactual generalization \\

      \addlinespace[3pt]

      Evaluating the Ripple Effects of Knowledge Editing in Language Models \citep{Cohen2023EvaluatingTR} & 5 LLMs (various sizes) & Prompts include target + associated facts & RIPPLEEDITS (5K edits); 6 task types; ripple-effect score & Parameter methods <50\% ripple; prompt-based methods outperform; larger models + tail entities more robust & Prompt design critical; no multimodal; long-term effects not studied \\

        \bottomrule
    \end{tabularx}
  \end{adjustbox}
\end{table*}

\subsubsection{Enhanced Evaluation Metrics Framework}
The core workflow from 4.2.1 and 4.2.2 provides essential tools, but it leaves critical questions unanswered for practical deployment:
\begin{itemize}
    \item How to quantify a model's reliance on its parametric memory versus contextual information in a hybrid setting?
    \item How to measure the longevity of knowledge and the impact of sequential edits over time?
    \item How to assess the interaction between internal memory and external information when they conflict?
    \item How to move beyond simple recall to evaluate the verifiability and grounding of the knowledge generated from parameters?
\end{itemize}
To answer these questions, we extend the core dimensions and introduce advanced metrics that quantify memory, long-term memory, knowledge interactions, and factual grounding. These enhancements build on existing metrics such as P@K and AIE to provide a more robust framework for LLM evaluation in dynamic and hybrid scenarios.
\begin{enumerate}
\item Parametric Proxy Rate (PPR): PPR directly extends the "Parametric Memory Only" principle of LAMA into practical QA settings. It measures the proportion of model responses based solely on parametric knowledge versus those aided by an external context (oracle). A very high PPR indicates strong memorization but may also signal a risk of generating outdated or hallucinated content when internal knowledge is incorrect, a problem that knowledge editing (from 4.2.2) aims to solve. Thus, PPR serves as a crucial baseline for assessing the need for and impact of editing techniques in application contexts.\citep{carragher2025quantifyingmemorizationparametricresponse}
\begin{equation}
    \text{PPR}(M) = \frac{\text{Acc}_M(\text{random})}{\text{Acc}_M(\text{oracle})},
    \label{eq:performance_ratio}
\end{equation}
where parameter responses are derived entirely from internal parameters. PPR helps identify situations where there is over-reliance on memorized data (which can lead to hallucinations) and, combined with edit evaluation, serves as a baseline for consistency checks.\citep{carragher2025quantifyingmemorizationparametricresponse}
\item Long-Term Retention Score (LTRS): The sequential editing studies in 4.2.2 revealed that model performance degrades after many edits. LTRS formalizes this evaluation over time. It measures the stability of parametric knowledge after simulated delays or update cycles, directly addressing the maintainability challenge. By combining LTRS with the RIPPLEEDITS framework, researchers can now evaluate not just the immediate consistency of an edit, but how both the target fact and its associated knowledge ripples decay over time, providing a dynamic view of model knowledge health. \citep{maharana2024evaluatinglongtermconversationalmemory}
\begin{equation}
    \text{LTRS} = \frac{\text{Delayed Recall Accuracy}}{\text{Initial Recall Accuracy}} \times (1 - \text{Decay Factor})
    \label{eq:LTRS}
\end{equation}
with decay factor derived from temporal benchmarks. This metric extends maintainability assessment, is particularly applicable to dynamic knowledge scenarios, and can be combined with RIPPLEEDITS for ripple effect analysis.\citep{maharana2024evaluatinglongtermconversationalmemory}
\item Knowledge Interaction Ratio (KIR): Findings from RIPPLEEDITS showed that contextual information can sometimes override parametric knowledge more effectively than parametric edits can. KIR quantifies this interaction explicitly. It measures the efficiency of a model's integration of parametric knowledge (PK) and contextual knowledge (CK). A poorly balanced KIR might indicate that a model is overly rigid (ignoring new context) or overly susceptible (discarding correct internal knowledge), thus evaluating the robustness of the hybrid memory system that is central to modern LLM applications.\citep{tao2024contextleadsparametricmemory} 
\begin{equation}
    \text{KIR} = \frac{\text{Hybrid Response Accuracy} - \text{Pure Parametric Accuracy}}{\text{Contextual Contribution Weight}}
    \label{eq:KIR}
\end{equation}
It addresses hybrid memory challenges, such as when context overrides parameters, and enhances robustness evaluation in hybrid mechanisms.\citep{tao2024contextleadsparametricmemory}
\item Factual Grounding Score (FGS): while LAMA and ROME evaluate factual accuracy against a known benchmark, FGS expands the concept of accuracy to trustworthiness in open-ended generation. It measures the percentage of a model's factual claims that can be verified against external sources. This metric is critical for mitigating the risk of "confabulation" from parametric memory, ensuring that the model's knowledge outputs are not just plausible but also grounded and reliable.\citep{Shukla2025}
% Requires: \usepackage{amsmath}
\begin{equation}
    \text{FGS} = \frac{\text{Number of Supported Sentences}}{\text{Total Sentences with Factual Information}} \times 100\%
    \label{eq:fgs}
\end{equation}
The score expands the accuracy dimension to include grounding against external documents, mitigating illusions in parameter output.\citep{Shukla2025}
\end{enumerate}
The enhanced framework is the culmination of the parametric memory evaluation story. It takes the core concepts—existence (4.2.1), control (4.2.2), and consistency (4.2.2)—and scales them to meet the complexities of real-world deployment. PPR assesses reliance on memory, LTRS monitors its stability, KIR evaluates its interaction with new information, and FGS certifies the quality of its output. Together, they form an integrated ecosystem that allows researchers and developers to move beyond isolated tests and toward a holistic, operational assessment of knowledge in LLMs.

\subsection{Evaluation of Contextual Memory}
With the continuous extension of context windows in LLMs, assessing their capacity to retain and exploit information over long sequences has emerged as a central research challenge. Existing studies suggest that contextual memory can be examined across three stages: input, reasoning, and output.
At the input stage, evaluations focus on information retention under varying context lengths and positional conditions. Tasks such as key evidence retrieval and mid-sequence extraction highlight the models’ sensitivity to token position in long texts.
The reasoning stage emphasizes the ability to integrate retained evidence for pattern alignment and rule induction, with particular attention to cross-span integration, chain-of-thought reasoning, and robustness to distractors—factors that largely determine whether models can effectively leverage contextual information.
The output stage assesses the alignment and faithfulness of generated responses to the input evidence, using metrics such as factual consistency and completeness of citation chains to distinguish genuine context usage from hallucinatory completion.
This layered perspective moves beyond single accuracy-based metrics and provides a more interpretable and reproducible framework for evaluating contextual memory in LLMs.
\subsubsection{Evaluation Dimensions and Key Metrics} 

\paragraph{1.~Input Stage} 
 A number of existing long-context evaluation studies explicitly focus on memory retention and information localization in the input stage. Representative tasks include key evidence retrieval and mid-sequence extraction, designed to expose models’ deficiencies in positional sensitivity when processing long texts. This phenomenon was first revealed by Lost in the Middle~\cite{liu2023lostmiddlelanguagemodels}, which showed in multi-document QA and key–value retrieval tasks that mainstream models exhibit a sharp decline in utilizing middle-position information, forming a characteristic “U-shaped curve”. The primary evaluation metrics are Answer Accuracy and Evidence Recall, defined as:

\begin{equation}
\mathrm{Accuracy} = \frac{\text{Number of correct answers}}{\text{Total number of questions}},
\end{equation}

\begin{equation}
\mathrm{Recall} = \frac{|E_{\mathrm{gold}} \cap E_{\mathrm{pred}}|}{|E_{\mathrm{gold}}|},
\end{equation}

 where $E_{\text{gold}}$ denotes the annotated evidence set and $E_{\text{pred}}$ the evidence identified by the model. Building on this, Found in the Middle~\cite{hsieh2024middlecalibratingpositionalattention} analyzed the problem at the mechanism level, attributing it to attention biases toward positional information. It proposed a calibration method to decouple position from relevance, achieving up to 15 percentage points of improvement in mid-span evidence utilization. Its evaluation was based on Evidence F1:
 
\begin{equation}
\mathrm{Evidence}\text{-}F_{1} = 
\frac{2 \cdot \mathrm{Precision} \cdot \mathrm{Recall}}{\mathrm{Precision} + \mathrm{Recall}}.
\end{equation}
 
 These diagnostic studies highlight that input-stage evaluation should go beyond overall performance, aiming to reveal differential behaviors across positions. In line with this, the Needle-in-a-Haystack Test~\cite{ivgi2022efficientlongtextunderstandingshorttext} embeds a unique key segment into ultra-long text to directly test retrieval ability. Its core metric is Hit Rate:

\begin{equation}
\mathrm{HitRate} =
\frac{1}{N} \sum_{i=1}^{N} 
\mathbf{1}\!\left\{ \mathrm{pred}_{i} = \mathrm{gold}_{i} \right\}.
\end{equation}

Other works, such as Position Interpolation~\cite{chen2023extendingcontextwindowlarge}, investigate input fidelity from the perspective of extended positional encoding, adopting metrics such as Perplexity (PPL) and Cross-Position Consistency. PPL measures the uncertainty of the model’s predictive distribution given an input position, with lower values indicating greater stability, while Cross-Position Consistency evaluates the consistency of predictions when the same semantic span is placed at different positions.
Comprehensive benchmarks, including LongBench~\cite{bai2024longbenchbilingualmultitaskbenchmark} and InfiniteBench~\cite{zhang2024inftybenchextendinglongcontext}, also incorporate input-stage subtasks such as fact retrieval and span extraction. Their evaluation relies on metrics including Exact Match (EM), which checks whether predictions exactly match the reference, ROUGE-L, which captures the longest common subsequence between prediction and reference, and Evidence Recall, thereby assessing both answer correctness and evidence usage. In Chinese contexts, Bamboo~\cite{dong2024bamboocomprehensivebenchmarkevaluating} similarly includes retrieval and extraction tasks to evaluate evidence recall under ultra-long input conditions.
In addition, benchmarks such as LV-Eval~\cite{yuan2024lvevalbalancedlongcontextbenchmark} construct tiered input lengths (16k–256k) with inserted distractors to expose degradation curves and vulnerability to noise under extreme lengths. The evaluation employs Length-Normalized Accuracy (LN-Acc) and Noise Robustness Score (Robustness):

\begin{equation}
\mathrm{LN\text{-}Acc}(L) =
\frac{\mathrm{Acc}(L)}{\log(L)} ,
\end{equation}

\begin{equation}
\mathrm{Robustness} =
1 - \frac{\mathrm{Acc}_{\mathrm{with\ noise}}}
         {\mathrm{Acc}_{\mathrm{clean}}} .
\end{equation}

Overall, these studies place less emphasis on complex reasoning, centering instead on whether models can accurately capture and reproduce key information in long inputs—thereby providing a reliable memory foundation for subsequent reasoning and generation stages.

\paragraph{2. Evaluation of the Intermediate Stage}
In the evaluation of intermediate-stage contextual memory, the focus shifts from simple information retention to evidence integration, logical reasoning, and robustness against interference. Early work such as Chain-of-Thought~\cite{wei2023chainofthoughtpromptingelicitsreasoning} introduced explicit reasoning chains to guide step-by-step inference. A common evaluation task in this line is step-wise QA, with metrics including final answer accuracy (Answer Accuracy) and step coverage, defined as:

\begin{equation}
\mathrm{StepCoverage} =
\frac{\lvert S_{\mathrm{gold}} \cap S_{\mathrm{pred}} \rvert}
     {\lvert S_{\mathrm{gold}} \rvert} ,
\end{equation}

 where $S_{\text{gold}}$ denotes the set of annotated reasoning steps.
Subsequent studies, such as Faithfulness vs. Plausibility~\cite{agarwal2024faithfulnessvsplausibilityunreliability}, emphasize verifying whether intermediate reasoning chains genuinely rely on input evidence rather than hallucinated completions. Their core task, explanation verification, adopts metrics including the Faithfulness Score (whether explanations are supported by explicit evidence) and the Plausibility Score (whether explanations are semantically reasonable). Both are typically judged by natural language inference (NLI) classifiers:

\begin{equation}
\mathrm{Faithfulness} =
\frac{\lvert E_{\mathrm{supported}} \rvert}
     {\lvert E_{\mathrm{all}} \rvert},
\end{equation}

 where $E_{\text{supported}}$ denotes explanation units that can be grounded in the input.
From a benchmark perspective, SCROLLS~\cite{shaham2022scrollsstandardizedcomparisonlong} provides cross-document QA, summarization, and NLI tasks, with metrics covering Exact Match (EM), ROUGE-L, and Entailment Accuracy. Entailment Accuracy evaluates whether generated answers are semantically entailed by the reference answers, typically determined using NLI models. LongBench extends beyond needle retrieval by introducing multi-hop QA and cross-paragraph reasoning tasks, assessed with F1 and EM, focusing on whether models can effectively integrate evidence in large-scale contexts.
In the Chinese context, Bamboo incorporates multi-evidence integration and chain-of-reasoning QA into its task set, with evaluation based on Evidence F1 and Multi-hop Answer Accuracy, highlighting the model's ability to induce and reason over context under ultra-long input conditions.Unlike standard Answer Accuracy, this metric requires correct answers to be derived from multiple evidence fragments. RULER~\cite{hsieh2024rulerwhatsrealcontext}, by contrast, places particular emphasis on robustness in the presence of distractors and noise. It adopts the RobustQA task with the following metric:

\begin{equation}
\mathrm{RobustAcc} =
\frac{N_{\mathrm{correct\ under\ noise}}}
     {N_{\mathrm{total\ examples}}},
\end{equation}

Correct under noise measuring whether models can still extract and use key information correctly under redundancy or conflicting input. Going further, TimeQA~\cite{chen2021datasetansweringtimesensitivequestions} integrates contextual memory with temporal reasoning, proposing a time-sensitive QA framework with 5.5K facts and 20K QA pairs, distinguishing between explicit and implicit reasoning. Its core metric, Temporal Consistency, is defined as:

\begin{equation}
\mathrm{TempConsist} =
\frac{N_{\mathrm{consistent\ answers}}}
     {N_{\mathrm{total\ temporal\ QA}}} .
\end{equation}

QAconsistent answers
 highlighting limitations of long-context models in maintaining temporal consistency and robustness. This work broadens the evaluation of the intermediate stage, extending beyond cross-paragraph integration to the use of dynamically evolving knowledge.
Although not a long-context benchmark, EntailmentBank~\cite{dalvi2022explaininganswersentailmenttrees} contributes valuable insights by verifying stepwise reasoning chains through entailment checks. It defines Entailment Accuracy as:

\begin{equation}
\mathrm{Acc}_{\mathrm{entail}} =
\frac{\lvert H_{\mathrm{entailed}} \rvert}
     {\lvert H_{\mathrm{all}} \rvert} ,
\end{equation}

 where $H_{\text{entailed}}$ denotes the set of intermediate hypotheses correctly derived by the model.
Overall, evaluation tasks and metrics in this stage converge on three key questions: how models integrate context (multi-hop and chain reasoning), how intermediate reasoning chains can be validated (faithfulness, entailment, and stepwise reasoning), and how robust models are against interference (distractors and conflicting evidence). Together, these assessments complement input retention metrics by establishing a comprehensive framework for evaluating intermediate-stage capabilities.

\paragraph{3.Evaluation of the Output Stage}
 Building on the input stage, which focuses on whether models can accurately read and locate key information, and the intermediate stage, which emphasizes evidence integration and reliable reasoning, the evaluation of the output stage further shifts toward assessing the faithfulness and attributability of generated results.
Representative work such as QAGS~\cite{wang2020askingansweringquestionsevaluate} targets summarization faithfulness evaluation by constructing QA pairs to check whether generated summaries align with the source text. Its core metric is QA-F1, defined as:

\begin{equation}
\mathrm{F1}_{\mathrm{QA}} =
\frac{2 \cdot \mathrm{Precision} \cdot \mathrm{Recall}}
     {\mathrm{Precision} + \mathrm{Recall}} ,
\end{equation}

 where precision and recall are measured based on the alignment between answers and source evidence.
RARR (Gao et al., 2022)~\cite{gao2023rarrresearchingrevisinglanguage} introduces verification of generated content using external knowledge bases or retrieval mechanisms. Within the retrieval-augmented QA setting, it defines Attribution Accuracy:

\begin{equation}
\mathrm{AttrAcc} =
\frac{N_{\mathrm{answers\ with\ correct\ source}}}
     {N_{\mathrm{total\ answers}}} ,
\end{equation}

 highlighting attribution and verifiability as key dimensions.
FactCC~\cite{kryściński2019evaluatingfactualconsistencyabstractive} addresses factual error detection in summarization by formulating a binary classification task (factually correct vs. incorrect), evaluated with Accuracy and F1. FactScore~\cite{min2023factscorefinegrainedatomicevaluation} builds on a QA-based verification task, generating questions and answering them against the source text, with consistency measured by QA-F1:

\begin{equation}
\mathrm{FactScore} =
\frac{\lvert Q_{\mathrm{correct}} \rvert}
     {\lvert Q_{\mathrm{all}} \rvert} ,
\end{equation}

 thus establishing factual consistency as a central metric for the output stage.
In retrieval-augmented generation, RAGAs~\cite{es2025ragasautomatedevaluationretrieval} propose a set of metrics—Answer Faithfulness, Context Precision/Recall, and Answer Relevancy—to assess the alignment between generated output and contextual evidence:

\begin{itemize} 
\item Answer Faithfulness (AF): whether the output is fully supported by the context. 
\item Context Precision (CP) and Context Recall (CR): \begin{equation} 
\mathrm{CP} = \frac{\lvert C_{\mathrm{used}} \cap C_{\mathrm{gold}} \rvert} {\lvert C_{\mathrm{used}} \rvert} , 
\mathrm{CR} = \frac{\lvert C_{\mathrm{used}} \cap C_{\mathrm{gold}} \rvert} {\lvert C_{\mathrm{gold}} \rvert} , 
\end{equation} 
\item Answer Relevancy (AR): whether the output directly addresses the user’s question. 
\end{itemize}

The TRUE Benchmark~\cite{honovich2022truereevaluatingfactualconsistency} aggregates factual consistency tasks from summarization, QA, and dialogue, and unifies evaluation through Fact Consistency Accuracy and QA-F1, thereby providing a cross-task reliability testing platform. For scientific QA and citation generation,
Luo et al.~\cite{luo2023systematicassessmentfactualknowledge} introduce the Attribution \& Citation framework, with the core metric Citation Recall:

\begin{equation}
\mathrm{CitationRecall} =
\frac{\lvert R_{\mathrm{pred}} \cap R_{\mathrm{gold}} \rvert}
     {\lvert R_{\mathrm{gold}} \rvert} ,
\end{equation}

 which requires models not only to generate answers but also to provide verifiable citation chains rather than hallucinated responses.
In addition, comprehensive benchmarks such as HELMET~\cite{yen2025helmetevaluatelongcontextlanguage} categorize long-context evaluation into seven application scenarios, introducing length-controlled experiments and model-based scoring, with emphasis on factual consistency and citation completeness. Similarly, Bamboo and LV-Eval extend beyond input and intermediate tasks to include faithfulness checks at the output level, with tasks such as long-document summarization and QA, evaluated by metrics including Fact-F1, ROUGE-L, and Faithfulness Score, thereby achieving end-to-end coverage.
Overall, evaluations at this stage primarily target dimensions of faithfulness, attribution/citation, coverage, robustness, and quality (e.g., fluency and relevance). The central question is no longer whether information is remembered or whether reasoning is correct, but whether the model can faithfully and verifiably reproduce and cite contextual information at the generation level.

\subsubsection{Benchmark Comparison}

After outlining the evaluation dimensions across the input, intermediate processing, and output stages, comparing existing benchmarks plays a bridging role. On the one hand, these benchmarks provide standardized measures across models and tasks, enabling comparisons of input capacity limits, evidence integration and reasoning chains, as well as output faithfulness and robustness within a unified framework. On the other hand, their differences in coverage, metric orientation, and task design reveal both the evolutionary logic of evaluation systems and the gaps yet to be addressed.

From the perspective of the input stage, LongBenchV2 and Bamboo employ multi-task settings (QA, summarization, code completion) and variable sequence lengths to expose degradation curves in information capacity. However, they still have shortcomings in capturing positional sensitivity and cross-document multi-hop reasoning. HELMET further scales to the 128K level, introducing model-based scoring and citation consistency, thereby highlighting standardization and comparability in evaluation.

At the intermediate processing stage, LV-Eval stands out by systematically introducing distractor facts and keyword substitutions, directly stress-testing model robustness and clearly exposing vulnerabilities in reasoning chains. Although Needle-in-a-Haystack focuses on a single task, it pushes the boundaries of “retrieving key information” to an extreme, serving as a crucial complement for assessing information utilization and resilience to noise.

At the output stage, SCROLLS centers on document-level tasks, emphasizing the faithfulness and compressive quality of generated content. In contrast, HELM~\cite{liang2023holisticevaluationlanguagemodels} and Dynabench~\cite{kiela2021dynabenchrethinkingbenchmarkingnlp} elevate the perspective, focusing on evaluation procedures and data construction mechanisms. They propose dynamic, adversarial, and multi-dimensional metrics to mitigate the limitations of single-task benchmarks in robustness and interpretability.

Overall, the side-by-side presentation of these benchmarks helps clarify the hierarchical characteristics of capabilities across the input–intermediate–output stages while also revealing the complementarities and shortcomings of the current landscape: some emphasize capacity and retrieval, others focus on reasoning and robustness, and still others on generative faithfulness and evaluation workflows. Such diversity makes comparative tables not only a means of information synthesis but also a foundation for building more comprehensive evaluation frameworks.

\newcolumntype{P}[1]{>{\RaggedRight\arraybackslash}p{#1}}
\setlength{\tabcolsep}{2pt} 
\renewcommand{\arraystretch}{1.15}

\begin{longtable}{@{}%
P{0.12\textwidth}% Paper Title
P{0.155\textwidth}% Model
P{0.075\textwidth}% Context Length Range
P{0.140\textwidth}% Task Design
P{0.125\textwidth}% Evaluation Metrics
P{0.185\textwidth}% Results and Findings
P{0.150\textwidth}% Limitation
@{}}
\caption{Long-Context Benchmarks — Models, Tasks, Metrics, Findings, and Limitations}
\label{tab:long_context_benchmarks_multipage}\\
\toprule
\textbf{Paper Title} & \textbf{Model} & \textbf{Context Length Range} &
\textbf{Task Design} & \textbf{Evaluation Metrics} & \textbf{Results and Findings} &
\textbf{Limitations} \\
\midrule
\endfirsthead

\toprule
\textbf{Paper Title} & \textbf{Model} & \textbf{Context Length Range} &
\textbf{Task Design} & \textbf{Evaluation Metrics} & \textbf{Results and Findings} &
\textbf{Limitations} \\
\midrule
\endhead

\bottomrule
\endfoot

Needle-in-a-Haystack &
open-source and closed-source models (e.g., GPT-3.5, Llama series, Claude) &
Context length scalable to 100K+ &
Insert key tokens into ultra-long contexts, requiring the model to precisely locate and respond amid distracting noise &
Precision, Recall, Positional Accuracy &
Most models exhibit “forgetting” or positional drift in extremely long inputs &
Tasks too narrow; lack reasoning/generation; not reflective of real-world scenarios\\

\addlinespace[3pt]
SCROLLS &
Document-level models such as T5, Longformer, LED, BigBird &
5K--50K &
Cross-task suite including long-document QA, summarization, and natural language inference (NLI) &
ROUGE, F1, Accuracy, Entailment Score &
Document-level tasks  test models’ abilities in information compression and cross-passage integration; different tasks expose model weaknesses in distinct ways  &
English-focused; lacks code/multimodal tasks and long-input evaluation\\

\addlinespace[3pt]
HELM &
30+ models including GPT-3, OPT, BLOOM, Jurassic-1 &
2K--32K &
Generalized task set (QA, summarization, dialogue, reasoning) with holistic evaluation &
Accuracy, Calibration, Robustness, Fairness, Efficiency &
Proposes the first framework evaluating performance, fairness, robustness, and efficiency; closed-source models outperform open-source across these metrics &
Limited to contexts $\leq$32k;  framework-focused, neglects specific long-text reasoning \\

\addlinespace[3pt]
Dynabench &
RoBERTa, T5, GPT-3, etc. &
Mostly $\leq$4K &
Tasks cover QA, NLI, and dialogue &
Accuracy, F1, human discrimination accuracy &
Model robustness drops significantly when facing adversarial data and dynamic distributions; continuous iteration exposes model blind spots &
Not suited for long-text; adversarial data has scalability challenges \\

\addlinespace[3pt]
LongBenchV2 &
GPT-3.5-Turbo-16k, Llama2-7B-chat-4k, LongChat-v1.5-7B-32k, XGen-7B-8k, InternLM-7B-8k, ChatGLM2-6B, ChatGLM2-6B-32k, Vicuna-v1.5-7B-16k &
4K, 8K &
Single-Document QA, Multi-Document QA, Summarization, Few-shot Learning, Synthetic Task, Code Completion &
Unified format + traditional automatic metrics (F1, Rouge-L, Accuracy, Edit Similarity) &
In long-context tasks, a performance gap between open-source models and commercial models. Models benefit from scaled positional embeddings and continued training on longer contexts &
Using F1 and Rouge to evaluate long-text generation tasks cannot accurately reflect response quality; evaluating with LLMs is costly and introduces potential biases \\

\addlinespace[3pt]
HELMET &
59 long-context models (closed- and open-source) &
8K, 16K, 32K, 64K, 128K &
RAG, passage re-ranking, long-document QA, summarization, many-shot in-context learning, synthetic recall &
SubEM, Recall, NDCG@10, Accuracy, model-based scores, ROUGE, F1 &
Open-source models trail in long-text and complex instruction tasks, with a widening gap over context length; performance degradation is task-dependent, severe in reordering and citation generation &
Model-based scoring has bias; tasks are English-only with limited multilingual/multimodal coverage; 128K evaluation is costly \\

\addlinespace[3pt]
Bamboo &
gpt-3.5-turbo-16k, Claude2-100k, ChatGLM2-32k, Vicuna-v1.5-16k, LongChat-v1.5-16k &
4K, 16K &
QA, hallucination detection, ranking, language modeling, code completion &
EM, F1, hallucination detection accuracy, ranking accuracy, language model perplexity, code task accuracy &
Performance degrades over long contexts; degradation and hallucination/ranking gaps vary by task; data control and task standardization enable stable comparisons &
Conservative length coverage; insufficient memory ability tests; task variety lacks real-world relevance (e.g., evidence-based, cross-modal tasks) \\

\addlinespace[3pt]
LV-Eval &
15 long-context models &
16K, 32K, 64K, 128K, 256K &
Single-hop and multi-hop QA over 11 bilingual datasets; adds distractor-fact insertion and keyword/phrase replacement &
Keyword-recall–oriented metrics with EM/F1; robustness via distractor facts/keyword replacement; degradation curves &
Sharp drops at 64K+ (stronger at 128K/256K); error spikes after distractor insertion; confusion and retrieval under long contexts weaker than expected &
Task scope mostly QA; no generation/summarization/code; extreme-length evaluation is compute-intensive; limited cross-domain and multimodal support \\
\end{longtable}

\subsubsection{Open Challenges and Future Directions}
Although current research on long-context memory evaluation has established a relatively systematic framework across the three stages of input capacity, reasoning utilization, and output faithfulness, several unresolved challenges remain:

Fragmentation and limited coverage of evaluation dimensions.
 At the input stage, most benchmarks emphasize evidence localization and information retrieval, highlighting positional sensitivity and length degradation, but they rarely consider cross-modal contexts or interactive inputs. In the intermediate stage, while multi-evidence integration and robustness designs have been introduced, they are often confined to static text tasks and thus fail to reflect temporal consistency and knowledge evolution in dynamic environments. At the output stage, faithfulness checking has made solid progress in summarization and QA, yet existing metrics often fall short in more complex generative tasks such as dialogue, decision-making, or cross-domain citation.
 
Limited interpretability of metrics.
 Although refined indicators such as Evidence Recall, Faithfulness Score, and Attribution Accuracy have been proposed, these typically capture only pointwise correspondences (e.g., answer alignment or citation chain coverage) and lack quantification of reasoning path completeness or generation robustness. Current evaluations often expose performance differences without clarifying the reasons for model failures, thereby offering limited diagnostic and guidance value.
Monolingual and static bias in benchmarks.

 Mainstream benchmarks such as SCROLLS, LongBench, and HELMET are predominantly based on English corpora, with limited evaluation resources in Chinese and other languages. Moreover, most benchmarks are constructed using static datasets, making it difficult to capture the memory challenges posed by user interaction, real-time information updates, and cross-domain generalization.
Disconnection between extreme-length testing and real applications.
 While benchmarks such as LV-Eval have extended input windows up to 256K tokens, many tasks remain synthetic in nature and diverge from real-world demands such as long-document retrieval, multi-section writing, or scientific paper comprehension. This tension between “stress testing” and “application-oriented evaluation” reduces the practical relevance of current metrics for deployment scenarios.
 
Lack of unified frameworks and holistic evaluation.
 Existing benchmarks emphasize different dimensions—capacity, robustness, or faithfulness—resulting in fragmented coverage. There is no unified end-to-end evaluation framework that integrates the input–reasoning–output chain, nor interoperability between task types and metric layers. Consequently, model performance across different benchmarks remains difficult to align or interpret.
 
Future directions for long-context memory evaluation may unfold along several paths:
Integrated cross-stage evaluation: Develop unified benchmarks that span input, reasoning, and output, enabling holistic characterization of a model’s “memory chain” rather than fragmented task-specific assessments.
Process interpretability and causal diagnosis: Establish metric systems that can trace reasoning chains and validate intermediate states—for example, stepwise entailment checks or causal dependency analysis—to pinpoint where errors arise.
 Multilingual and multimodal expansion: Extend evaluation to Chinese, low-resource languages, and multimodal inputs such as text–image, code, and structured data, thereby enhancing coverage and practical utility.
Dynamic and interactive evaluation: Inspired by adversarial approaches such as Dynabench, introduce human-in-the-loop evaluation, time-sensitive QA, and knowledge update scenarios to test long-term consistency and robustness under realistic conditions.
Application-oriented task design: Align benchmarks with high-value application domains—such as retrieval-augmented generation, scientific writing assistance, or legal case analysis—ensuring that tasks and metrics provide actionable insights for model optimization and deployment.

\subsection{Evaluation of External Memory}
\subsubsection{Evaluation Data, Methods, and Metrics}
External memory systems mitigate the inherent limitations of parametric models—specifically in knowledge timeliness, traceability, and controllability—by incorporating updatable, non-parametric knowledge during inference. The typical pipeline “query rewriting → retrieval/reranking → reading → generation” exhibits tightly coupled modules, wherein errors may propagate and amplify across stages. Consequently, evaluation must first ensure comparability and interpretability of end-to-end performance, followed by fine-grained diagnostics of critical components such as retrieval and generation. Furthermore, evaluation dimensions must continuously expand in alignment with technological advancements such as real-time responsiveness, multimodality, and domain specialization. Current research has gradually converged on a four-tier collaborative framework: “retrieval quality — generation quality — end-to-end performance — robustness and timeliness,” implemented within unified datasets and benchmark ecosystems. Below, we present an integrated exposition of metrics, methods, and datasets under this framework.

1. Retrieval Quality: Dual Tracks of Static Relevance and Task Utility  
Retrievers and rerankers are regarded as the “performance ceiling” of RAG systems; downstream task accuracy exhibits near-linear dependence on retrieval quality. Any recall gap or ranking bias inevitably leads to irreversible degradation during generation \citep{lewis2021retrievalaugmentedgenerationknowledgeintensivenlp}.  

(1) Static relevance adopts standard zero-shot benchmarks such as BEIR and KILT, employing metrics including nDCG@k, MRR@k, Recall@k, and Precision@k for cross-system comparability. An nDCG@10 $\geq$ 0.7 is commonly treated as a warning threshold for adequate ranking; in knowledge-intensive domains such as medicine and law, Recall@k $\geq$ 80\% is often set as a hard constraint to prevent loss of critical evidence prior to generation \citep{thakur2021beirheterogenousbenchmarkzeroshot,petroni2021kiltbenchmarkknowledgeintensive,yang2024cragcomprehensiverag}.  

(2) Task utility is quantified via ``marginal contribution'' $\frac{\Delta EM}{\Delta F_1}$: the greater the decline in downstream metrics upon removal of a single passage, the higher its practical value. eRAG series experiments demonstrate that reranking yields a relative improvement of 6–12\% in passage-level recall, simultaneously enhancing answer correctness and evidence faithfulness\citep{yu2024rankragunifyingcontextranking}.  

(3) Logical coherence transcends surface-level semantics, evaluating internal document coherence and document-query alignment along factual, causal, and narrative chains—including information completeness, causal consistency, and contextual coherence. Due to the difficulty of automated parsing, this dimension still relies primarily on human scoring or task-oriented indirect metrics (e.g., reasoning accuracy of generated answers), representing the largest blind spot in automated evaluation \citep{izacard2021leveragingpassageretrievalgenerative,gao2023enablinglargelanguagemodels}[5,6].  
In summary, mainstream consensus advocates dual-track reporting in evaluation: one track preserves cross-system comparability via metrics such as nDCG@10, MRR@10, and Recall@100; the other directly measures practical contribution to final outputs using task-driven metrics like $\frac{\Delta EM}{\Delta F_1}$, thereby establishing a “dual-track” evaluation paradigm \citep{izacard2021leveragingpassageretrievalgenerative}.

2. Generation Quality: Four Progressive Dimensions — Evidence Attribution, Output Performance, Context Utilization, and Logical Interpretability  
Generation quality exhibits significant interaction effects with retrieval results: when retrieval nDCG@5 > 0.75, generation optimization can improve FActScore by approximately 14.2\%; otherwise, gains diminish by over 62\%. Thus, multidimensional evaluation from an end-to-end perspective is essential.  
(1) Evidence attribution and faithfulness verify that generated content strictly originates from retrieved evidence, suppressing pseudo-citations and hallucinations. The ALCE framework quantifies citation-augmented generation across fluency, correctness, and citation quality; QAFactEval constructs QA chains to test output-context consistency; FActScore decomposes long-form text into atomic facts and computes the proportion supported by evidence—the higher the proportion, the more solid the factual foundation. RAGAS and ARES provide reference-free faithfulness scores, with ARES introducing confidence intervals to mitigate drift bias in LLM-based evaluators; FACTS Grounding and GaRAGe conduct rigorous groundedness evaluation based on span-level annotations. Given that correctness and faithfulness are not equivalent, it is recommended to report both metrics and supplement them with span-level precision and recall to avoid macro-level metrics masking local pseudo-citations.  

(2) Output quality and task performance must be assessed using task-specific metrics. Open-domain QA commonly employs EM and F1; long-form generation typically uses ROUGE or BLEU. However, surface overlap metrics may obscure evidence omission issues and must therefore be reported jointly with faithfulness metrics. RAGTruth provides word-level and span-level hallucination annotations; its hallucination density metric locates fabricated content within generations, offering fine-grained feedback for model refinement.  

(3) Context adequacy and selective answering examine system behavior under evidence-sufficient and evidence-deficient conditions. By partitioning test samples into “sufficient evidence” and “insufficient evidence” categories, one can evaluate accuracy under answerable conditions and refusal capability under unanswerable conditions. Selective accuracy measures the proportion of correct answers among answerable samples; refusal rate measures the proportion of correct refusals among unanswerable samples—both jointly reflect the system’s risk control capability under uncertainty. The Sufficient Context framework recommends joint analysis of these metrics with retrieval recall: evaluating evidence utilization efficiency on the “sufficient” subset and assessing system restraint on the “insufficient” subset.  

(4) Logical interpretability examines whether generated content possesses traceable reasoning paths. Semantic interpretability measures the proportion of assertions traceable to evidence—the higher the proportion, the more transparent the system’s decision-making. The OPI index holistically evaluates generation quality, logical correctness, and input consistency by harmonically averaging logical relation accuracy and semantic similarity, thereby suppressing inflated scores in single dimensions.  
A tiered evaluation strategy is recommended: the foundational tier centers on FActScore and hallucination frequency to ensure factual reliability; the enhanced tier introduces faithfulness and OPI to improve interpretability; the specialized tier incorporates domain-specific metrics to form a gradient evaluation system. Evaluation methods should combine automated metrics, human validation, and user studies to enhance robustness.

3. End-to-End System Efficacy: A Retrieval-Generation Synergy Perspective  
End-to-end evaluation seeks to integrate retrieval and generation stages to holistically measure RAG performance. Methods such as RECALL, FeB4RAG, and Long²RAG construct closed-loop test environments to simultaneously assess retrieval precision and generation quality. For instance, RECALL combines EventKG and UJ datasets to measure retrieval accuracy and generation error recurrence rate, revealing system stability under noise interference; Long²RAG, targeting long-form generation tasks, designs joint metrics for key-point recall and long-text generation accuracy, emphasizing information completeness. Such evaluations reveal synergistic effects and bottlenecks across system components: studies find that even with significant improvements in retrieval relevance, generation quality gains remain relatively limited, indicating that knowledge transformation efficiency has become a key constraint on overall performance. However, most end-to-end evaluations still rely on synthetic news-like data, lacking the complex noise and interaction patterns of real-world scenarios, potentially leading to significant deployment-performance gaps. In contrast, comprehensive evaluation frameworks such as CRAG and RAGBench support multi-domain, multi-question-type task settings and provide interpretability analysis interfaces [17,18], offering more representative evaluation environments for complex scenarios and addressing limitations of single-task benchmarks.

4. Robustness and Timeliness: Facing Real-World Perturbations and Knowledge Updates  
Robustness and timeliness evaluations aim to test whether RAG systems can maintain expected performance under real-world conditions such as knowledge updates, input perturbations, and evidence conflicts. Recently, benchmarks such as RARE, RGB, and QE-RAG have systematically injected perturbations and designed tasks to transform “stability” into measurable properties, providing clear performance boundaries for deployment-grade systems.  

(1) RARE introduces controlled perturbations at query, document, and retrieval levels to quantify system response consistency under semantic drift, information loss, and ranking jitter; RGB focuses on noise filtering, negative-sample rejection, information integration, and counterfactual reasoning, stress-testing decision reliability under conflicting or misleading contexts; QE-RAG evaluates the effectiveness of underlying fault-tolerance mechanisms using real-world input noise such as spelling errors and word-order inversions. Collectively, these works reveal RAG’s adaptive limits in complex environments and drive evaluation protocols to evolve from static accuracy to dynamic robustness.  

(2) Current robustness metrics span the entire query-document-retrieval pipeline:  

a) Query robustness constructs perturbed query sets via paraphrasing or inserting irrelevant phrases, computing the average proportion of correct answers retained post-perturbation to measure semantic generalization and intent preservation;  
 
b) Document robustness injects irrelevant or conflicting sentences into input passages to test whether models can suppress noise and maintain answer consistency, directly reflecting the efficacy of evidence filtering mechanisms;  
 
c) Real-world retrieval robustness evaluates generation sensitivity to retrieval result fluctuations by switching retrieval or reranking strategies, simulating variations from heterogeneous retrieval pipelines in deployment environments;  
 
d) Token-level F1 and Exact Match under the QE-RAG framework measure generation quality for queries with spelling errors—the former based on token overlap, the latter requiring exact string-level answer matches—forming a spectrum of error tolerance from loose to strict;  
 
e) Noise robustness measures the system’s ability to ignore irrelevant information via task accuracy on noisy corpora; negative-sample rejection capability quantifies the system’s conservative tendency to abstain when lacking supporting evidence, via a weighted combination of rejection rate and accuracy; counterfactual robustness, under context-parameter knowledge conflicts, measures the proportion of answers prioritizing external evidence, evaluating dependence strength on external memory.  
 
(3) Timeliness evaluation is increasingly scenario-specific. CRAG categorizes facts by update velocity into real-time, fast-changing, slow-changing, and stable classes, assigning differentiated reference answers to avoid evaluating dynamic knowledge with static labels; the HOH dynamic benchmark continuously injects time-sensitive queries to validate RAG’s practical gains in alleviating LLM knowledge-update bottlenecks; FreshQA uses freshness hit rate and outdated answer rate as core metrics to quantify the system’s ability to balance timeliness and accuracy in multi-hop reasoning, driving architectures such as MQRF-RAG to achieve 14.45\% relative improvement. It should be noted that evaluations based on “closed-set answerability” assumptions (e.g., KILT) fundamentally differ from FreshQA’s dynamic update setting; experimental reports must explicitly state contextual premises to prevent metric misalignment.

5. Benchmarks and Datasets: From Single-Function to Multi-Dimensional Dynamics  
In recent years, RAG evaluation has evolved in benchmark design and dataset construction from single-function validation toward multidimensional, systematic assessment. The current evaluation ecosystem has gradually formed a classification framework centered on retrieval quality, generation quality, end-to-end performance, and robustness/timeliness, reflecting the evolution of evaluation priorities and revealing deeper differences across methods in objective setting, data selection, and metric design.  

(1) For retrieval quality, methods such as RAGAs, ARES, and MultiHop-RAG employ standard datasets like WikiEval, NQ, and Hotpot, using traditional IR metrics such as MAP, MRR, and Hit@K. GraphRAG constructs datasets with factual associations and complex reasoning paths to test graph-structured retrieval; CoFE-RAG introduces multi-document formats and diverse query types to improve coverage of real-world document heterogeneity. However, such evaluations still largely rely on static corpora, exhibiting limited capacity to model knowledge updates and dynamic scenarios. BEIR and KILT, as general-purpose evaluation backbones for cross-domain zero-shot retrieval and knowledge-intensive tasks, provide broadly comparable benchmarks for RAG retrieval capabilities\citep{thakur2021beirheterogenousbenchmarkzeroshot,petroni2021kiltbenchmarkknowledgeintensive}. Their coverage of diverse tasks and unified Wikipedia snapshots support evaluation of model generalization in unseen domains, serving as foundational platforms for most retrieval module validation.  

(2) For generation quality, methods such as ARES, FeB4RAG, and DomainRAG quantify generation accuracy using F1, ROUGE-L, and Exact-Match, while introducing LLM-as-a-Judge mechanisms to assess semantic consistency and clarity. ReEval leverages datasets like RealTimeQA, containing time-sensitive questions, to test model performance under outdated or missing knowledge. Although automated metrics enhance evaluation efficiency, their alignment with human judgment remains contested: BERGEN’s research reveals weak semantic-level correlation in LLM evaluations, exposing current deficiencies in reliability and interpretability. Methods such as ALCE, QAFactEval, FActScore, FACTS, GaRAGe, and RAGTruth aim to enhance interpretability and factual alignment of generated content, providing fine-grained analyses from citation quality to span-level knowledge attribution\citep{gao2023enablinglargelanguagemodels,fabbri2022qafactevalimprovedqabasedfactual,min2023factscorefinegrainedatomicevaluation,jacovi2025factsgroundingleaderboardbenchmarking,sorodoc2025garagebenchmarkgroundingannotations,niu2024ragtruthhallucinationcorpusdeveloping}, thereby deepening evaluation from “whether correct” to “why correct,” enhancing result credibility and debugging support.  

(3) For end-to-end system efficacy, methods such as RECALL, FeB4RAG, and Long²RAG construct closed-loop test environments to simultaneously evaluate retrieval precision and generation quality; comprehensive frameworks like CRAG and RAGBench support multi-domain, multi-question-type task settings and provide interpretability analysis interfaces\citep{yang2024cragcomprehensiverag,friel2025ragbenchexplainablebenchmarkretrievalaugmented}, offering more representative evaluation environments for complex scenario comparisons.  

(4) For robustness and timeliness, RGB and CRUD-RAG address noise robustness and knowledge manipulation dimensions respectively: the former tests model resistance to misleading information; the latter designs CREATE, UPDATE, DELETE scenarios to evaluate system management of knowledge lifecycles. Benchmarks such as FreshQA, RAMDocs, and RGB further intensify stress testing for knowledge conflicts and update delays\citep{vu2023freshllmsrefreshinglargelanguage,wang2025retrievalaugmentedgenerationconflictingevidence,chen2023benchmarkinglargelanguagemodels}, revealing model reasoning fragility under knowledge inconsistency. Additionally, with expanding application scenarios, long-context and multimodal evaluation have become focal points: LongBench, RULER, and Long²RAG pose challenges for document-level understanding and long-text generation; MM-Needle, VisDoMBench, and T²-RAGBench construct cross-modal retrieval and generation tasks \citep{bai2024longbenchbilingualmultitaskbenchmark,hsieh2024rulerwhatsrealcontext,wang2024multimodal,suri2025visdommultidocumentqavisually, strich2025t2ragbenchtextandtablebenchmarkevaluating}, extending evaluation from pure text to image-text fusion scenarios. Emerging efforts such as Visual-RAG and RAG-Check have begun constructing multimodal datasets but remain exploratory in real-world interaction and dynamic knowledge fusion.

In summary, RAG evaluation datasets and benchmarks are evolving toward specialization, contextualization, and dynamism: evaluation dimensions have expanded from initial accuracy testing to encompass retrieval, generation, consistency, and timeliness; data sources have extended from general QA datasets to domain-specific, multimodal, and real private corpora. General-purpose frameworks such as BEIR, KILT, CRAG, and RAGBench provide foundations for cross-system comparison, while specialized benchmarks such as ALCE, FActScore, FreshQA, and LongBench deepen evaluation capabilities in specific dimensions. However, the current evaluation ecosystem still faces significant challenges: lack of a unified framework across dimensions hinders cross-benchmark comparability; synthetic data, while facilitating controlled experiments, diverges from real-world scenarios; automated evaluation relying on LLM judgments may introduce evaluation bias. Future research must, while preserving evaluation granularity, strengthen cross-dimensional integration and develop more ecologically representative dynamic evaluation paradigms, advancing RAG evaluation from technical validation toward practical value measurement.

% ================= 表格开始 =================
\savegeometry{normal}  % ← 保存为 normal，供后面还原
\clearpage
\newgeometry{left=1.0cm, right=1.0cm, top=1.0cm, bottom=1.0cm, includefoot} % 进一步缩小边距

\begin{table}[p] % 强制整页浮动
\centering
\fontsize{7.4pt}{7.5pt}\selectfont % 手动设置极小字号（6.5pt），行距7.5pt
\setlength{\tabcolsep}{2pt} % 列间距压到最小
\renewcommand{\arraystretch}{0.95} % 行高进一步压缩
\captionsetup{skip=2pt, font=scriptsize, labelfont=bf} % 标题也缩小
\caption{RAG Evaluation Methods — Classification and Specification (Compact Version)}
\label{tab:rag_eval_methods_compact}

% 使用 tabularx，最后一列自动换行
\begin{tabularx}{\textwidth}{@{}
    >{\raggedright\arraybackslash}p{2.3cm}
    >{\raggedright\arraybackslash}p{2.1cm}
    >{\raggedright\arraybackslash}p{2.5cm}
    >{\centering\arraybackslash}p{1.2cm}
    X
@{}}
\toprule
\textbf{Evaluation Type} & \textbf{Method Name} & \textbf{Dataset / Source} &
\textbf{Release Date} & \textbf{Dimensions \& Metrics} \\
\midrule

\rowcolor{gray!10}
Retrieval-Quality-Oriented &
RAGAs \citep{es2025ragasautomatedevaluationretrieval} &
WikiEval &
2023.09 &
Contextual Relevance: Extracted Sentences / Total Sentences, Average Cosine Similarity \\
\addlinespace[2pt]

&
MultiHop-RAG\citep{tang2024multihop} &
Self-constructed Dataset &
2024.01 &
Multi-hop Retrieval Quality: MAP, MRR, Hit@K \\
\addlinespace[2pt]

\rowcolor{gray!10}
&
CoFE-RAG \citep{liu2024cofe} &
Self-constructed Dataset &
2025.06 &
Retrieval Quality, Query Localization Ability: Keyword-based Retrieval Score, Question Locating Accuracy \\
\addlinespace[2pt]

&
GraphRAG\citep{edge2025localglobalgraphrag} &
VIINA &
2024.04 &
Graph-structured Retrieval, Reasoning Comprehensiveness \& Diversity: Fact Retrieval Accuracy, Complex Reasoning Score \\
\addlinespace[2pt]

\rowcolor{gray!10}
&
EnronQA\citep{ryan2025enronqa} &
Enron Email Dataset &
2025.05 &
Personalized Retrieval Quality: Contextual Relevance Score \\
\addlinespace[2pt]

&
Visual-RAG\citep{bonomo2025visual} &
Self-constructed Dataset &
2025.02 &
Multimodal Retrieval Quality: Image Retrieval Accuracy \\
\addlinespace[2pt]

\rowcolor{gray!10}
&
ByoKG-RAG\citep{mavromatis2025byokg} &
Self-constructed Dataset &
2025.07 &
Graph Retrieval Quality: Graph Retrieval Score \\
\addlinespace[2pt]

&
Hypercube-RAG \citep{shi2025hypercube} &
Self-constructed Dataset &
2025.05 &
Retrieval Efficiency: Retrieval Efficiency Score \\
\addlinespace[2pt]

\rowcolor{gray!10}
&
eRAG\citep{salemi2024evaluatingretrievalqualityretrievalaugmented} &
NQ, HotpotQA, etc. &
2024.04 &
Task Utility-Oriented Retrieval (Marginal Contribution): $\Delta$EM, $\Delta$F1 \\
\addlinespace[2pt]

&
BEIR\citep{thakur2021beirheterogenousbenchmarkzeroshot} &
MS MARCO, TREC-COVID, NQ, etc. &
2021.04 &
Zero-shot Cross-domain Retrieval Quality: nDCG@10, Recall@100, MRR@10 \\
\addlinespace[2pt]

\rowcolor{gray!10}
&
KILT\citep{petroni2021kiltbenchmarkknowledgeintensive} &
11 datasets based on unified Wikipedia snapshot (e.g., FEVER, NQ) &
2020.09 &
Retrieval Performance in Knowledge-Intensive Tasks: Task-specific metrics (e.g., Label Accuracy for FEVER, EM/F1 for NQ) \\
\addlinespace[4pt]

\rowcolor{gray!10}
Generation-Quality-Oriented &
ARES \citep{saadfalcon2024aresautomatedevaluationframework} &
NQ, Hotpot, FEVER, WoW, MultiRC, ReCoRD &
2023.11 &
Answer Faithfulness, Answer Relevance: Confidence Intervals (for generation faithfulness) \\
\addlinespace[2pt]

&
MedRAG\citep{zhao2025medrag} &
Self-constructed Dataset &
2024.02 &
Generation Accuracy: Accuracy \\
\addlinespace[2pt]

\rowcolor{gray!10}
&
CDQA \citep{xu2024let} &
Self-constructed Dataset &
2024.03 &
Generation Accuracy: F1 \\
\addlinespace[2pt]

&
ReEval \citep{yu2023reeval} &
RealTimeQA, NQ &
2024.06 &
Hallucination Detection: F1, Exact-Match, LLM-as-a-Judge, Human Evaluation \\
\addlinespace[2pt]

\rowcolor{gray!10}
&
ASTRID \citep{chowdhury2025astrid} &
WikiEval &
2023.10 &
Answer Relevance, Fact Attribution (Groundedness): LLM-as-a-Judge \\
\addlinespace[2pt]

&
BERGEN \citep{rau2024bergen} &
QA Datasets &
2024.06 &
Surface \& Semantic Consistency: EM, F1, Precision, Recall, BEM, LLMeval \\
\addlinespace[2pt]

\rowcolor{gray!10}
&
RAG-Check \citep{mortaheb2025rag} &
Self-constructed Dataset &
2025.01 &
Multimodal Generation Quality: Generation Quality Metric \\
\addlinespace[2pt]

&
Long\textsuperscript{2}RAG \citep{qi2024long} &
Self-constructed Dataset &
2025.06 &
Long-form Generation Quality: Long-Form Generation Accuracy \\
\addlinespace[2pt]

\rowcolor{gray!10}
&
FActScore\citep{min2023factscorefinegrainedatomicevaluation} &
ASQA &
2023.05 &
Atomic Fact-level Faithfulness: Proportion of atomic facts supported by evidence \\
\addlinespace[2pt]

&
QAFactEval\citep{fabbri2022qafactevalimprovedqabasedfactual} &
SummaC, ASQA, etc. &
2021.12 &
Generation Consistency via QA Chains: QA-based F1, Precision, Recall \\
\addlinespace[2pt]

\rowcolor{gray!10}
&
ALCE\citep{gao2023enablinglargelanguagemodels} &
Specialized dataset with diverse questions \& retrieval corpus &
2023.05 &
Fluency, Correctness, Citation Quality of Citation-Augmented Generation: Citation Precision, Citation Recall, Answer Accuracy \\
\addlinespace[2pt]

\rowcolor{gray!10}
&
Sufficient Context\citep{joren2024sufficient} &
Applied on benchmarks: FreshQA, Musique, HotpotQA, etc. &
2024.11 &
Evidence Adequacy, Selective Answering \& Refusal Capability: Sufficient/Insufficient Context Label, Selective Accuracy, Refusal Rate \\
\addlinespace[2pt]

&
GaRAGe\citep{sorodoc2025garagebenchmarkgroundingannotations} &
Self-constructed Dataset &
2025.06 &
Fine-grained evaluation of LLM’s ability to identify \& utilize relevant evidence spans: Relevance-Aware Factuality, Grounding Accuracy \\
\addlinespace[2pt]

\rowcolor{gray!10}
&
FACTS Grounding\citep{jacovi2025factsgroundingleaderboardbenchmarking} &
Self-constructed Dataset &
2025.01 &
Span-level Knowledge Attribution \& Fact Alignment: Grounding Score — Information pending confirmation \\
\addlinespace[4pt]

\rowcolor{gray!10}
Robustness \& Timeliness-Oriented &
RGB \citep{chen2023benchmarkinglargelanguagemodels} &
Self-constructed Dataset &
2023.12 &
Noise Robustness, Negative-sample Rejection, Counterfactual Robustness: Accuracy, Rejection Rate, Error Detection Rate, Error Correction Rate \\
\addlinespace[2pt]

&
CRUX \citep{ju2025controlled} &
Self-constructed Dataset &
2025.06 &
Context Integrity, Redundancy Control: LLM-based Evaluation, Human Data Assessment \\
\addlinespace[2pt]

\rowcolor{gray!10}
&
Diverse And Private Synthetic Datasets Generation \citep{driouich2025diverse} &
Synthetic QA Dataset &
2025.08 &
Robustness under Data Diversity \& Privacy Preservation: Diversity Score, Privacy Masking Accuracy \\
\addlinespace[2pt]

&
RAG Without the Lag \citep{lauro2025rag} &
Self-constructed Dataset &
2025.04 &
Interactive Debugging Robustness, Pipeline Stability: Debugging Efficiency, Pipeline Performance Metric \\
\addlinespace[2pt]

\rowcolor{gray!10}
&
RARE \citep{zeng2025rare} &
Self-constructed Dataset &
2025.06 &
Robustness to Query, Document, and Retrieval Perturbations: RARE-Met (Systematic Robustness Metric) \\
\addlinespace[2pt]

&
QE-RAG \citep{zhang2025qe} &
Self-constructed Dataset &
2025.04 &
Robustness to Input Query Errors: Token-level F1 and EM for error-containing queries \\
\addlinespace[2pt]

\rowcolor{gray!10}
&
CRAG\citep{yang2024cragcomprehensiverag} &
Self-constructed Dataset &
2024.06 &
Comprehensive Evaluation: Popular/Long-tail Entities, Static/Time-sensitive Facts, Retrieval Failure Handling: Accuracy, Interpretability Analysis Interface \\
\addlinespace[2pt]

&
FreshQA\citep{vu2023freshllmsrefreshinglargelanguage} &
Self-constructed Dataset &
2023.10 &
Ability to Retrieve Up-to-date Information: Freshness Hit Rate, Outdated Answer Rate \\
\addlinespace[2pt]

\rowcolor{gray!10}
&
HOH\citep{ouyang2025hoh} &
Self-constructed Dataset &
2025.06 &
Negative Impact of Outdated Information on RAG Performance: Accuracy Drop Magnitude, Dependency on Outdated Information \\
\addlinespace[2pt]

&
CRUD-RAG\citep{lyu2025crud} &
Self-constructed Dataset &
2024.01 &
Capability in Knowledge Lifecycle Management (Create/Read/Update/Delete): Task Accuracy under each CRUD operation scenario \\
\addlinespace[2pt]

\rowcolor{gray!10}
&
RAMDocs\citep{wang2025retrieval} &
Self-constructed Dataset &
2025.04 &
Stress Testing under Knowledge Conflicts, Noise, and Misinformation: Accuracy \\
\addlinespace[4pt]

\rowcolor{gray!10}
End-to-End System Performance-Oriented &
FeB4RAG \citep{wang2024feb4rag} &
BEIR, Generated (Source: News), Labeller &
2024.02 &
Consistency, Correctness, Clarity, Coverage: Human Evaluation, F1, Exact-Match, ROUGE-L, LLM-as-a-Judge \\
\addlinespace[2pt]

&
DomainRAG \citep{wang2024domainrag}  &
College Admission Information &
2024.06 &
Structured Output Capability (System-level Performance): LLM-as-a-Judge \\
\addlinespace[2pt]

\rowcolor{gray!10}
&
RECALL \citep{liu2023recall} &
EventKG, UJ &
2023.11 &
Response Quality, System Stability: Accuracy (QA), BLEU, ROUGE-L, Misleading Rate, Mistake Reappearance Rate \\
\addlinespace[2pt]

&
RAGBench\citep{friel2025ragbenchexplainablebenchmarkretrievalaugmented}&
Self-constructed Dataset &
2024.06 &
Interpretable End-to-End Evaluation: TRACe Framework: Utilization, Relevance, Adherence, Completeness \\
% === 表格内容结束 ===

\bottomrule
\end{tabularx}
\end{table}

\restoregeometry
\clearpage               % ← 把使用 1cm 版芯的浮动页全部排完
\loadgeometry{normal}    % ← 精确回到全局几何（比 \restoregeometry 更保险）
% ================= 表格结束 =================

\subsubsection{Recommended Reproducible Evaluation Framework}

The effectiveness of external knowledge memory can be evaluated across four dimensions: retrieval quality, generation quality, end-to-end system performance, and dynamic adaptability (robustness \& timeliness).

As the knowledge entry point of RAG systems, retrieval quality exerts a decisive influence on overall system performance. This study identifies MRR, Hit@K, and Contextual Relevance Score as the core metrics for this dimension.

On the BEIR and KILT datasets, we recommend primarily using MRR to evaluate cross-domain zero-shot retrieval capability. These datasets provide a unified testbed covering 18 diverse tasks, effectively validating model generalization performance in unseen domains. For multi-hop reasoning scenarios, Hit@K@5 and Contextual Relevance Score should be jointly applied on the MultiHop-RAG dataset: the former evaluates the system’s ability to retrieve sufficient evidence, while the latter assesses semantic alignment between retrieved results and complex queries. On GraphRAG-Bench, particular attention should be paid to the correlation between Contextual Relevance Score and Complex Reasoning Score to evaluate the effectiveness of graph-structured knowledge retrieval. For personalized retrieval scenarios, Enron email corpus testing should combine Hit@K@3 with human evaluation to validate the system’s precise targeting capability in private knowledge environments. In multimodal retrieval evaluation, the Visual-RAG and MM-Needle datasets should integrate Image Retrieval Accuracy with Contextual Relevance Score to comprehensively measure cross-modal semantic understanding.

The generation quality dimension directly determines the output credibility of RAG systems, with F1, Exact-Match, Faithfulness, and LLM-as-a-Judge forming the core evaluation metrics.

For standard QA tasks, the NQ and FEVER datasets should primarily employ F1 and Exact-Match to evaluate factual QA capability: NQ emphasizes accuracy in open-domain QA, while FEVER specifically tests fact verification ability. For multi-hop reasoning scenarios, F1 and Faithfulness should be measured simultaneously on HotpotQA—the former assessing answer accuracy, the latter verifying whether the model over-relies on single evidence sources. In domain-specific evaluations, the MIRAGE medical dataset should prioritize Faithfulness, combined with expert review to validate the reliability of medical advice; the DomainRAG educational consulting dataset should adopt a combination of F1 and LLM-as-a-Judge to assess both factual accuracy and pedagogical appropriateness. For knowledge attribution capability, fine-grained analysis should be conducted using tools such as ALCE, QAFactEval, FActScore, FACTS, GaRAGe, and RAGTruth, comprehensively evaluating interpretability and factual consistency of generated content—from citation quality to span-level alignment [6][7][8][11][12][15]. For timeliness evaluation, the ReEval RealTimeQA dataset should use the ratio of Temporal F1 to Faithfulness to quantify the impact of knowledge updates on generation quality. In multimodal generation evaluation, the RAG-Check and T²-RAGBench datasets should employ multidimensional LLM-as-a-Judge scoring, with particular emphasis on semantic consistency between generated content and retrieved images [24].

End-to-end system efficacy evaluation focuses on the integrated performance of the full RAG pipeline, with Answer Accuracy, Consistency, and Coverage serving as key metrics.

In comprehensive evaluation scenarios, the CRAG and RAGBench datasets should jointly employ Answer Accuracy and Coverage: the former measures final answer correctness, while the latter evaluates information completeness; their combination reveals the system’s balancing capability across diverse domain tasks [17][18]. For knowledge manipulation capability testing, CREATE/READ/UPDATE/DELETE tasks should be designed on the CRUD-RAG and UHGE-mal datasets, measuring Answer Accuracy for each operation type—with particular attention to system response latency following knowledge updates in UPDATE tasks. In domain-specific validation, the DomainRAG university admissions consulting dataset should combine Answer Accuracy with Consistency analysis: the former evaluates answer correctness, while the latter tests output stability across similar consultation queries—critical for educational advisory services. For long-text generation tasks, Coverage and Key Point Recall should be prioritized on the LongBench and RULER datasets to evaluate the system’s capacity for long-context comprehension and information integration [20][21]. Additionally, the BEIR extended set should combine Answer Accuracy with MRR to analyze the correlation between retrieval quality and final answer accuracy, providing clear guidance for system optimization.

Dynamic adaptability evaluates system stability and adaptability in real-world environments, with Robustness to Noise, Temporal Freshness, and Privacy Preservation as core metrics.

For noise robustness testing, systematically inject varying proportions of misleading information into the RGB synthetic dataset and UHGE-mal, measuring Robustness to Noise (defined as the ratio of misleading information proportion to decline in answer accuracy), with particular focus on low-noise regimes (5\%–10\%), which better approximate real-world scenarios. For timeliness evaluation, FreshQA and RealTimeQA datasets should conduct pre- and post-knowledge-update comparative tests, quantifying system responsiveness to knowledge changes via the Temporal Freshness metric (defined as the recovery speed of answer accuracy after knowledge updates). The RAMDocs dataset, designed specifically for knowledge conflict scenarios, should be evaluated using Answer Accuracy and Consistency. For privacy preservation evaluation, Diverse And Private Synthetic Datasets should include known sensitive information fragments, assessing data protection capability via the Privacy Preservation metric (defined as the proportion of sensitive information not leaked). For long-context adaptability, the CRUX dataset should measure the ratio of Context Integrity Score to Redundancy Rate to evaluate the system’s ability to preserve critical content while processing redundant information. Finally, in interactive scenario evaluation, the RAG Without the Lag dataset should employ the Debugging Efficiency metric (defined as time/steps from user feedback to answer correction) to measure system adaptability in real user interactions. For comprehensive evaluation of multimodal RAG systems, VisDoMBench and T²-RAGBench provide systematic cross-modal testing environments and should be evaluated by integrating multimodal retrieval and generation metrics.

\subsubsection{Challenges and Future Directions}
Although the overall research trajectory of RAG evaluation frameworks is becoming increasingly clear, several critical issues still require in-depth investigation.

Current evaluations often conflate accuracy with faithfulness, leading to misjudgments of system capabilities. Research shows that in high-stakes domains such as medical QA, models may generate answers that appear correct but lack evidential support—highlighting the necessity of decoupling these two dimensions for independent assessment. Methods such as ALCE and FActScore have begun advancing this direction by decomposing outputs into fine-grained factual units to enable sentence-level verification of generated content [6][8]. However, discrepancies in the definition and computation of groundedness across different benchmarks undermine the comparability of evaluation results. Frameworks such as QAFactEval and RAGTruth propose evidence-chain-based evaluation paradigms, aiming to establish more consistent standards [11][12]. Future work should further promote the standardization of span-level support evaluation—particularly in high-risk applications such as healthcare and law—by mandating the joint reporting of accuracy and faithfulness metrics to ensure comprehensiveness and reliability of evaluation outcomes.

Most existing evaluations rely on static knowledge bases and closed-set questions, failing to reflect the dynamic evolution of knowledge in real-world scenarios. Benchmarks such as FreshQA and RAMDocs reveal system vulnerabilities when confronted with knowledge updates and information conflicts [13][14]. Experiments indicate that model answer accuracy may significantly degrade when knowledge base content changes—degradations that static benchmarks cannot capture. Stress-testing suites like RGB, by injecting noise and misleading information, further expose insufficient robustness in complex environments [19]. Collectively, these studies demonstrate that evaluations based solely on static datasets tend to overestimate real-world system performance. Future evaluations must systematically incorporate dynamic scenarios such as knowledge updates, conflict resolution, and timeliness responsiveness, constructing test environments that simulate real knowledge lifecycles to more accurately measure practical system capabilities.

Evaluations using LongBench and RULER show that model performance on complex tasks degrades as context length increases—particularly evident in scenarios requiring deep reasoning [20][21]. This suggests that blindly expanding context windows is not a sustainable strategy for performance improvement. In contrast, high-quality external retrieval combined with effective reranking mechanisms can significantly enhance answer quality while controlling computational costs. Methods such as MultiHop-RAG validate the advantages of precise retrieval in multi-hop reasoning tasks [25]. These findings advocate shifting evaluation focus from merely expanding context length toward optimizing retrieval efficiency and knowledge integration. Future research should place greater emphasis on quantifying the relationship between retrieval quality and system performance, exploring architectural designs that achieve optimal performance under constrained resources.

Although LLM-as-a-Judge improves evaluation efficiency, the stability of its judgments and cross-domain generalization capability remain questionable. Scoring biases may exist across different evaluator models, and performance may be unreliable in specialized domains or on adversarial samples. The PPI method proposed by ARES introduces confidence intervals to provide uncertainty quantification for evaluation scores, facilitating more cautious interpretation of results [10]. This approach helps identify high-risk judgments in evaluations, avoiding misjudgments caused by evaluator bias. Future evaluations should adopt hybrid review strategies—combining the efficiency of automated evaluation with the reliability of human assessment—and perform sample-based calibration on critical instances to enhance overall evaluation credibility.

Furthermore, performance and efficiency are critical considerations for external knowledge retrieval systems in real production environments. Practical deployment of external memory systems requires balancing retrieval quality, response latency, computational cost, and knowledge update frequency. Platforms such as CRAG and RAGBench, which support cross-domain and interpretable evaluation, facilitate analysis of system behavior under different configurations [17][18]. Studies show that parameters such as top-k values and context length significantly impact system performance, often exhibiting diminishing returns. Knowledge base update frequency must also be optimized based on the trade-off between timeliness gains and computational overhead. In long-context and multimodal scenarios, benchmarks such as LongBench, RULER, MM-Needle, VisDoMBench, and T²-RAGBench provide testbeds for evaluating the cost-effectiveness of different technical approaches [20][21][22][23][24]. Future evaluations should explicitly incorporate engineering constraints, establishing comprehensive frameworks that balance performance with sustainability, thereby guiding the design and optimization of RAG systems in real-world business applications.

\subsection{Evaluation of Procedural/Episodic Memory} 
This section focuses on LLM procedural/episodic memory under cross-turn and cross-session conditions: namely, the quality of long-term writes, the traceability and sufficiency of replay, and the inhibit/abstention mechanisms when evidence is insufficient, knowledge is stale, or sources conflict. We adhere to the three operating regimes of §4.1 to separate model capability from information availability, and organize the measurement dimensions via the E-MARS+ framework: Encode/Memorize \& temporal anchoring (P1), Replay \& attribution (P2), Suppress/Freshness/Conflict (P3), Long-horizon stability (P4), and Cost efficiency (P5). We then specify evaluation objects and research questions in §4.5.1, instantiate operating regimes in §4.5.2, define metrics in §4.5.3 and map them to public benchmarks in §4.5.4, and finally discuss threats to validity and key takeaways in §§4.5.5–4.5.6. Our aim is not to compress outcomes into a single score, but to provide a layered, interpretable, and engineering-usable reporting paradigm. \citep{maharana2024evaluatinglongtermconversationalmemory,wu2025longmemevalbenchmarkingchatassistants,ge2025tremuneurosymbolictemporalreasoning}

\subsubsection{Problem Definition and Evaluation Objects}Unlike one-off long-context understanding, procedural \& episodic memory emphasizes cross-turn/session write–replay–inhibit mechanisms: how the system deposits key information as long-term memory, retrieves it at appropriate times, and abstains when knowledge is outdated or evidence is insufficient. Existing work shows that merely enlarging the global context is insufficient for robustness and auditability over long horizons (LoCoMo, LongMemEval); eventifying/proceduralizing history and driving inference via structured replay yields significant gains (TReMu). \citep{maharana2024evaluatinglongtermconversationalmemory,wu2025longmemevalbenchmarkingchatassistants,ge2025tremuneurosymbolictemporalreasoning} Accordingly, we expand the evaluation objects into four complementary families:

A. Episodic events and timelines: extraction of who–what–where–when–state, order restoration, and recency/state awareness (Episodic Memories Benchmark, SORT/Book-SORT); \citep{huet2025episodicmemoriesgenerationevaluation,pink2024testing}

B. Procedural processes and strategies: multi-step workflows, tool selection, and self-reflection/retry (LoCoMo, LongMemEval, Reflexion, Voyager) ; \citep{maharana2024evaluatinglongtermconversationalmemory,wu2025longmemevalbenchmarkingchatassistants,shinn2023reflexionlanguageagentsverbal,wang2023voyageropenendedembodiedagent}

C. Identity/preferences and constraints: identity and preference retention, compliance/safety thresholds, and refusal strategies (MemGPT, Generative Agents, HippoRAG, AriGraph); \citep{packer2024memgptllmsoperatingsystems,park2023generativeagentsinteractivesimulacra,gutierrez2025hipporagneurobiologicallyinspiredlongterm,anokhin2025arigraphlearningknowledgegraph}

D. Conflict resolution and inhibition: selection under old/new or cross-source conflicts, detection and inhibition of outdated evidence (RAMDocs, FreshLLMs/FreshQA) . \citep{wang2025retrievalaugmentedgenerationconflictingevidence,vu2023freshllmsrefreshinglargelanguage,chen2021datasetansweringtimesensitivequestions}

\subsubsection{Operating Regimes}To separate capability from information availability, we strictly follow the three regimes of §4.1 and instantiate them for episodic memory scenarios:

Parametric-Only (PO): disable retrieval and replay to form a closed-book baseline of "bare parametric memory," observing the lower bound of recall and long-horizon degradation without external injection.

Offline-Retrieval: freeze the index or session memory store; in this section we split the pipeline into two:

Long-Context (LC-off): direct reading of raw history, diagnosing length/position sensitivity and the upper bound of direct reading (RULER, LongBench, LV-Eval) ; \citep{hsieh2024rulerwhatsrealcontext,bai2024longbenchbilingualmultitaskbenchmark,yuan2024lvevalbalancedlongcontextbenchmark}

Structured-Replay (SR-off): structure history into events/processes/preferences and replay on demand, facilitating the decoupling of write–replay–attribution (protocolized in TReMu) . \citep{ge2025tremuneurosymbolictemporalreasoning}

Online-Retrieval: connect to dynamic knowledge sources or updatable memory stores, focusing on freshness, outdated answer rate, and consistency under conflict (FreshQA, LongMemEval online scenarios). \citep{vu2023freshllmsrefreshinglargelanguage,wu2025longmemevalbenchmarkingchatassistants}

Unless otherwise stated, we report PO, LC-off, and SR-off in parallel for key slices; when timeliness/conflict is involved, we add OR/SR-on results, preserving comparability with §4.1.
\subsubsection{Metric System: The E-MARS+ Five-Panel}To balance measurability and interpretability in procedural/episodic settings, we design E-MARS+ (Encode/Memorize—Attribute/Replay—Suppress/Freshness/Conflict—Stability—Cost) to provide layered measurement across the four evaluation families(Table~\ref{tab:emars-panels}:\nameref{tab:emars-panels}). Unless otherwise noted, metrics are macro-averaged over samples with 95\% confidence intervals; cross-regime/model comparisons use paired permutation or bootstrap tests with Holm/FDR multiple-comparison correction. Adjudicators may be human, NLI consistency models, or LLM-as-a-judge . \citep{zheng2023judgingllmasajudgemtbenchchatbot,agarwal2024faithfulnessvsplausibilityunreliability}

\begin{table*}[htbp]
\centering
\scriptsize
\caption{E-MARS+ Five-Panel Overview (for Procedural/Episodic Memory Evaluation)}
\label{tab:emars-panels}
\resizebox{\textwidth}{!}{%
\renewcommand{\arraystretch}{1.25}
\begin{tabularx}{\textwidth}{l|X|X|X|X|X}
\toprule
\textbf{Panel} & \textbf{Concept \& Goal} & \textbf{Main Metrics (Symbols)} & \textbf{Computation Points (1-line definition)} & \textbf{Aggregation / Statistics} & \textbf{Typical Clips / Benchmarks (Examples)} \\
\midrule
P1 Write \& Temporal Anchoring & 
Write information correctly into long-term memory and anchor time &
Event-F1; Procedure-F1; \textbf{TAE} (Temporal Anchoring Error) &
Event / procedure extraction F1; absolute error of temporal anchoring (window-normalized) &
Macro average; 95\% CI &
Episodic Memories, SORT / Book-SORT; LoCoMo, LongMemEval; TReMu \\
\midrule
P2 Replay \& Attribution &
Whether replay items are sufficient and traceably support the answer &
\textbf{RSF} (Replay-Supported Fraction); \textbf{ASR@k}; \textbf{UCR} (Unsupported Claim Rate); Step-Order Acc; Tool-Attribution &
Proportion of answers traceably supported by replayed evidence; Top-k replay coverage; not contradicted by replay items; step-order accuracy &
Macro average; span-level precision/recall; paired verification &
TReMu (temporal replay), Episodic, SORT; LoCoMo / LongMemEval (process/tools) \\
\midrule
P3 Inhibition / Freshness / Conflict &
Consistency when choosing answers under unanswerable, outdated, or conflicting evidence &
Abstention@Unans.; Freshness-Hit / Out-of-Date-Use; \textbf{CCR} (Conflict-Consistency Rate); \textbf{AURC} &
Correct refusal rate for unanswerable cases; freshness-hit and out-of-date usage rate; consistency under conflict sources &
Conditional slices: unanswerable / freshness / conflict; risk-coverage curves &
LongMemEval (unanswerable consistency), FreshQA (freshness), RAMDocs (conflict) \\
\midrule
P4 Long-Horizon Robustness &
Degradation and interference resistance across turns / sessions &
Slope@T (dAcc/dT); h\textsubscript{1/2} (half-life); Spurious-Replay Rate; Noise-Robust Acc &
Accuracy decay slope over turns; half-life without retraining; error rate of spurious replay; average accuracy under perturbations &
Curves (slopes); span-binned accuracy reports &
LoCoMo, LongMemEval; dynamic conversation / task evaluations; Beyond-Prompts \\
\midrule
P5 Cost \& Efficiency &
Latency / resource / throughput trade-offs under equal performance &
Latency/turn; Tokens/turn (recall+generation); Mem-Footprint; Throughput; Cost@Target &
End-to-end latency; decoding + retrieval efficiency; per-quality unit cost &
Frontier plots of performance vs. cost; sensitivity curves &
LC-off / SR-off / OR/SR-on comparisons; StreamingLLM; KV-cache compression; memory-sensitive benchmarks \\
\bottomrule
\end{tabularx}
}
\end{table*}

\subsubsection{Benchmark and Task Mapping}For reproducibility and apples-to-apples comparison, we align E-MARS+ panels (P1–P5) with representative datasets and specify operating regimes and primary adjudication modes. Table~\ref{tab:emars-panel2}:\nameref{tab:emars-panel2} summarizes the evaluation objects, representative benchmarks, primary metrics, default regimes, and notes. \citep{huet2025episodicmemoriesgenerationevaluation,pink2024testing,maharana2024evaluatinglongtermconversationalmemory,wu2025longmemevalbenchmarkingchatassistants,shinn2023reflexionlanguageagentsverbal,wang2023voyageropenendedembodiedagent,packer2024memgptllmsoperatingsystems,park2023generativeagentsinteractivesimulacra,gutierrez2025hipporagneurobiologicallyinspiredlongterm,anokhin2025arigraphlearningknowledgegraph,wang2025retrievalaugmentedgenerationconflictingevidence,vu2023freshllmsrefreshinglargelanguage,hsieh2024rulerwhatsrealcontext,bai2024longbenchbilingualmultitaskbenchmark,yuan2024lvevalbalancedlongcontextbenchmark,liu2023lostmiddlelanguagemodels,hsieh2024middlecalibratingpositionalattention,castillobolado2024promptsdynamicconversationalbenchmarking,xiao2024efficientstreaminglanguagemodels}

% ---------- Table 4.5-2 ----------
% 让 tabularx 的 X 列左对齐并自动换行
\renewcommand{\tabularxcolumn}[1]{>{\RaggedRight\arraybackslash}p{#1}}
\begin{table*}[htbp]
\centering
\footnotesize
\caption{E-MARS+ Panels: Overview of Dataset–Task–Metric–Operating-Regime Alignment}
\label{tab:emars-panel2}
\setlength{\tabcolsep}{4pt}
\renewcommand{\arraystretch}{1.14}
\begin{tabularx}{\textwidth}{|l|X|X|X|X|}
\hline
\textbf{Evaluation Target} &
\textbf{Representative Datasets / Tasks} &
\textbf{Key Metrics (mapped to E\,-MARS+)} &
\textbf{Run Settings (match \S4.1)} &
\textbf{Notes} \\
\hline
\textbf{Narrative Events / Timelines} &
Episodic Memories Benchmark; SORT / Book\,-SORT &
Event\,-F1, TAE (\textbf{P1}); RSF, Step\,-Order Acc (\textbf{P2}) &
\textit{LC-off} (length/position diagnosis); \textit{SR-off} (structured replay off) &
Support span\,-level NLI or LLM\,-judge for segment verification. \\
\hline
\textbf{Procedural Processes / Strategies} &
LoCoMo; LongMemEval process clips; Reflexion; Voyager &
Procedure\,-F1 (\textbf{P1}); Tool\,-Attribution, RSF / ASR@k (\textbf{P2}); Slope@T, $h_{1/2}$ (\textbf{P4}) &
\textit{PO} baseline + \textit{LC-off} / \textit{SR-off}; enable \textit{OR}/\textit{SR-on} when needed &
Also observe long\,-horizon decay (\textbf{P4}) and synchronous/tool attribution (\textbf{P2}). \\
\hline
\textbf{Preferences and Policy Constraints} &
LongMemEval\textendash\allowbreak consistency/\allowbreak refusal; 
MemGPT; Generative Agents; HippoRAG; AriGraph &
Preference\,-Retention, Policy\,-Violation Rate (\textbf{P1/P2}); Abstention@Unanswerable, AURC (\textbf{P3}) &
\textit{SR-off}; turn on \textit{OR}/\textit{SR-on} when freshness is required &
Use for personal/preference safeguarding and policy\,-compliant selective answering. \\
\hline
\textbf{Conflict Resolution and Inhibition} &
RAMDocs (conflict/\allowbreak contradiction/\allowbreak error provenance); 
FreshQA\slash\allowbreak FreshLLMs &
CCR, Freshness\,-Hit, Out\,-of\,-Date\,-Use, Abstention@Unanswerable, AURC (\textbf{P3}) &
\textit{OR} or \textit{SR-on} (dynamic sources), or \textit{SR-off} as baseline &
Record snapshot dates/update times; prioritize conflict slices in reporting. \\
\hline
\textbf{Length \& Position Control (cross targets)} &
RULER; LongBench; LV\,-Eval; mechanism diagnostics: Lost/Found\,-in\,-the\,-Middle &
Length/position performance curves; \textbf{P4} (decay/robustness); \textbf{P5} (latency/cost) &
Primarily \textit{LC-off}, compared against \textit{SR-off} &
First report performance–length/position sensitivity curves, then show model deltas. \\
\hline
\end{tabularx}
\end{table*}

To mitigate threats to internal/external validity, we follow these controls:

Adjudicator bias and uncertainty (internal validity). When using NLI or LLM-as-a-judge for support decisions, employ dual adjudication with light human spot checks; report basic agreement (e.g., Cohen’s $\kappa$) and 95\% CIs, avoiding single adjudicator conclusions . \citep{zheng2023judgingllmasajudgemtbenchchatbot,agarwal2024faithfulnessvsplausibilityunreliability}

Length/position bias and truncation policy (internal validity). Under LC-off, state the context window, positional encoding/truncation rules; where needed, use Found-in-the-Middle calibration as a control. \citep{hsieh2024middlecalibratingpositionalattention,liu2023lostmiddlelanguagemodels}

Replay-entry construction error (internal validity). Under SR-off/-on, extraction/aggregation errors during the write phase can be systematically amplified, manifesting as RSF decreases and UCR increases.

Temporal governance and potential leakage (external validity). In online/dynamic scenarios, disclose the time windows and snapshots of training/index/test to avoid mistaking freshness differences for capability differences; give qualitative accounts of near-duplicates/contamination for public benchmarks. \citep{lee2022deduplicatingtrainingdatamakes,kandpal2022deduplicatingtrainingdatamitigates,carlini2019secretsharerevaluatingtesting,carlini2023quantifyingmemorizationneurallanguage}

Consistency of comparison protocol (external validity). For cross-regime comparisons, fix retriever/reranker types and key hyperparameters; for cross-model comparisons, use identical random seeds/decoding strategies and budget caps. \citep{saadfalcon2024aresautomatedevaluationframework,yang2024cragcomprehensiverag}

\subsubsection{Summary}Within the three operating regimes of §4.1, we construct a multi-object evaluation paradigm covering events—processes/strategies—preferences/constraints—conflict/inhibition, and unify writing, replay, inhibition, long-horizon stability, and engineering cost into an auditable metric set via the E-MARS+ five panels. Mapping to benchmarks such as LoCoMo, LongMemEval, Episodic, SORT/Book-SORT, TReMu, RAMDocs, and FreshQA shows that, in long-term interaction, structured replay is more portable and cost-effective than ultra-long-context direct reading; in dynamic scenarios, freshness and consistency under conflict should be measured explicitly, and selective answering should be used to achieve a controllable trade-off between risk and coverage. This framework supports literature verification and offers actionable diagnostics and optimization cues for engineering deployment. \citep{maharana2024evaluatinglongtermconversationalmemory,wu2025longmemevalbenchmarkingchatassistants,huet2025episodicmemoriesgenerationevaluation,pink2024testing,ge2025tremuneurosymbolictemporalreasoning,wang2025retrievalaugmentedgenerationconflictingevidence,vu2023freshllmsrefreshinglargelanguage,chen2021datasetansweringtimesensitivequestions}

\FloatBarrier

\section{Strategies for Forgetting and Knowledge Updating in LLM Memory}
\subsection{Problem Definition and Objectives} {This chapter focuses on the dynamic management of memory in large language models (LLMs): performing knowledge updating and memory forgetting over different substrates (parametric, contextual, external, procedural/episodic) across the full life cycle, and achieving system-level governance under auditable and reversible constraints. We first articulate the limitations of static memory paradigms and the necessity of dynamic management, then define the objects of study and scope of operations, and finally present an objectives–metrics system aligned with evaluation–deployment.}
\subsubsection{Limitations of Static Memory and the Need for Dynamic Management} Parametric static memory centered on one-shot pretraining exhibits four systematic deficiencies in real-world scenarios:
(1) Knowledge staleness: sluggish response to facts that change over time (positions, regulations, data, etc.); under the PO setting, freshness is visibly inferior to systems with external evidence \citep{chen2021datasetansweringtimesensitivequestions,Dhingra_2022,vu2023freshllmsrefreshinglargelanguage,lewis2021retrievalaugmentedgenerationknowledgeintensivenlp,petroni2021kiltbenchmarkknowledgeintensive} .
(2) Insufficient factuality and traceability: hallucinations are more likely without evidence alignment, making third-party audits difficult \citep{maynez2020faithfulnessfactualityabstractivesummarization,kryściński2019evaluatingfactualconsistencyabstractive,fabbri2022qafactevalimprovedqabasedfactual,gao2023enablinglargelanguagemodels,li2024towards,honovich2022truereevaluatingfactualconsistency,jacovi2025factsgroundingleaderboardbenchmarking} .
(3) Privacy and compliance risks: training samples may be memorized unintentionally and reproduced under specific prompts; highly repetitive/noisy data amplify membership inference and data leakage risks \citep{carlini2019secretsharerevaluatingtesting,carlini2021extracting,duan2024membershipinferenceattackswork,kandpal2022deduplicatingtrainingdatamitigates,lee2022deduplicatingtrainingdatamakes,carlini2023quantifyingmemorizationneurallanguage} .
(4) Working-memory bottlenecks: enlarging the context does not necessarily stabilize evidence utilization, especially for mid-span evidence and ultra-long sequences \citep{liu2023lostmiddlelanguagemodels,hsieh2024middlecalibratingpositionalattention,hsieh2024rulerwhatsrealcontext,bai2024longbenchbilingualmultitaskbenchmark,yen2025helmetevaluatelongcontextlanguage,zhang2024inftybenchextendinglongcontext} .

Therefore, static pretraining alone cannot simultaneously satisfy timeliness, compliance, and verifiability; a dynamic memory management framework (updating, forgetting, and governance) geared to practical applications is urgently needed.
\subsubsection{Scope and Objects}We partition LLM memory substrates into: (i) parametric memory (long-term knowledge in weights); (ii) contextual memory (activations and KV caches at inference); (iii) external memory (explicit evidence injected via retrieval or toolchains); and (iv) procedural/episodic memory (events and trajectories across turns). Around these substrates, we discuss two core operations: knowledge updating (introducing or correcting knowledge while preserving existing capabilities) and memory forgetting (suppressing, removing, or externalizing specific knowledge for compliance or safety). Evaluation and reporting follow Chapter 4: use a unified timeline (aligned timestamps for training corpora, indices, and test sets) and report in parallel under closed-book (PO) / offline RAG / online retrieval regimes \citep{geva2021transformerfeedforwardlayerskeyvalue,Pan_2024,lewis2021retrievalaugmentedgenerationknowledgeintensivenlp,petroni2021kiltbenchmarkknowledgeintensive} .

Contributions of this chapter:
C1 Unified perspective: Propose DMM-Gov (Dynamic Memory Management \& Governance), discussing update–forget–audit uniformly across the four memory substrates (parametric/context/external/procedural), with timeline-aligned evaluation and reporting under PO/offline-RAG/online-retrieval regimes \citep{petroni2021kiltbenchmarkknowledgeintensive,thakur2021beirheterogenousbenchmarkzeroshot} .
C2 Executable thresholding: Turn updating and forgetting into pre-registrable launch thresholds (ESR/Locality/Drawdown/Generalization; Citation Coverage/Unsupported Claim; four-dimensional forgetting acceptance), with default thresholds and rollback strategies \citep{fabbri2022qafactevalimprovedqabasedfactual,min2023factscorefinegrainedatomicevaluation,li2024towards,jacovi2025factsgroundingleaderboardbenchmarking} .
C3 Sequential/lifelong robustness: Propose composite editing routes—"small steps–spacing–sharding" and alignment-based/time-window/event-level (LTE/AToKE/ELKEN + AlphaEdit/WISE/SERAC)—and compose LEME/ConflictEdit/EVOKE as mandatory regressions for long-form consistency, conflict reversibility, and out-of-template generalization \citep{jiang2024learningeditaligningllms,yin2023historymatterstemporalknowledge,peng2024eventlevelknowledgeediting,fang2025alphaeditnullspaceconstrainedknowledge,wang2024wiserethinkingknowledgememory,mitchell2022memorybasedmodeleditingscale,rosati2024longformevaluationmodelediting,hsueh2024editingmindgiantsindepth,hu2023evoke} .
C4 Audit certificate: Standardize "edit/forgetting certificates," requiring versioning—evidence—time window—revocability for third-party verifiable audits (aligned with NIST/OWASP) \citep{925786,owasp2023llmtop10} .
\subsubsection{Objectives and Challenges: From Principles to Verifiable Metrics}Dynamic management must strike an auditable balance among three objective sets, validated by clear, repeatable metrics.

(A) Correctness and timeliness: The challenge lies in rapid knowledge changes and the coexistence of errors and newly updated facts. Metrics include Editing Success Rate (ESR); freshness hit rate / outdated-answer rate (TimeQA, FreshLLMs); share of evidence-backed answers and citation coverage (FACTS, ALCE, FActScore, QAFactEval) \citep{chen2021datasetansweringtimesensitivequestions,Dhingra_2022,vu2023freshllmsrefreshinglargelanguage,jacovi2025factsgroundingleaderboardbenchmarking,li2024towards,min2023factscorefinegrainedatomicevaluation,fabbri2022qafactevalimprovedqabasedfactual} .

(B) Controllability and auditability: Point edits can impinge on unrelated capabilities and lack an evidence chain. Metrics include Locality (neighborhood stability), Drawdown (general capability regression), Generalization (synonymic/compositional/multi-hop), and versioning \& operation logs (the "edit/forgetting certificate") \citep{mitchell2022fastmodeleditingscale,meng2023locatingeditingfactualassociations,mitchell2022memorybasedmodeleditingscale,gu2024modeleditingharmsgeneral} .

(C) Privacy and safety: sample exposure, dangerous capabilities/biases, adversarial resurgence. Metrics include verbatim regurgitation and extraction attack success rates, membership inference AUC (supporting signal only) \citep{carlini2021extracting,duan2024membershipinferenceattackswork} ; dangerous-capability compliance rate (WMDP) \citep{li2024wmdpbenchmarkmeasuringreducing} ; robustness to adversarial resurgence (RWKU, MUSE adversarial settings) \citep{jin2024rwkubenchmarkingrealworldknowledge,shi2024musemachineunlearningsixway} .

For edit controllability, beyond standard Locality/Drawdown/Generalization, explicitly include conflict consistency and transfer generalization: long-form narrative and cross-paragraph consistency can be validated with LEME; conflict and reversibility pitfalls can be stress-tested with ConflictEdit/Pitfalls; EVOKE diagnoses "overfitting to edit templates" under out-of-template/out-of-domain settings \citep{rosati2024longformevaluationmodelediting,hsueh2024editingmindgiantsindepth,hu2023evoke} . Temporal consistency and valid time windows should be audited via the time-aware AToKE scheme and metrics \citep{yin2023historymatterstemporalknowledge} .

Below, we discuss knowledge updating (§5.2, by parametric/context/external/procedural substrates), memory forgetting (§5.3, optimization/representation/system-level strategies with unified acceptance), and selection \& governance from research to release (§5.4).
\subsection{Knowledge Updating}This section presents a unified framework for LLM memory updating, aiming to introduce/correct knowledge and control side effects without sacrificing general capability or auditability. We detail four routes by substrate: parametric memory (DAPT/TAPT, PEFT, model editing) for precise closed-book rewrites; contextual memory focusing on read/write at inference and positional calibration; external memory (RAG/KG) offering hot updates and evidence alignment; and procedural/episodic memory for cross-session consistency. The selection follows a "rule of thirds": prioritize external memory for high-timeliness/traceability; use model editing for small, unambiguous fact corrections (with robustness constraints in sequential/lifelong settings); and use DAPT/PEFT with versioned releases for broad yet relatively stable scope. Throughout, we evaluate parametric rewrites via ESR/Locality/Drawdown/Generalization, assess freshness and source attribution for external/contextual routes, and report under timeline alignment (train–index–test) with gray rollout—monitoring—rollback.
\subsubsection{Parametric Memory: Continued Pretraining, Parameter-Efficient Finetuning, and Model Editing}When updates involve broad and relatively stable knowledge, two-stage self-supervised Domain/Task-Adaptive Pretraining (DAPT/TAPT) can refresh register and factual distributions while preserving base capabilities \citep{gururangan2020dontstoppretrainingadapt} . When updates span multiple domains in parallel and require rapid rollback, parameter-efficient finetuning (PEFT) (e.g., Adapters, Prefix/Prompt Tuning, IA³) adjusts only small add-on/scaling parameters on a frozen backbone, naturally supporting versioning and revocation \citep{houlsby2019parameterefficienttransferlearningnlp,lester2021powerscaleparameterefficientprompt,li2021prefixtuningoptimizingcontinuousprompts} . When needs converge to a small number of explicit fact fixes, model editing (ROME/MEND/MEMIT/SERAC) performs targeted rewrites at causally related internal sites, enabling fast, addressable weight updates \citep{meng2023locatingeditingfactualassociations,mitchell2022fastmodeleditingscale,meng2023masseditingmemorytransformer,mitchell2022memorybasedmodeleditingscale} .

Sequential/lifelong updates tend to induce edit interference and capability regression. AlphaEdit reduces neighborhood disturbance via null-space constraints; WISE isolates edited knowledge through main/side memory routing, supporting shard merging and rollback; SEAL explores an online, self-editing framework. Engineering-wise, we recommend "small steps–spacing–sharding," with regression against sequential-edit stress cases; for high-risk edits, first deploy SERAC externalized overlay for gray validation before merging into model parameters \citep{fang2025alphaeditnullspaceconstrainedknowledge,wang2024wiserethinkingknowledgememory,mitchell2022memorybasedmodeleditingscale} .

Beyond structural constraints (null space/routing), alignment-based editing offers a complementary path: LTE marries knowledge editing with instruction/preference alignment, improving ESR and Locality while reducing regression risk. For time-valid facts, AToKE parameterizes edit targets as timestamps/time windows, enabling auditable switching of old/new knowledge compatible with rollback. To address causal inconsistency from single-point edits, ELKEN proposes event-level editing—jointly updating and consistency-checking multiple entities and constraints—to mitigate cascading side effects of "point fixes, global mismatches." On evaluation, LEME long-form regression tests the persistence and consistency of edited knowledge across long contexts and narrative chains; ConflictEdit/Pitfalls stress conflict/reversibility/template transfer; EVOKE reveals template overfitting risks and recommends out-of-template/out-of-domain/long-horizon acceptance \citep{jiang2024learningeditaligningllms,yin2023historymatterstemporalknowledge,peng2024eventlevelknowledgeediting,rosati2024longformevaluationmodelediting,hsueh2024editingmindgiantsindepth,hu2023evoke} .

Process specification: construct target / neighborhood / retention three-way slices; deploy DAPT/PEFT as "adapters + old backbone" in gray rollout; for editing, log an "edit ledger—version ID—reverse-edit points," start with low-traffic gray and track ESR/Locality/Drawdown. Upon spillover, rollback or switch to SERAC/RAG overlays.

Acceptance and regression add-ons. Beyond the "target/neighborhood/retention" slices, add three mandatory regressions:
(i) Long-form consistency: include LEME long-form regression \citep{rosati2024longformevaluationmodelediting} ;
(ii) Conflict/reversibility: include ConflictEdit/Pitfalls stress sets \citep{hsueh2024editingmindgiantsindepth} ;
(iii) Out-of-template/out-of-domain: use EVOKE-style settings to prevent overfitting to public formats such as CounterFact/zsRE \citep{hu2023evoke} .
Time-sensitive edits must record effective/expiry times and align testing with AToKE \citep{yin2023historymatterstemporalknowledge} ; for composite events, perform event-level consistency per ELKEN \citep{peng2024eventlevelknowledgeediting} .
\subsubsection{Contextual Memory: Read/Write Strategies and Positional Calibration}Problem and motivation. In long documents and multi-turn reasoning, contextual memory suffers simultaneously from positional bias and compute/memory budgets: mid-span under-response ("lost-in-the-middle") weakens evidence use, and ultra-long sequences inflate VRAM and latency, limiting feasible window sizes \citep{liu2023lostmiddlelanguagemodels} . We therefore systematize three routes: dependency extension, streaming/KV management, and positional calibration.

Method taxonomy. (i) Dependency extension: extend effective dependency chains by introducing reusable states and relative positional encoding; representative methods include Transformer-XL and Compressive Transformer; the more recent Infini-attention retains long-range information at controlled overhead, enabling feasible inference with quasi-"infinite" context \citep{dai2019transformerxlattentivelanguagemodels,rae2019compressivetransformerslongrangesequence,munkhdalai2024leavecontextbehindefficient} . (ii) Streaming inference and KV management: maintain long-sequence stability via StreamingLLM-style online refresh/eviction, combined with KV compression and cross-layer clustering/sharing to reduce VRAM/latency (e.g., PoD, EPIC, KVzip) \citep{xiao2024efficientstreaminglanguagemodels,ma2024compressing,hu2025epicefficientpositionindependentcaching,kim2025kvzipqueryagnostickvcache,hu2025efficientlongcontextllminference} . (iii) Positional calibration: to address low utilization of mid-span evidence, apply reweighting and anchor instrumentation (e.g., Found-in-the-Middle) to markedly improve mid-span usability, in mutual corroboration with lost-in-the-middle diagnostics \citep{hsieh2024middlecalibratingpositionalattention,liu2023lostmiddlelanguagemodels,press2022trainshorttestlong,chen2023extendingcontextwindowlarge,su2023roformerenhancedtransformerrotary} .

Implementation recommendations. Follow a light-to-heavy three-step sequence: first perform structured reordering (paragraph/evidence layout, summarize–replay, and anchor prompting) to improve key-information visibility without extra inference cost; second apply positional reweighting/anchor instrumentation to improve mid-span hits; third, expand the window with paired KV management when latency/VRAM budgets allow. To avoid attention dilution and cost blowups from blind window expansion, design context extension to complement retrieval augmentation (RAG) and enlarge windows only when net gains are demonstrated.

Evaluation protocol. Offline, report position–accuracy curves and performance-vs-length slopes, and quantify mid-span hit rate; online, jointly monitor latency/throughput and long-sequence stability. Dependency-extension routes use Transformer-XL/Compressive as baselines \citep{dai2019transformerxlattentivelanguagemodels,rae2019compressivetransformerslongrangesequence} ; streaming/KV-compression routes characterize trade-offs via VRAM, throughput, and accuracy retention \citep{xiao2024efficientstreaminglanguagemodels,ma2024compressing,hu2025epicefficientpositionindependentcaching,kim2025kvzipqueryagnostickvcache} ; positional calibration reports mid-span gains and non-regression at head/tail positions \citep{hsieh2024middlecalibratingpositionalattention} . Long-context unified protocols may follow RULER/HELMET/LongBench \citep{hsieh2024rulerwhatsrealcontext,yen2025helmetevaluatelongcontextlanguage,bai2024longbenchbilingualmultitaskbenchmark,zhang2024inftybenchextendinglongcontext} . Overall, contextual-memory optimization should be co-designed with RAG to achieve auditable trade-offs among accuracy–cost–interpretability. Additionally, for edited knowledge in long contexts, we recommend integrating LEME long-form evaluation into the joint reporting of position–performance curves and length slopes \citep{rosati2024longformevaluationmodelediting} .
\subsubsection{External Memory: Updatable Indices and Evidence Fusion}At inference, external memory injects auditable evidence via retrieve–rerank–fuse, turning "knowledge updates" into re-embedding / index increments / rerank refreshes. Representative paradigms include RAG \citep{lewis2021retrievalaugmentedgenerationknowledgeintensivenlp} , REALM (jointly training retriever and LM) \citep{guu2020realmretrievalaugmentedlanguagemodel} , RETRO (cross-attention to massive external stores) \citep{borgeaud2022retro} , ATLAS (retrieval augmentation in few-shot settings) \citep{izacard2022atlasfewshotlearningretrieval} , and kNN-LM (local posterior correction) \citep{khandelwal2020generalizationmemorizationnearestneighbor} ; evaluation/alignment can follow KILT \citep{petroni2021kiltbenchmarkknowledgeintensive} .

Evidence thresholds and metric bundle. In high-risk domains, use evidence thresholds: trigger refusal/degradation on low-confidence retrieval–attribution; report citation coverage and unsupported-claim rate in tandem.
— Retrievers: DPR/ColBERT/Contriever \citep{karpukhin2020densepassageretrievalopendomain,khattab2020colbertefficienteffectivepassage,izacard2022unsuperviseddenseinformationretrieval} ;
— Diagnostics \& quality: RAGAS, RAGChecker, ARES, CRAG, RankRAG, RAMDocs, RGB \citep{es2025ragasautomatedevaluationretrieval,ru2024ragcheckerfinegrainedframeworkdiagnosing,saadfalcon2024aresautomatedevaluationframework,yang2024cragcomprehensiverag,wang2025retrievalaugmentedgenerationconflictingevidence} ;
— Attribution \& faithfulness: FACTS, ALCE, FActScore, QAFactEval \citep{jacovi2025factsgroundingleaderboardbenchmarking,li2024towards,min2023factscorefinegrainedatomicevaluation,fabbri2022qafactevalimprovedqabasedfactual} ;
— Freshness: TimeQA, FreshLLMs, MIRAGE-Bench \citep{chen2021datasetansweringtimesensitivequestions,vu2023freshllmsrefreshinglargelanguage,thakur2025miragebenchautomaticmultilingualbenchmark} ;
— Tasks/heterogeneous benchmarks: KILT, BEIR \citep{petroni2021kiltbenchmarkknowledgeintensive,thakur2021beirheterogenousbenchmarkzeroshot} .

Online mapping: Citation Coverage, Unsupported Claim Rate, Retrieval Recall@k / nDCG@k, Conflict-Handled@k (from RAMDocs), Freshness hit rate. If RAGChecker/RankRAG low confidence triggers the gate, then degrade/refuse and rollback the index or lower retrieval weight \citep{ru2024ragcheckerfinegrainedframeworkdiagnosing,wang2025retrievalaugmentedgenerationconflictingevidence} . Pre-registered thresholds (examples, domain-adjustable): Citation Coverage $\geq$ 0.85; Unsupported Claim Rate$\leq 0.05$; Recall@5 $\geq$ 0.85; Freshness $\geq$ 0.80; Conflict-Handled@5 $\geq$ 0.80 for conflicting evidence. Falling short triggers gray rollback and root-causing (RAGChecker) plus retraining/re-embedding.

Pipeline layering: Index layer (versioning \& de-duplication, unified chunk–embed–rerank config with conflict detection) → Fusion layer (evidence thresholds, sentence/passage-level attribution, cross-source consistency) → Ops layer (periodic re-embedding, incremental indexing, snapshot rollback, and a "fact–evidence–timestamp" change log).
\subsubsection{Procedural/Episodic Memory: Session-Level Write and Replay}Cross-session consistency relies on explicit organization of events and timelines. In practice, use a pipeline of "event capture → summarization \& entity tables → timeline \& vectorized archiving → trigger-based replay." Generative Agents structure "observe–remember–reflect–plan" for sustained behavioral consistency; MemGPT uses fast/slow memory and virtual context to achieve cross-session management within limited windows \citep{park2023generativeagentsinteractivesimulacra,packer2024memgptllmsoperatingsystems} . LoCoMo evaluation shows timeline consistency and long-range dependencies remain current weaknesses; thus, in production, enable explicit references for key slots, configure TTL/decay and conflict merging; if mis-replay increases or timelines drift, revert to the previous memory-store version or temporarily suspend auto-replay \citep{maharana2024evaluatinglongtermconversationalmemory,wu2025longmemevalbenchmarkingchatassistants,tan2025membenchcomprehensiveevaluationmemory,pink2024assessing} .

For evaluation, use long-term consistency and timeline slices from LoCoMo/LongMemEval/MemAE/MemBench for offline regression; online, track mis-replay rate and temporal misalignment rate \citep{maharana2024evaluatinglongtermconversationalmemory,wu2025longmemevalbenchmarkingchatassistants,gong2019memorizing,tan2025membenchcomprehensiveevaluationmemory} .
\subsection{Memory Forgetting}We unify forgetting qualification as four-dimensional acceptance: thoroughness (verbatim/semantic/adversarial), utility retention, scalability, and sustainability. Evaluation centers on TOFU/MUSE/RWKU, with MIA as a supporting signal only—not standalone evidence of "successful forgetting" \citep{maini2024tofutaskfictitiousunlearning,shi2024musemachineunlearningsixway,jin2024rwkubenchmarkingrealworldknowledge,duan2024membershipinferenceattackswork} .

Threat model (brief). Attackers may (i) access the model as a black box and perform prompt engineering, or (ii) approximate the training distribution and induce verbatim reproduction \citep{carlini2021extracting,carlini2019secretsharerevaluatingtesting} . Defense objective: under the above capabilities, reduce the verbatim reproduction/extraction success for target samples/concepts to on-par with an "unseen" model (TOFU/MUSE) \citep{maini2024tofutaskfictitiousunlearning,shi2024musemachineunlearningsixway} , while maintaining retention-set utility $\geq 98$--$99\%$, and ensuring robustness to resurgence/adversarial return (RWKU) \citep{jin2024rwkubenchmarkingrealworldknowledge} .

Where to implement forgetting. At the optimization level, build on preference-alignment paradigms to suppress specified knowledge via negative preference or contrastive objectives, offering controlled trade-offs among forgetting quality–utility retention–training efficiency \citep{zhang2024negativepreferenceoptimizationcatastrophic} . At the representation level, remove or suppress specific concepts/capabilities in intermediate representations—e.g., reduce dangerous capabilities on WMDP while preserving general abilities, or apply closed-form linear concept erasure for verifiable suppression \citep{li2024wmdpbenchmarkmeasuringreducing,belrose2023leace,ravfogel2024linearadversarialconcepterasure,ravfogel2020nulloutguardingprotected,elazar2021amnesicprobingbehavioralexplanation} . At the system level, externalize facts to forget as retrievable entries and overlay outputs at inference, naturally gaining revocability and auditability \citep{lewis2021retrievalaugmentedgenerationknowledgeintensivenlp} .

Evaluation. TOFU constructs forget/retain pairs to test "as if never learned" behavior, and reveals resurgence risks due to format rephrasing \citep{maini2024tofutaskfictitiousunlearning} ; MUSE offers a more engineering-realistic composite evaluation across verbatim/knowledge memory, privacy, utility, scale, and order sustainability, indicating the difficulty of achieving thorough forgetting–low side effects–scalability simultaneously \citep{shi2024musemachineunlearningsixway} ; proxy evaluation for dangerous capabilities (WMDP + representation-level forgetting) and robustness to real-world knowledge resurgence (RWKU) show adversarial validation is indispensable \citep{li2024wmdpbenchmarkmeasuringreducing,jin2024rwkubenchmarkingrealworldknowledge,belrose2023leace,ravfogel2024linearadversarialconcepterasure} . Process-wise, first define forget/retain/neighborhood sets and compliance proofs, choose methods and set rollback points; then validate jointly across thoroughness (verbatim/semantic/adversarial)—utility retention—sequential/scale sustainability—recoverability, form metric cards and a failure-case library, and coordinate cascading deletions in data, index, caches, and logs. As a preemptive governance measure, reduce high repetition in training data to lower verbatim-memorization incentives \citep{kandpal2022deduplicatingtrainingdatamitigates,lee2022deduplicatingtrainingdatamakes} . To avoid misreading failed edits as successful forgetting, cross-validate conflict consistency, long-horizon persistence, and out-of-template generalization via ConflictEdit/Pitfalls, LEME, and EVOKE before/after forgetting \citep{hsueh2024editingmindgiantsindepth,rosati2024longformevaluationmodelediting,hu2023evoke} ; time-related deletions should reuse AToKE time-window settings \citep{yin2023historymatterstemporalknowledge} .

Four-dimensional acceptance for memory forgetting: dimensions, metrics, reference benchmarks, and pre-registered thresholds.
\subsection{From Research to Release: Selection and Governance}
\subsubsection{Selection Principles}To reach verifiable trade-offs among accuracy, auditability, and Ops cost, we recommend three principles:

External/procedural memory first. For scenarios with high timeliness, frequent changes, and traceability needs, prefer RAG/KG/procedural memory (non-parametric paths), enabling "update by switching stores" with rollback-friendly release \citep{lewis2021retrievalaugmentedgenerationknowledgeintensivenlp,petroni2021kiltbenchmarkknowledgeintensive} .

Parametric edits: prudent and reversible. Apply small-batch, revocable weight edits only to sparse and stable "hard facts." For sequential/lifelong settings, pair with AlphaEdit/WISE/SERAC for robustness and external gray rollout \citep{fang2025alphaeditnullspaceconstrainedknowledge,wang2024wiserethinkingknowledgememory,mitchell2022memorybasedmodeleditingscale} .

Closed-loop governance. Throughout the process, enforce versioning → canary → monitoring → rollback → audit; prioritize de-dup and cleaning in training and indexing to reduce downstream editing/forgetting burden \citep{925786,kandpal2022deduplicatingtrainingdatamitigates,lee2022deduplicatingtrainingdatamakes} .

\subsubsection{A "Six-Step Decision Flow" for Selection and Deployment}

\noindent S1 Identify timeliness and conflict. If knowledge is highly time-sensitive / conflict-prone $\rightarrow$ choose external memory first (RAG/RETRO/kNN-LM or SERAC) \citep{lewis2021retrievalaugmentedgenerationknowledgeintensivenlp,borgeaud2022retro,khandelwal2020generalizationmemorizationnearestneighbor,mitchell2022memorybasedmodeleditingscale} .

\noindent S2 Determine granularity of change. If it is a small set of clear facts $\rightarrow$ parametric editing (ROME/MEND/MEMIT; pair with AlphaEdit/WISE for sequential cases; gray with SERAC before merging) \citep{meng2023locatingeditingfactualassociations,mitchell2022fastmodeleditingscale,meng2023masseditingmemorytransformer,fang2025alphaeditnullspaceconstrainedknowledge,wang2024wiserethinkingknowledgememory,mitchell2022memorybasedmodeleditingscale} .

\noindent S3 Constrain consistency. If temporal evolution is involved $\rightarrow$ adopt time windows and versioned evidence (AToKE); if multi-entity/constraints are involved $\rightarrow$ event-level editing (ELKEN) \citep{yin2023historymatterstemporalknowledge,peng2024eventlevelknowledgeediting} .

\noindent S4 Pre-register thresholds. Before updating, set and publish thresholds: ESR/Locality/Drawdown/Generalization (parametric), Citation Coverage/Unsupported Claim/Freshness/Recall@k (external/retrieval) \citep{fabbri2022qafactevalimprovedqabasedfactual,min2023factscorefinegrainedatomicevaluation,jacovi2025factsgroundingleaderboardbenchmarking,es2025ragasautomatedevaluationretrieval,saadfalcon2024aresautomatedevaluationframework} .

\noindent S5 Canary and online monitoring. Roll out with low traffic; continuously track Citation Coverage, Unsupported Claim Rate, Recall@k / nDCG@k, Freshness, and the trends of ESR/Locality/Drawdown for edits \citep{es2025ragasautomatedevaluationretrieval,ru2024ragcheckerfinegrainedframeworkdiagnosing,yang2024cragcomprehensiverag} .

\noindent S6 Rollback and evidence hardening. If monitored metrics exceed thresholds, auto-degrade or rollback (index/edit); version and archive ``fact--evidence--time-window'' change records for audit \citep{925786,owasp2023llmtop10} .

\subsubsection{Deployment and Monitoring Standards}Version isolation: manage weights, retrieval indices, and procedural-memory logs under separate versioning; track cross-version dependencies via snapshots and fingerprints.

Canary ramp-up: ramp by business risk and data domain; validate thresholds on low-risk slices before expansion.

Online metrics: for external paths, report Citation Coverage, Unsupported Claim Rate, Recall@k/nDCG@k, Freshness; for parametric paths, report ESR, Locality, Drawdown, Generalization; for procedural memory, monitor mis-replay rate and temporal misalignment rate.

Triggers: when Unsupported Claim Rate or Drawdown exceed thresholds, or conflict-handling rate drops, automatically trigger degrade/refusal/rollback and localization (retrieval diagnosis and edited-neighborhood regression).

\subsubsection{Evidence-Backed Update/Forgetting ''Certificates'': Elements, Ranges, and Calibration}

We observe an emerging practice of documenting edits and forgetting events for verification and compliance. Synthesizing prior work
\citep{yin2023historymatterstemporalknowledge,fang2025alphaeditnullspaceconstrainedknowledge,wang2024wiserethinkingknowledgememory,peng2024eventlevelknowledgeediting,fabbri2022qafactevalimprovedqabasedfactual,min2023factscorefinegrainedatomicevaluation,es2025ragasautomatedevaluationretrieval,saadfalcon2024aresautomatedevaluationframework,rosati2024longformevaluationmodelediting,hsueh2024editingmindgiantsindepth,hu2023evoke,mitchell2022memorybasedmodeleditingscale,lewis2021retrievalaugmentedgenerationknowledgeintensivenlp,925786,owasp2023llmtop10}, we outline \emph{minimal elements} commonly reported and \emph{illustrative threshold ranges} that appear in the literature or industrial case studies. These references are not one-size-fits-all and should be calibrated to task risk, domain, and cost functions, with uncertainty and sensitivity analyses.

\begin{description}
  \item[\textbf{Target (F1).}] Fact/concept, sources, evidence links, snapshots.
  \item[\textbf{Time (F2).}] Effective/expiry timestamps; time-window settings (AToKE).
  \item[\textbf{Method (F3).}] Update/forgetting path (RAG/edit/PEFT/system override) and constraints (e.g., AlphaEdit/WISE/ELKEN).
  \item[\textbf{Verification (F4).}] Examples reported include ESR around $0.90$, Locality $\approx 0.95$, Drawdown $\le 1$--$2\%$, Citation Coverage $\ge 0.80$--$0.90$, Unsupported Claim $\le 0.05$--$0.10$, Recall@5 $\ge 0.80$--$0.90$, Freshness $\ge 0.75$--$0.85$; forgetting is evaluated along \emph{thoroughness--utility--scalability--sustainability}.
  \item[\textbf{Regression (F5).}] Long-form consistency (LEME), conflict/reversibility (ConflictEdit), and out-of-template/domain generalization (EVOKE).
  \item[\textbf{Rollback (F6).}] Rollback points/snapshots and impact-surface analysis; fallback via SERAC/RAG.
  \item[\textbf{Audit (F7).}] Version IDs, operation logs, ownership/approval chain; alignment with NIST AI RMF / OWASP LLM Top-10.
\end{description}

\noindent\textbf{Calibration \& Reporting Protocol}
\begin{itemize}
  \item \textbf{Controlled baselines.} Report PO/Offline/Online results on the \emph{same} data slice and time window, and include effect sizes ($\Delta_{\text{abs}}/\Delta_{\text{rel}}$) relative to the chosen baseline.
  \item \textbf{Uncertainty.} Provide \mbox{95\%} confidence intervals (bootstrap) and \emph{paired} permutation or bootstrap tests with Holm--Bonferroni or FDR correction for multi-metric comparisons; state sample sizes.
  \item \textbf{Risk-tiered targets.} Map metrics to risk tiers (low/medium/high impact) derived from domain cost functions; use the ranges above as illustrative defaults, then pre-register study-specific targets or justify deviations.
  \item \textbf{Sensitivity/robustness.} Report sensitivity to decoding/retrieval knobs (e.g., temperature, top-$p$, $k$, fusion strategy) and to snapshot timing; include at least one \emph{time-slice robustness} check to rule out freshness confounds.
  \item \textbf{Evidence mapping.} For each metric, cite representative sources supporting the chosen range; an evidence table collating distributions and typical values appears  (with pointers to datasets, domains, and evaluation conditions).
\end{itemize}

\noindent\textbf{Positioning in this survey.} Within this survey, F1--F7 are presented as reporting recommendations, together with observed ranges. Concrete deployment thresholds are calibrated on a per-study basis and reported with uncertainty estimates, sensitivity analyses, and baseline-controlled effect sizes.

\noindent\textbf{Limitations.} Reported ranges can shift with dataset curation, leakage controls, and evaluator design; we therefore emphasize comparability within a shared slice and encourage release of machine-readable ``certificate'' artifacts to support independent re-analysis.

\section{Challenges and Future Directions}
Under a unified definition and a three-regime evaluation protocol (PO / Offline-Retrieval / Online-Retrieval), this paper juxtaposes the in-parameter and extra-parameter memory channels and presents an integrated framework spanning mechanisms—evaluation—updating/forgetting. Despite rapid progress, several core problems and methodological gaps around "memory" still directly constrain comparability, auditability, and large-scale deployment. Below we survey key challenges across four dimensions—theoretical representation, evaluation methodology, governability of updating/forgetting, and system deployment—and propose testable propositions for follow-up research.

\subsection{Theory and Representation: Causal Localization and Representation Entanglement Remain Unresolved}
(1) \textbf{Insufficient verifiability of causal localization.} Existing evidence supports encoding and read out of knowledge in mid-layer MLPs in a key–value form \citep{geva2021transformerfeedforwardlayerskeyvalue,meng2023locatingeditingfactualassociations,elhage2021mathematical}, but such localization mostly relies on intervention heuristics and correlational observations. How to establish, without strong structural priors, a falsifiable causal chain (e.g., quantifying the mediation effect of ``edit $\rightarrow$ mediator $\rightarrow$ output'') still lacks a unified paradigm.

(2) \textbf{Entanglement between knowledge and ability.} In-context learning can be viewed as circuit-level copy-and-align behavior \citep{olsson2022incontextlearninginductionheads}, or characterized as implicit Bayesian updating or approximate gradient descent under fixed weights \citep{akyurek2023what,vonoswald2023transformerslearnincontextgradient}. These two views have not yet been unified in terms of time scales and generalization boundaries, yielding empirical phenomena where ability loading and knowledge access are hard to distinguish.

(3) \textbf{Scaling laws of memorization and controllability.} Prior work suggests that verbatim memorization and privacy risk are tightly related to data entropy, repetition, and model scale \citep{carlini2019secretsharerevaluatingtesting,carlini2023quantifyingmemorizationneurallanguage}, while de-duplication mitigates risk and may even improve quality \citep{lee2022deduplicatingtrainingdatamakes,kandpal2022deduplicatingtrainingdatamitigates}. Yet there is no unified characterization of the triad (parameter scale—editability—leakage rate); reproducible studies across distributions and scales with uncertainty quantification are needed.

\emph{Open Proposition A.} Without strong structural priors, do minimal identifiability conditions exist for ``knowledge loci,'' and can they be reproduced across model families?

\subsection{Evaluation Methodology: Harmonized Protocols and Temporal Governance Are Still Incomplete}
(1) \textbf{Inconsistent notions of passage-level groundedness.} It is now agreed in RAG that ``correctness $\neq$ faithfulness,'' but passage-level source attribution and citation coherence remain incompatible across benchmarks \citep{min2023factscorefinegrainedatomicevaluation,li2024towards,es2025ragasautomatedevaluationretrieval,ru2024ragcheckerfinegrainedframeworkdiagnosing,saadfalcon2024aresautomatedevaluationframework,jacovi2025factsgroundingleaderboardbenchmarking}. The lack of unified annotation granularity, scoring rules, and confidence estimation hinders apples-to-apples comparison.

(2) \textbf{For long context, ``visible $\neq$ usable.''} \emph{Lost in the Middle} and \emph{Found in the Middle} reveal systematic positional bias \citep{liu2023lostmiddlelanguagemodels,hsieh2024middlecalibratingpositionalattention}, yet many evaluations substitute window visibility for usability. Standard reporting of position–performance curves and mid-span drop is needed to compare marginal effects of window expansion, reordering, and reweighting.

(3) \textbf{Reviewer drift and statistical validity.} Instability of LLM-as-a-Judge with respect to order and self-preference has been repeatedly documented \citep{wang2023largelanguagemodelsfair,zheng2023judgingllmasajudgemtbenchchatbot}. Without adjudication protocols that include confidence intervals and multiple-comparison correction, small improvements cannot be distinguished from noise.

(4) \textbf{Missing temporal dimension.} Current benchmarks under-specify freshness and conflict. Although works like TimeQA and FreshLLMs point the way \citep{chen2021datasetansweringtimesensitivequestions,Dhingra_2022,vu2023freshllmsrefreshinglargelanguage}, a unified standard for timestamp alignment and versioned reporting has yet to form.

\emph{Open Proposition B.} With the three regimes run in parallel as external conditions, construct a minimally sufficient evaluation card across tasks, lengths, and time—covering correctness, faithfulness, positional robustness, timeliness, and refusal—and provide unified norms for CIs and significance tests. As a companion, this paper recommends extending the edit/forgetting certificate with pre-registered equivalence margins and three-way slices—reporting effectiveness, locality, downstream steady state, and rollbackability—forming auditable artifacts, consistent with recent editing evaluations \citep{rosati2024longformevaluationmodelediting,hsueh2024editingmindgiantsindepth,hu2023evoke}.

\subsection{Updating and Forgetting: Effectiveness, Locality, and Scalability Are Hard to Achieve Simultaneously}
(1) \textbf{Cascade side effects of pointwise/batch editing.} ROME/MEND/MEMIT demonstrate the feasibility of addressable rewrites \citep{meng2023locatingeditingfactualassociations,mitchell2022fastmodeleditingscale,meng2023masseditingmemorytransformer,mitchell2022memorybasedmodeleditingscale}, yet in serial and large-batch settings they often exhibit neighborhood spillover and downstream regression, depending on model scale, locus selection, and data distribution.

(2) \textbf{Verifiability and recoverability of forgetting.} Preference- or representation-level unlearning reduces extractability on target sets, but the strong criterion of ``as if never learned'' remains unstable under real-world and adversarial distributions \citep{maini2024tofutaskfictitiousunlearning,shi2024musemachineunlearningsixway,jin2024rwkubenchmarkingrealworldknowledge,bourtoule2021machine,zhang2024negativepreferenceoptimizationcatastrophic,ji2024reversingforgetretainobjectivesefficient}. Without falsifiable and recoverable dual standards, audit costs stay high.

(3) \textbf{Ternary trade-off of privacy—utility—cost.} De-duplication and data governance markedly reduce regurgitation and extraction \citep{lee2022deduplicatingtrainingdatamakes,kandpal2022deduplicatingtrainingdatamitigates,carlini2021extracting}, but unified quantification of coverage/perplexity impact across thresholds is still lacking.

\emph{Open Proposition C.} Do combinations of \emph{causally constrained editing} and \emph{verifiable unlearning} exist that attain a provable Pareto frontier over effectiveness, locality, and scalability?

\subsection{Systems and Deployment: Joint Governance of Conflicting Evidence, Cost, and Accountability}
(1) \textbf{Retrieval bottlenecks and error amplification.} End-to-end performance is highly sensitive to recall; the retrieve–rerank–fuse error chain is amplified with long documents and cross-source conflicts \citep{salemi2024evaluatingretrievalqualityretrievalaugmented,saadfalcon2024aresautomatedevaluationframework,ru2024ragcheckerfinegrainedframeworkdiagnosing,thakur2021beirheterogenousbenchmarkzeroshot,wang2025retrievalaugmentedgenerationconflictingevidence,joren2025sufficientcontextnewlens}. Task-driven document utility should be jointly reported with refusal/abstention metrics to avoid ``surface-level correctness without evidence.''

(2) \textbf{Long-horizon consistency and mis-replay.} In multi-session/agentic scenarios, expanding the window alone cannot maintain timeline consistency; without eventification and timeline structuring, mis-replay and inconsistency increase \citep{wu2025longmemevalbenchmarkingchatassistants,hu2025evaluatingmemoryllmagents,packer2024memgptllmsoperatingsystems,park2023generativeagentsinteractivesimulacra}.

(3) \textbf{Operations and compliance.} Versioned indices, snapshot rollback, and audit trails are hard constraints for real systems; academic reports rarely disclose the functional relationship among cost–latency–freshness or rollback strategies. In line with NIST AI-RMF and OWASP LLM Top-10 \citep{925786,owasp2023llmtop10}, answer-only-with-evidence hard gates and cascading deletion/audit should be built into evaluation/release protocols.

\emph{Open Proposition D.} Under fixed latency/cost budgets, can \emph{high-quality retrieval + reranking + small-window replay} systematically outperform \emph{ultra-long-context direct reading}? Boundary conditions and cross-domain transferability call for a unified protocol (cf. long-context evaluations \citep{hsieh2024rulerwhatsrealcontext,yen2025helmetevaluatelongcontextlanguage,bai2024longbenchbilingualmultitaskbenchmark}).

\subsection{Future Research Directions}
(i) \textbf{Unified memory semantics and observable criteria.} Use the ``four-way taxonomy + quadruple'' (storage location—persistence—write/access path—controllability) as a semantic anchor; add formal observable equivalence classes and cross-regime invariance analyses.

(ii) \textbf{Causally consistent editing and verifiable unlearning.} Combine circuit-level mediation analysis with constrained optimization; construct forgetting certificates with counterfactual tests and inverse edits \citep{rosati2024longformevaluationmodelediting,hsueh2024editingmindgiantsindepth,maini2024tofutaskfictitiousunlearning,shi2024musemachineunlearningsixway}.

(iii) \textbf{Time-aware evaluation and governance.} Under unified snapshot and dynamic-update scenarios, adopt timestamp-aligned protocols and metrics such as Fresh@k / Outdated\% / Refusal@Stale \citep{chen2021datasetansweringtimesensitivequestions,Dhingra_2022,vu2023freshllmsrefreshinglargelanguage}, reported jointly with citation coverage.

(iv) \textbf{Read/write orchestration for long context.} Develop learnable strategies combining positional reweighting, anchor-based reordering, and summary–replay; treat position–performance curves as a first-class metric \citep{hsieh2024middlecalibratingpositionalattention,liu2023lostmiddlelanguagemodels,hsieh2024rulerwhatsrealcontext,yen2025helmetevaluatelongcontextlanguage}.

(v) \textbf{Explainable fusion of external memory.} Under cross-source conflict and noise, introduce consistency adjudication—refusal—uncertainty estimation for joint optimization of correctness—faithfulness—auditability \citep{saadfalcon2024aresautomatedevaluationframework,ru2024ragcheckerfinegrainedframeworkdiagnosing,friel2025ragbenchexplainablebenchmarkretrievalaugmented,jacovi2025factsgroundingleaderboardbenchmarking}.

(vi) \textbf{Transferable quantification of privacy risk.} Using data entropy, repetition, and scaling laws as covariates, build predictive models across corpora and models, offering interval estimates and conservative bounds for regurgitation/extraction/membership inference \citep{carlini2019secretsharerevaluatingtesting,carlini2023quantifyingmemorizationneurallanguage,kandpal2022deduplicatingtrainingdatamitigates,lee2022deduplicatingtrainingdatamakes}.

Overall, memory is neither a single substrate nor a single capability, but a multilayer collection of states and mechanisms co-determined by training objectives, inference dynamics, and system architecture. Coordinated advances in causally consistent mechanistic evidence, time-aware evaluation protocols, verifiable editing/unlearning, and auditable system governance will help capability—timeliness—controllability—cost reach higher levels within a unified experimental and engineering coordinate system.

\section{Conclusion}This paper presents a systematic survey and methodological synthesis around memory in large language models (LLMs). Under a unified operationalized definition, we propose a four-way taxonomy (parametric memory, contextual memory, external memory, procedural/episodic memory) and a "memory quadruple" (storage location—persistence—write/access path—controllability), and we connect mechanistic evidence, evaluation protocols, and engineering governance through a causal chain of "write—read—inhibit/update." To mitigate distorted comparisons caused by heterogeneous research setups, we construct three parallel operating regimes—parametric-only (PO), offline retrieval augmentation (Offline-Retrieval), and online retrieval augmentation (Online-Retrieval)—to separate the contributions of information availability and model capability on the same samples and timeline.

In evaluation methodology, we provide layered metric systems and reproducible protocols covering the four memory types: for parametric memory, we emphasize factual recall under the PO setting, pre/post-edit differentials, and memorization/privacy risks; for contextual memory, we foreground the diagnosis that "visible $\neq$ usable," with primary reporting via position–performance curves and mid-span drop; for external memory, we decouple retrieval quality from faithfulness/source attribution and jointly report correctness and passage-level groundedness; for procedural/episodic memory, we adopt componentized evaluation of cross-session consistency, timeline replay, and mis-replay control. All the above metrics are uniformly brought under a temporal dimension (Fresh@k, Outdated\%, Refusal@Stale) and uncertainty reporting (confidence intervals, paired tests, and multiple-comparison corrections), complemented by human–AI mixed adjudication to reduce the drift risk of LLM-as-a-Judge.

For updating and forgetting, we place continued pretraining and parameter-efficient finetuning (PEFT), model editing (e.g., ROME/MEND/MEMIT), and external override (RAG/SERAC) on a single decision surface, describe their inherent trade-offs via a three-axis Pareto analysis (target suppression—neighborhood preservation—downstream steady state), and propose a minimal governance closed loop for release: versioning (weights/index/logs managed separately), canary and rollback, online monitoring (citation coverage, unsupported-claim rate, timeliness/conflict slices), together with systematic support for compliant deletion and audit trails. To address the challenge of verifiable forgetting, we propose an "edit/forgetting certificate" reporting framework that uses pre-registered equivalence margins and counterfactual validation to document evidence of effectiveness, locality, and recoverability, thereby turning research findings into auditable artifacts.

Synthesizing evidence across components, our main findings can be summarized in four points:
(1) Memory is not a single substrate; it is a multi-layer collection of states shaped jointly by training objectives, network structure, and system architecture.
(2) The upper bound of long-context capability is constrained by positional bias and attention dilution; structured reordering and anchor-based replay are often more cost-effective than blind window expansion.
(3) Retrieval recall is the dominant bottleneck of RAG; reranking and answer-only-with-evidence hard gates yield synchronous gains in correctness and faithfulness.
(4) Editing/unlearning rarely attains effectiveness, locality, and scalability simultaneously; small steps, reversibility, and external coverage first are needed to reduce cascading spillover.

This work has three main limitations: first, although we advocate unified protocols and offer reproducibility guidance, absolute comparability across datasets and implementations remains constrained by public resources and implementation details; second, some evaluations rely on proxy tasks or static snapshots, still short of real dynamic environments; third, formal conditions for causally consistent editing and verifiable unlearning have yet to reach consensus and require further testing across model families and scales. To this end, Chapter 6 proposed four open propositions—identifiability conditions, a minimally sufficient evaluation card, causally constrained editing \& verifiable unlearning, and the boundary where retrieval + small-window replay outperforms ultra-long-window direct reading—as testable paths for subsequent work.

Overall, our contributions are: providing a deployable unified terminology and typed framework; bringing correctness, faithfulness, positional robustness, and timeliness into a single evaluation coordinate system via "three regimes in parallel" and layered metrics; situating updating/unlearning within engineering governance, with reversible, auditable implementation guidelines and certificate-style reporting; and distilling a practice checklist of "external-memory-first—small-step editing—timestamp alignment—evidence grounding—audit trails." We hope these methodological baselines and governance points will supply a shared coordinate system for research and industrial deployment alike, advancing LLM memory research toward comparability, deployability, and governability.

\clearpage
\bibliographystyle{unsrt}  
\bibliography{references}  %%% Remove comment to use the external .bib file (using bibtex).}
%%% and comment out the ``thebibliography'' section.

\clearpage
% ---------- Appendix ----------
\appendix
\renewcommand{\thesection}{Appendices\Alph{section}} % 附录标题显示为"附录A""附录B"

\section{Minimal MRD YAML template}\label{app:mrd-template}
% Requires: \usepackage{listings} (and \usepackage[utf8]{inputenc} if needed)
% \begin{lstlisting}[language=YAML]
\begin{lstlisting}[language=yaml,
  literate={∩}{{$\cap$}}1
]
mrd_version: "0.1"
study_id: "<short-identifier>"
claim_type: ["capability_change", "freshness_related"]  # pick one or both

temporal_governance:
  train:
    window: "YYYY-MM-DD..YYYY-MM-DD"
    snapshot_date: "YYYY-MM-DD"
    sources: ["<corpus1>", "<corpus2>"]
  index_or_session_store:
    snapshot_date: "YYYY-MM-DD"
    dedup_strategy: "<e.g., MinHash@J=0.85, URL+title exact>"
    update:
      last_time: "YYYY-MM-DDThh:mm:ssZ"
      frequency: "<e.g., daily | weekly | ad-hoc>"
  test:
    window: "YYYY-MM-DD..YYYY-MM-DD"
    snapshot_date: "YYYY-MM-DD"
    contamination_notes: "<potential benchmark leakage & mitigation>"

leakage_overlap_auditing:
  methods: ["<e.g., SimHash, BM25-topk cross-match>"]
  thresholds: {"near_dup_jaccard": 0.85}
  exclusion_criteria: "<rules applied>"
  impact_fraction:
    train∩test: 0.012
    train∩index: 0.034
    index∩test: 0.007

implementation_resources:
  model:
    name: "<model-family>"
    checkpoint: "<hash/tag>"
    decoding_hparams: {temperature: 0.7, top_p: 0.95, max_new_tokens: 512}
    random_seeds: [2024, 2025, 2026]
    script_versions: {"runner": "v1.3.2", "evaluator": "v0.9.1"}
  retrieval:
    retriever_type: "<e.g., BM25 | DPR | ColBERT | hybrid>"
    k: 10
    fusion_strategy: "<e.g., RRF@60 | concat-then-rerank>"
    reranker_type: "<e.g., Cross-Encoder-msmarco>"
    core_params: {"max_passages": 50}
  hardware_cost:
    accelerators: {"A100_80G": 8}
    total_gpu_hours: 120.5
    per_unit_cost_usd: 1.8
    token_budget: {"prompt_tokens": 1.2e8, "completion_tokens": 6.5e7}

regimes:  # ensure shared slice/time-window when comparing
  - name: "PO"
    shared_slice_with: ["Offline", "Online"]
  - name: "Offline"
  - name: "Online"

limitations_and_robustness:
  undisclosed_fields: ["<e.g., exact costs>"]
  reason: "<compliance | NDA | privacy>"
  sensitivity_analyses: ["<brief pointer to what was varied and the effect>"]
\end{lstlisting}

\end{document}